%% file: main.tex
\documentclass[10pt, letter, onecolumn]{arxiv}

\usepackage{kantlipsum, lipsum}
\usepackage{dm-colors}
\usepackage{amsmath}
\usepackage{pstricks, pst-node}
\usepackage{verbatim}
\usepackage{multirow}
\usepackage{scalerel}
\usepackage{booktabs}
\usepackage{enumitem}
\usepackage{xspace}
\usepackage{bm}
\usepackage{bbm}
\usepackage{mathtools}
\usepackage{soul}
\usepackage{epsfig}
\usepackage{graphicx}
\usepackage{amssymb}
\usepackage{colortbl}
\usepackage{csquotes}
\usepackage{etex}
\usepackage{setspace}
\usepackage{colortbl}
\usepackage{tabularx,ragged2e}
\usepackage{placeins}
\usepackage[symbol]{footmisc}
\usepackage[bibstyle=nature,citestyle=numeric-comp,%
            natbib=true,backend=biber,maxbibnames=99,%
            giveninits=false]{biblatex}
\usepackage{nameref}
\usepackage{varioref}
\usepackage[pagebackref=false,breaklinks=false,%
            colorlinks=true,bookmarks=true,citecolor=ourdarkblue,%
            urlcolor=ourdarkblue,linkcolor=ourdarkblue]{hyperref}
\usepackage[noabbrev,capitalize]{cleveref}
\usepackage{etoc}

\graphicspath{{figures/}}

\title{{Large Language Models Encode \\ Clinical Knowledge}}
\author[$\ast$,1]{Karan Singhal}
\author[$\ast$,1]{Shekoofeh Azizi}
\author[$\ast$,1]{Tao Tu}
\author[1]{\\S. Sara Mahdavi}
\author[1]{Jason Wei}
\author[1]{Hyung Won Chung}
\author[1]{Nathan Scales}
\author[1]{Ajay Tanwani}
\author[1]{\\Heather Cole-Lewis}
\author[1]{Stephen Pfohl}
\author[1]{Perry Payne}
\author[1]{Martin Seneviratne}
\author[1]{Paul Gamble}
\author[1]{Chris Kelly}
\author[1]{Nathaneal Schärli}
\author[1]{Aakanksha Chowdhery}
\author[1]{Philip Mansfield}
\author[1]{Blaise Agüera y Arcas}
\author[1]{\\Dale Webster}
\author[1]{Greg S. Corrado}
\author[1]{Yossi Matias}
\author[1]{Katherine Chou}
\author[1]{Juraj Gottweis}
\author[2]{\\Nenad Tomasev}
\author[1]{Yun Liu}
\author[1]{Alvin Rajkomar}
\author[1]{Joelle Barral}
\author[1]{Christopher Semturs}
\author[$\dagger$,1]{\\Alan Karthikesalingam}
\author[$\dagger$,1]{Vivek Natarajan}

\affil[1]{Google Research, }
\affil[2]{DeepMind}

\renewcommand{\correspondingauthor}[1]{$\ast$~Equal contributions. %
                                       $\dagger$~Equal leadership. \\%
                                       $\ddagger$~Corresponding authors: \{karansinghal, shekazizi, alankarthi, natviv\}@google.com }

\begin{abstract}

Large language models (LLMs) have demonstrated impressive capabilities in natural language understanding and generation, but the quality bar for medical and clinical applications is high. Today, attempts to assess models’ clinical knowledge typically rely on automated evaluations on limited benchmarks. There is no standard to evaluate model predictions and reasoning across a breadth of tasks. To address this, we present MultiMedQA, a benchmark combining six existing open question answering datasets spanning professional medical exams, research, and consumer queries; and HealthSearchQA, a new free-response dataset of medical questions searched online. We propose a framework for human evaluation of model answers along multiple axes including factuality, precision, possible harm, and bias.

In addition, we evaluate PaLM (a 540-billion parameter LLM) and its instruction-tuned variant, Flan-PaLM, on MultiMedQA. Using a combination of prompting strategies, Flan-PaLM achieves state-of-the-art accuracy on every MultiMedQA multiple-choice dataset (MedQA, MedMCQA, PubMedQA, MMLU clinical topics), including 67.6\% accuracy on MedQA (US Medical License Exam questions), surpassing prior state-of-the-art by over 17\%. However, human evaluation reveals key gaps in Flan-PaLM responses. To resolve this we introduce instruction prompt tuning, a parameter-efficient approach for aligning LLMs to new domains using a few exemplars. The resulting model, Med-PaLM, performs encouragingly, but remains inferior to clinicians.

We show that comprehension, recall of knowledge, and medical reasoning improve with model scale and instruction prompt tuning, suggesting the potential utility of LLMs in medicine. Our human evaluations reveal important limitations of today’s models, reinforcing the importance of both evaluation frameworks and method development in creating safe, helpful LLM models for clinical applications.

\end{abstract}

\begin{document}

\maketitle
\begin{refsection}

\section{Introduction}
\label{sec:introduction}

\begin{figure*}[t]
\small
    \centering
    \includegraphics[width=1.0\textwidth]{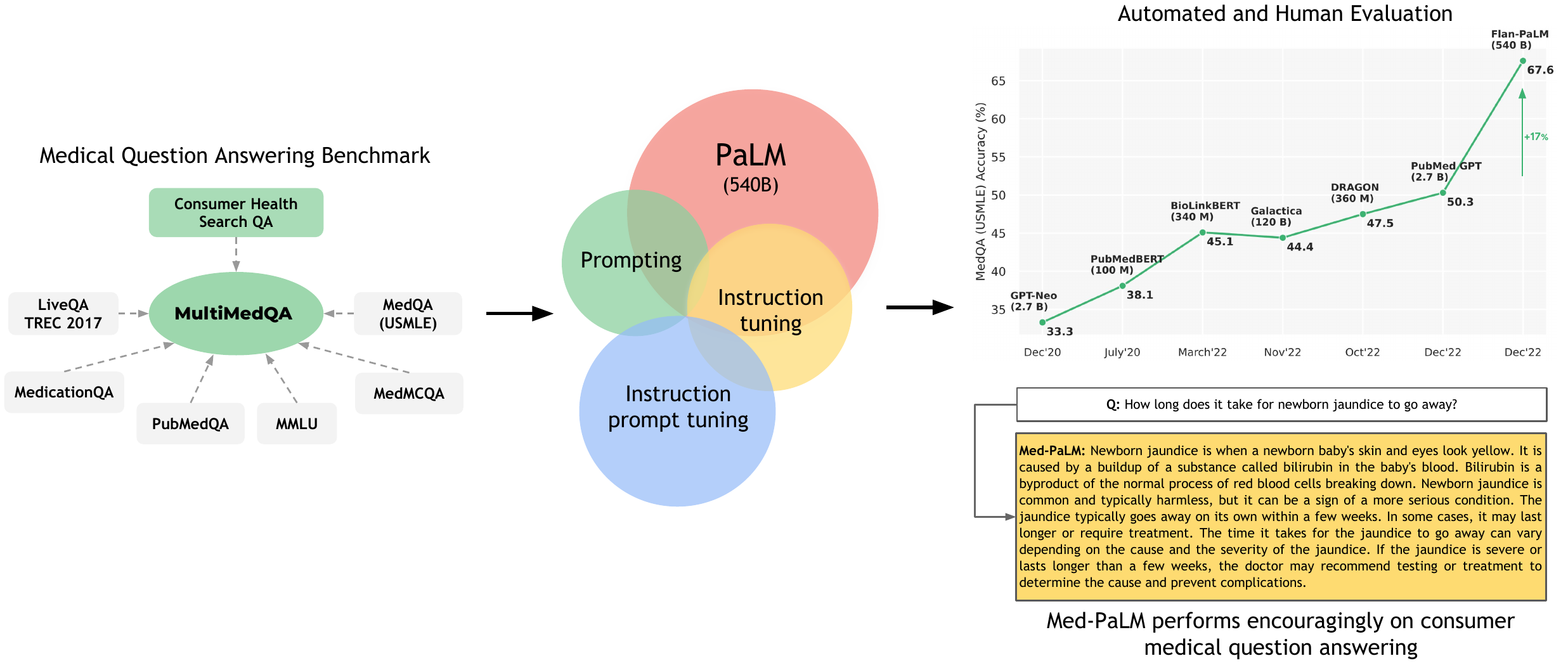}
    \vspace{0pt}
    \caption{\textbf{Overview of our contributions} We curated MultiMedQA, a benchmark for medical question answering spanning medical exam, medical research, and consumer medical questions. We evaluated PaLM and its instructed-tuned variant, Flan-PaLM, on MultiMedQA. With a combination of prompting strategies, Flan-PaLM exceeded SOTA performance on MedQA (USMLE), MedMCQA, PubMedQA, and MMLU clinical topics. In particular, it improved over the previous SOTA on MedQA (USMLE) by over 17\%. We next proposed instruction prompt tuning to further align Flan-PaLM to the medical domain, producing Med-PaLM. Med-PaLM’s answers to consumer medical questions compared favorably with clinician-generated answers under our human evaluation framework, demonstrating the effectiveness of instruction prompt tuning.}
    \vspace{-0pt}
    \label{fig:contributions-overview}
\end{figure*}

Medicine is a humane endeavor where language enables key interactions for and between clinicians, researchers, and patients. Yet, today’s AI models for applications in medicine and healthcare have largely failed to fully utilize language. These models, while useful, are predominantly single-task systems (e.g., classification, regression, segmentation), lacking expressivity and interactive capabilities~\cite{esteva2021deep, tomavsev2021use, yim2020predicting}. As a result, there is a discordance between what today’s models can do and what may be expected of them in real-world clinical workflows~\cite{lakkaraju2022rethinking, schaekermann2020expert}.

Recent advances in large language models (LLMs) offer an opportunity to rethink AI systems, with language as a tool for mediating human-AI interaction. LLMs are “foundation models”~\cite{bommasani2021opportunities}, large pre-trained AI systems that can be repurposed with minimal effort across numerous domains and diverse tasks. These expressive and interactive models offer great promise in their ability to learn generally useful representations from the knowledge encoded in medical corpora, at scale. There are several exciting potential applications of such models in medicine, including knowledge retrieval, clinical decision support, summarisation of key findings, triaging patients’ primary care concerns, and more.

However, the safety-critical nature of the domain necessitates thoughtful development of evaluation frameworks, enabling researchers to meaningfully measure progress and capture and mitigate potential harms. This is especially important for LLMs, since these models may produce generations misaligned with clinical and societal values. They may, for instance, hallucinate convincing medical misinformation or incorporate biases that could exacerbate health disparities.

To evaluate how well LLMs encode clinical knowledge and assess their potential in medicine, we consider medical question answering. This task is challenging: providing high-quality answers to medical questions requires comprehension of medical context, recall of appropriate medical knowledge, and reasoning with expert information. Existing medical question answering benchmarks~\cite{jin2021disease} are often limited to assessing classification accuracy or automated natural language generation metrics (e.g., BLEU~\cite{papineni2002bleu}), and do not enable the detailed analysis required for real-world clinical applications. This creates an unmet need for a broad medical question answering benchmark to assess LLMs’ response factuality, use of expert knowledge in medical and scientific reasoning, helpfulness, precision, health equity, and potential harm to humans accepting model outputs as facts.  

To address this, we curate MultiMedQA, a benchmark comprising seven medical question answering datasets, including six existing datasets: MedQA~\cite{jin2021disease}, MedMCQA~\cite{pal2022medmcqa}, PubMedQA~\cite{jin2019pubmedqa}, LiveQA~\cite{abacha2017overview}, MedicationQA~\cite{abacha2019bridging}, and MMLU clinical topics~\cite{hendrycks2020measuring}. We newly introduce the seventh dataset, HealthSearchQA, which consists of commonly searched health questions.

To assess LLMs using MultiMedQA, we build on PaLM, a 540-billion parameter LLM~\cite{chowdhery2022palm}, and its instruction-tuned variant Flan-PaLM~\cite{chung2022scaling}. Using a combination of few-shot~\cite{brown2020language}, chain-of-thought (CoT)~\cite{wei2022chain}, and self-consistency~\cite{wang2022self} prompting strategies, Flan-PaLM achieves state-of-the-art (SOTA) performance on MedQA, MedMCQA, PubMedQA, and MMLU clinical topics, often outperforming several strong LLM baselines by a significant margin. On the MedQA dataset comprising USMLE questions, FLAN-PaLM exceeds previous SOTA by over 17\%.

Despite Flan-PaLM's strong performance on multiple-choice questions, its answers to consumer medical questions reveal key gaps. To resolve this, we propose instruction prompt tuning, a data- and parameter-efficient alignment technique, to further adapt Flan-PaLM to the medical domain. The resulting model, Med-PaLM, performs encouragingly on the axes of our pilot human evaluation framework. For example, a panel of clinicians judged only 61.9\% of Flan-PaLM long-form answers to be aligned with scientific consensus, compared to 92.6\% for Med-PaLM answers, on par with clinician-generated answers (92.9\%). Similarly, 29.7\% of Flan-PaLM answers were rated as potentially leading to harmful outcomes, in contrast with 5.8\% for Med-PaLM, comparable with clinician-generated answers (6.5\%).

While these results are promising, the medical domain is complex. Further evaluations are necessary, particularly along the dimensions of fairness, equity, and bias. Our work demonstrates that many limitations must be overcome before such models become viable for use in clinical applications. We outline some key limitations and directions of future research in our study.

Our key contributions are summarized below:

\begin{itemize}[leftmargin=1.5em,rightmargin=0em]
\setlength\itemsep{5pt}
\item \textbf{Approaches for evaluation of LLMs in medical question answering}

\begin{itemize}
    \item[-] \textbf{Curation of HealthSearchQA and MultiMedQA} We introduce HealthSearchQA, a dataset of 3375 commonly searched consumer medical questions. We present this dataset alongside six other existing open datasets for medical question answering, spanning medical exam, medical research, and consumer medical questions, as a diverse benchmark to assess the clinical knowledge and question answering capabilities of LLMs (see \cref{sec:datasets}).
    \item[-] \textbf{Pilot framework for human evaluation} We pilot a framework for physician and lay user evaluation to assess multiple axes of LLM performance beyond accuracy on multiple-choice datasets. Our evaluation assesses answers for agreement with scientific and clinical consensus, likelihood and possible extent of harm, reading comprehension, recall of relevant clinical knowledge, manipulation of knowledge via valid reasoning, completeness of responses, potential for bias, relevance, and helpfulness (see \cref{sec:framework-for-human-eval}).
    
\end{itemize}

\item \textbf{State-of-the-art results on medical question answering benchmarks} On the MedQA, MedMCQA, PubMedQA and MMLU clinical topics datasets, FLAN-PaLM achieves SOTA performance via a combination of prompting strategies, surpassing several strong LLM baselines. Specifically, we reach 67.6\% accuracy on MedQA (more than 17\% above prior SOTA), 57.6\% on MedMCQA, and 79.0\% on PubMedQA (see \cref{sec:results}).

\item \textbf{Instruction prompt tuning to align LLMs to the medical domain} We introduce instruction prompt tuning, a simple, data- and parameter-efficient technique for aligning LLMs to the safety-critical medical domain (see \cref{sec:instruction_prompt_tuning}). We leverage this to build Med-PaLM, an instruction prompt-tuned version of Flan-PaLM specialized for the medical domain. Our human evaluation framework reveals limitations of Flan-PaLM in scientific grounding, harm, and bias. However, Med-PaLM significantly reduces the gap (or even compares favorably) to clinicians on several of these axes, according to both clinicians and lay users (see \cref{sec:human_evaluation_results}).

\item \textbf{Key limitations of LLMs revealed through our human evaluation} While our results demonstrate the potential of LLMs in medicine, they also suggest several critical improvements are necessary in order to make these models viable for real-world clinical applications. We outline future research directions and mitigation strategies to address these challenges (see \cref{sec:limitations}).

\end{itemize}

\section{Related work}
\label{sec:related_work}
\paragraph{Large language models (LLMs)} Over the past few years, LLMs have shown impressive performance on natural language processing (NLP) tasks ~\cite{brown2020language, chowdhery2022palm, chung2022scaling, hoffmann2022training, scao2022bloom, rae2021scaling, raffel2020exploring, wei2021finetuned, wei2022chain, zhang2022opt}. They owe their success to scaling up the training of transformer-based models~\cite{vaswani2017attention}. It has been shown that model performance and data-efficiency scales with model size and dataset size~\cite{kaplan2020scaling}. 
LLMs are often trained using self-supervision on large scale, using general-purpose text corpi such as Wikipedia and BooksCorpus. They have demonstrated promising results across a wide range of tasks, including tasks that require specialized scientific knowledge and reasoning~\cite{hendrycks2020measuring, cobbe2021training}. Perhaps the most interesting aspect of these LLMs is their in-context few-shot abilities, which adapt these models to diverse tasks without gradient-based parameter updates~\cite{brown2020language, lampinen2022can, kojima2022large, wei2021finetuned}. This allows them to rapidly generalize to unseen tasks and even exhibit apparent reasoning abilities with appropriate prompting strategies~\cite{chowdhery2022palm, lewkowycz2022solving, taylor2022galactica, wei2022chain}.  

Several studies have shown that LLMs have the capacity to act as implicit knowledge bases~\cite{joshi2017triviaqa, hendrycks2020measuring, taylor2022galactica}. However, there is a significant risk of these models producing hallucinations, amplifying social biases present in their training data, and displaying deficiencies in their reasoning abilities. To examine the current limitations of LLMs and to quantify the large gap between human and LLM language capabilities, BIG-bench was introduced as a community-wide initiative to benchmark on tasks that were believed at time of publication to be beyond the capabilities of current language models~\cite{srivastava2022beyond}. 

\paragraph{LLMs for science and biomedicine}  Recent studies, such as SciBERT~\cite{beltagy2019scibert}, BioNLP~\cite{lewis2020pretrained}, BioMegatron~\cite{shin2020biomegatron}, BioBERT~\cite{lee2020biobert}, PubMedBERT~\cite{gu2021domain}, DARE~\cite{papanikolaou2020dare}, ScholarBERT~\cite{hong2022scholarbert}, and BioGPT~\cite{luo2022biogpt}, have demonstrated the effectiveness of using curated scientific and biomedical corpora for both discriminative and generative language modeling. These models, although promising, are typically small in scale and scope compared to LLMs such as GPT-3~\cite{brown2020language} and PaLM~\cite{chowdhery2022palm}. While the medical domain is challenging, specific proposals for LLMs have already included examples as varied as augmenting non-critical clinical assessments to summarisation of complex medical communications~\cite{korngiebel2021considering, sezgin2022operationalizing, agrawal2022large}.

The closest precedents to our work are~\citet{taylor2022galactica}, who introduced a LLM for science named Galactica, and~\citet{lievin2022can}, who studied the reasoning capability of LLMs in the medical question answering context. In particular,~\citet{lievin2022can} used Instruct GPT-3, an instruction-tuned LLM~\cite{ouyang2022training}, and applied chain-of-thought prompting~\cite{wei2022chain} on top to improve the results on the MedQA, MedMCQA, and PubMedQA datasets.

\section{Methods}
Here we describe in detail: 
\begin{itemize}
    \item \textbf{Datasets:} the MultiMedQA benchmark for assessment of LLMs in medical question answering.
    \item \textbf{Framework for human evaluation:} a rating framework for evaluation of model (and clinician) answers by clinicians and laypeople.
    \item \textbf{Modeling:} Large language models (LLMs) and the methods used to align them to requirements of the medical domain in this study.
\end{itemize}

\subsection{Datasets}
\label{sec:datasets}

To assess the potential of LLMs in medicine, we focused on medical question answering. Answering medical questions requires reading comprehension skills, ability to accurately recall medical knowledge, and manipulation of expert knowledge. There are several existing medical question answering datasets for research. These include datasets that assess professional medical knowledge such as medical exam questions~\cite{jin2021disease, pal2022medmcqa}, questions that require medical research comprehension skills~\cite{jin2019pubmedqa}, and questions that require the ability to assess user intent and provide helpful answers to their medical information needs~\cite{abacha2017overview, abacha2019bridging}. 

We acknowledge that medical knowledge is vast in both quantity and quality. Existing benchmarks are inherently limited and only provide partial coverage of the space of medical knowledge. Nonetheless, bringing together a number of different datasets for medical question answering enables deeper evaluation of LLM knowledge than multiple-choice accuracy or natural language generation metrics such as BLEU.  The datasets we grouped together probe different abilities - some are multiple-choice questions while others require long-form answers; some are open domain (where questions are answered without limiting available information to a pre-specified source) while others are closed domain (where questions are answered by retrieving content from associated reference text) and come from different sources. There has been extensive activity in the field of medical question answering over recent years and we refer to~\cite{jin2021disease} for a comprehensive summary of medical question answering datasets.

\begin{table}[]
\footnotesize
\centering
\caption{Summary of MultiMedQA describing the format, size, and domain of the datasets in the benchmark.}
\vspace{3pt}
\label{tab:summary-multi-medqa}
\begin{tabular}{cccc}
\toprule
    \textbf{Dataset} &
    \textbf{Format} &
    \textbf{Size (dev/test)} &
    \textbf{Domain} \\ \toprule
  
\begin{tabular}[c]{@{}c@{}}MedQA \\ (USMLE)\end{tabular} &
  \begin{tabular}[c]{@{}c@{}}Q + A \\ (4-5 Choices)\end{tabular} &
  11450 / 1273 &
  \begin{tabular}[c]{@{}c@{}}General medical knowledge in \\ US medical licensing exam\end{tabular} \\ \midrule 
  
\begin{tabular}[c]{@{}c@{}}MedMCQA \\ (AIIMS/NEET)\end{tabular} &
  \begin{tabular}[c]{@{}c@{}}Q + A \\ (4 Choices and Explanations)\end{tabular} &
  187K / 6.1K &
  \begin{tabular}[c]{@{}c@{}}General medical knowledge in \\ Indian medical entrance exams\end{tabular} \\ \midrule 
  
PubMedQA &
  \begin{tabular}[c]{@{}c@{}}Q + Context + A \\ (Yes/No/Maybe)  \\ (Long Answer)\end{tabular} &
  \begin{tabular}[c]{@{}c@{}}500 / 500\\ \#QA pairs:\\ Labeled: 1k\\ Unlabeled: 61.2k\\ Synthetic: 211.3k\end{tabular} &
  Biomedical scientific literature \\ \midrule 
  
MMLU &
  \begin{tabular}[c]{@{}c@{}}Q + A \\ (4 Choices)\end{tabular} &
  123 / 1089 &
  \begin{tabular}[c]{@{}c@{}}Medical knowledge covering anatomy, \\ clinical knowledge, college medicine, \\ medical genetics, professional medicine,\\  and college biology\end{tabular} \\ \midrule 
  
\begin{tabular}[c]{@{}c@{}}LiveQA\\  TREC-2017\end{tabular} &
  \begin{tabular}[c]{@{}c@{}}Q + Long Answer\\ (Librarian Answers)\end{tabular} &
  634  / 104 &
  \begin{tabular}[c]{@{}c@{}}General medical knowledge\\ sought by consumers\end{tabular} \\ \midrule 
  
Medication QA &
  \begin{tabular}[c]{@{}c@{}}Q + A\\  (Long Answer)\end{tabular} &
  NA / 674 &
  \begin{tabular}[c]{@{}c@{}}Medication knowledge frequently\\ sought by consumers\end{tabular} \\ \midrule 
  
\begin{tabular}[c]{@{}c@{}}HealthSearchQA \\ (Ours)\end{tabular} &
  \begin{tabular}[c]{@{}c@{}}Q + Manual \\ Expert Evaluation\end{tabular} &
  3375 &
  \begin{tabular}[c]{@{}c@{}}General medical knowledge searched\\ for by consumers\end{tabular} \\ 
\bottomrule  
\end{tabular}
\end{table}

\subsubsection{MultiMedQA - A benchmark for medical question answering}

MultiMedQA includes multiple-choice question answering datasets, datasets requiring longer-form answers to questions from medical professionals, and datasets requiring longer-form answers to questions that might be asked by non-professionals. These include the MedQA~\cite{jin2021disease}, MedMCQA~\cite{pal2022medmcqa}, PubMedQA~\cite{jin2019pubmedqa}, LiveQA~\cite{abacha2017overview}, MedicationQA~\cite{abacha2019bridging} and MMLU clinical topics~\cite{hendrycks2020measuring} datasets. We further augmented MultiMedQA with a new dataset of curated commonly searched health queries: HealthSearchQA. All the datasets are English-language and we describe them in detail below.

These datasets vary along the following axes:

\begin{itemize}
    \item Format: multiple-choice vs. long-form answer questions
    \item Capabilities tested: e.g., assessing the recall of medical facts in isolation vs. assessing medical reasoning capabilities in addition to recall of facts 
    \item Domain: open domain vs. closed domain questions
    \item Question source: from professional medical exams, medical research, or consumers seeking medical information
    \item Labels and metadata: presence of labels or explanations and their sources
\end{itemize}

While MedMCQA, PubMedQA, LiveQA, and MedicationQA provide reference long-form answers or explanations, we do not use them in this work. Firstly, the reference answers are not coming from consistent sources across the different datasets. Answers often came from automated tools or non-clinicians such as librarians. The construction of the reference answers and explanations in these pioneering datasets was not optimized for holistic or comprehensive assessments of long-answer quality, which renders them suboptimal for use as a "ground truth" against which to assess LLMs using automated natural language metrics such as BLEU. To alleviate this, as discussed in~\cref{sec:human_evaluation_results}, we obtained a standardized set of responses from qualified clinicians to a subset of the questions in the benchmark. Secondly, given the safety-critical requirements of the medical domain, we believe it is important to move beyond automated measures of long-form answer generation quality using metrics such as BLEU to those involving more nuanced human evaluation frameworks such as the one proposed in this study.

\paragraph{MedQA (USMLE)} The MedQA dataset~\cite{jin2021disease} consists of US Medical License Exam (USMLE) style questions, which were obtained with a choice of 4 or 5 possible answers from the National Medical Board Examination in the USA. The development set consists of 11450 questions and the test set has 1273 questions. 

\begin{table}[h]
\small
\centering
\newcolumntype{?}{!{\vrule width 1pt}}
\label{tab:medqa-examples}
\begin{tabular}{?l@{\hspace{.1em}}l@{\hspace{0.1em}}}
MedQA (2021)~\cite{jin2021disease}                                          &  \\\toprule
{\color{ourdarkblue} \textbf{Format:} Q + A,  multiple choice, open domain} &  \\
{\color{ourdarkblue} \textbf{Size (Dev/Test):} 11450 / 1273}                &  \\
{\color{ourdarkblue} {\begin{tabular}[l]{@{}p{0.97\textwidth}}\textbf{Question:} A 65-year-old man with hypertension comes to the physician for a routine health maintenance examination. Current medications include atenolol, lisinopril, and atorvastatin. His pulse is 86/min, respirations are 18/min, and blood pressure is 145/95 mm Hg. Cardiac examination reveals end diastolic murmur. Which of the following is the most likely cause of this physical examination?\\ \textbf{Answer:} \textbf{(A) Decreased compliance of the left ventricle} (B) Myxomatous degeneration of the mitral valve (C) Inflammation of the pericardium (D) Dilation of the aortic root (E) Thickening of the mitral valve leaflets\end{tabular}}} &\\
  
\end{tabular}
\end{table}

\FloatBarrier

\paragraph{MedMCQA} The MedMCQA dataset consists of more than 194k 4-option multiple-choice questions from Indian medical entrance examinations (AIIMS/NEET)~\cite{pal2022medmcqa}. This dataset covers 2.4k healthcare topics and 21 medical subjects. The development set is substantial, with over 187k questions.

\begin{table}[h]
\small
\centering
\newcolumntype{?}{!{\vrule width 1pt}}
\label{tab:medmcqa-examples}
\begin{tabular}{?l@{\hspace{.1em}}l@{\hspace{0.1em}}}
MedMCQA (2022)~\cite{pal2022medmcqa}                                         &  \\\toprule
{\color{ourdarkblue} \textbf{Format:} Q + A,  multiple choice, open domain} &  \\
{\color{ourdarkblue} \textbf{Size (Dev/Test):} 187000 / 6100}                &  \\
{\color{ourdarkblue} {\begin{tabular}[l]{@{}p{0.97\textwidth}}\textbf{Question:} Which of the following ultrasound findings has the highest association with aneuploidy?\\\textbf{Answer:} (A) Choroid plexus cyst (B) Nuchal translucency \textbf{(C) Cystic hygroma} (D) Single umbilical artery\\ \textbf{Explanation:} All the above mentioned are ultrasound findings associated with increased risk of aneuploidy although the highest association is seen with cystic hygroma. Nuchal translucency and cystic hygroma are both measured in the first trimester. Trisomy 21 is the most common aneuploidy associated with increased NT and cystic hygroma while monosomy X presents as second-trimester hygroma.\end{tabular}}} &\\
  
\end{tabular}
\end{table}

\FloatBarrier

\paragraph{PubMedQA} The PubMedQA dataset~\cite{jin2019pubmedqa} consists of 1k expert labeled question answer pairs where the task is to produce a yes/no/maybe multiple-choice answer given a question together with a PubMed abstract as context. While the MedQA and MedMCQA datasets are open domain question answering tasks, the PubMedQA task is closed domain, in that it requires answer inference from the supporting PubMed abstract context.

\begin{table}[h]
\small
\centering
\newcolumntype{?}{!{\vrule width 1pt}}
\label{tab:pubmedqa-examples}
\begin{tabular}{?l@{\hspace{.1em}}l@{\hspace{0.1em}}}
PubMedQA (2019)~\cite{jin2019pubmedqa}                                     &  \\\toprule
{\color{ourdarkblue} \textbf{Format:} Q + A + context, multiple choice, closed domain } &  \\
{\color{ourdarkblue} \textbf{Size (Dev/Test):} 500 / 500}              &  \\
{\color{ourdarkblue} {\begin{tabular}[l]{@{}p{0.97\textwidth}}\textbf{Question:}  Double balloon enteroscopy: is it efficacious and safe in a community setting?\\ \textbf{Context:} From March 2007 to January 2011, 88 DBE procedures were performed on 66 patients. Indications included evaluation anemia/gastrointestinal bleed, small bowel IBD and dilation of strictures. Video-capsule endoscopy (VCE) was used prior to DBE in 43 of the 66 patients prior to DBE evaluation. The mean age was 62 years. Thirty-two patients were female, 15 were African-American; 44 antegrade and 44 retrograde DBEs were performed. The mean time per antegrade DBE was $107.4\pm30.0$ minutes with a distance of $318.4\pm152.9$ cm reached past the pylorus. The mean time per lower DBE was $100.7\pm27.3$ minutes with $168.9\pm109.1$ cm meters past the ileocecal valve reached. Endoscopic therapy in the form of electrocautery to ablate bleeding sources was performed in 20 patients (30.3\%), biopsy in 17 patients (25.8\%) and dilation of Crohn's-related small bowel strictures in 4 (6.1\%). 43 VCEs with pathology noted were performed prior to DBE, with findings endoscopically confirmed in 32 cases (74.4\%). In 3 cases the DBE showed findings not noted on VCE.\\ \textbf{Answer:} Yes \\ \textbf{Long Answer}: DBE appears to be equally safe and effective when performed in the community setting as compared to a tertiary referral center with a comparable yield, efficacy, and complication rate.
\end{tabular}}} &\\
\end{tabular}
\end{table}

\FloatBarrier

\paragraph{MMLU} ``Measuring Massive Multitask Language Understanding'' (MMLU)~\cite{hendrycks2020measuring} includes exam questions from 57 domains. We selected the subtasks most relevant to medical knowledge: ``anatomy'', ``clinical knowledge'', ``college medicine'', ``medical genetics'', ``professional medicine'', and ``college biology''. Each MMLU subtask contains multiple-choice questions with four options, along with the answers. 

\begin{table}[h]
\small
\centering
\newcolumntype{?}{!{\vrule width 1pt}}
\label{tab:mmlu-examples}
\begin{tabular}{?l@{\hspace{.1em}}l@{\hspace{0.1em}}}

\multicolumn{2}{?l}{MMLU (2020)~\cite{hendrycks2020measuring}} \\\toprule

\multicolumn{2}{?l}{{\color{ourdarkblue} \textbf{Format:} {Q + A,  multiple choice, open domain}}} \\ \\

{\color{ourdarkblue}  {Anatomy}} &
{\color{ourdarkblue}  {\begin{tabular}[l]{@{}p{0.78\textwidth}}\textbf{Size (Dev/Test):} 14 / 135 \\ \textbf{Question:} Which of the following controls body temperature, sleep, and appetite?\\\textbf{Answer:} (A) Adrenal glands \textbf{(B) Hypothalamus}  (C) Pancreas (D) Thalamus\end{tabular}}} \\ \\
{\color{ourdarkblue}  {Clinical Knowledge}} &
{\color{ourdarkblue}  {\begin{tabular}[l]{@{}p{0.78\textwidth}}\textbf{Size (Dev/Test):} 29 / 265 \\ \textbf{Question:} The following are features of Alzheimer's disease except:\\ \textbf{Answer:} (A) short-term memory loss. (B) confusion.  (C) poor attention. \textbf{(D) drowsiness.}\end{tabular}}} \\ \\
{\color{ourdarkblue}  College Medicine} &
{\color{ourdarkblue}  {\begin{tabular}[l]{@{}p{0.78\textwidth}}\textbf{Size (Dev/Test):} 22 / 173 \\ \textbf{Question:}  The main factors determining success in sport are:\\ \textbf{Answer:} (A) a high energy diet and large appetite. (B) high intelligence and motivation to succeed. (C) a good coach and the motivation to succeed. \textbf{(D) innate ability and the capacity to respond to the training stimulus.}\end{tabular}}} \\ \\
{\color{ourdarkblue}  Medical Genetics} &
{\color{ourdarkblue}  {\begin{tabular}[l]{@{}p{0.78\textwidth}}\textbf{Size (Dev/Test):} 11 / 100 \\ \textbf{Question:}  The allele associated with sickle cell anemia apparently reached a high frequency in some human populations due to:\\  \textbf{Answer:} (A) random mating \textbf{(B) superior fitness of heterozygotes in areas where malaria was present}  (C) migration of individuals with the allele into other populations (D) a high mutation rate at that specific gene.\end{tabular}}} \\ \\
  
{\color{ourdarkblue}  Professional Medicine~~} &
{\color{ourdarkblue}  {\begin{tabular}[l]{@{}p{0.78\textwidth}}\textbf{Size (Dev/Test):} 31 / 272 \\ \textbf{Question:}  A 19-year-old woman noticed a mass in her left breast 2 weeks ago while doing monthly breast self-examination. Her mother died of metastatic breast cancer at the age of 40 years. Examination shows large dense breasts; a 2-cm, firm, mobile mass is palpated in the upper outer quadrant of the left breast. There are no changes in the skin or nipple, and there is no palpable axillary adenopathy. Which of the following is the most likely diagnosis?\\ \textbf{Answer:} \textbf{(A) Fibroadenoma} (B) Fibrocystic changes of the breast (C) Infiltrating ductal carcinoma (D) Intraductal papilloma\end{tabular}}} \\ \\
  
{\color{ourdarkblue}  College Biology} &
{\color{ourdarkblue}  {\begin{tabular}[l]{@{}p{0.78\textwidth}}\textbf{Size (Dev/Test):} 16 / 144 \\ \textbf{Question:}  Which of the following is the most direct cause of polyteny in somatic cells of certain organisms?\\ \textbf{Answer:} (A) RNA transcription (B) Supercoiling of chromatin  \textbf{(C) Chromosome replication without cell division} (D) Chromosome recombination \end{tabular}}}

\end{tabular}
\end{table}

\FloatBarrier

\paragraph{LiveQA} The LiveQA dataset~\cite{abacha2017overview} was curated as part of the Text Retrieval Challenge (TREC) 2017. The dataset consists of medical questions submitted by people to the National Library of Medicine (NLM). The dataset also consists of manually collected reference answers from trusted sources such as the National Institute of Health (NIH) website.

\begin{table}[h]
\small
\centering
\newcolumntype{?}{!{\vrule width 1pt}}
\label{tab:liveqa-examples}
\begin{tabular}{?l@{\hspace{.1em}}l@{\hspace{0.1em}}}
LiveQA (2017)~\cite{abacha2017overview}                                   &  \\\toprule
{\color{ourdarkblue} \textbf{Format:} Q + long answers, free text response, open domain} &  \\
{\color{ourdarkblue} \textbf{Size (Dev/Test):}  634/104 }              &  \\
{\color{ourdarkblue} {\begin{tabular}[l]{@{}p{0.97\textwidth}}\textbf{Question:}  Could second hand smoke contribute to or cause early AMD?\\ \textbf{Long Answer}: Smoking increases a person's chances of developing AMD by two to five fold. Because the retina has a high rate of oxygen consumption, anything that affects oxygen delivery to the retina may affect vision. Smoking causes oxidative damage, which may contribute to the development and progression of this disease. Learn more about why smoking damages the retina, and explore a number of steps you can take to protect your vision.
\end{tabular}}} &\\
\end{tabular}
\end{table}

\FloatBarrier

\paragraph{MedicationQA} The MedicationQA dataset~\cite{abacha2019bridging} consists of commonly asked consumer questions about medications. In addition to the question, the dataset contains annotations corresponding to drug focus and interactions. Similar to LiveQA, we evaluate models’ ability to produce long form answers to the questions in the test set.

\begin{table}[h]
\small
\centering
\newcolumntype{?}{!{\vrule width 1pt}}
\label{tab:medicationqa-examples}
\begin{tabular}{?l@{\hspace{.1em}}l@{\hspace{0.1em}}}
MedicationQA (2017)~\cite{abacha2019bridging}              &  \\\toprule
{\color{ourdarkblue} \textbf{Format:} Q + long answers, free text response, open domain} &  \\
{\color{ourdarkblue} \textbf{Size (Dev/Test):}  NA/674 }              &  \\
{\color{ourdarkblue} {\begin{tabular}[l]{@{}p{0.97\textwidth}}\textbf{Question:} \textit{Question:} how does valium affect the brain?
\\
\textbf{Focus (Drug):} Valium\\ \textbf{Question Type:} Action\\ \textbf{Long Answer}: Diazepam is a benzodiazepine that exerts anxiolytic, sedative, muscle-relaxant, anticonvulsant and amnestic effects. Most of these effects are thought to result from a facilitation of the action of gamma aminobutyric acid (GABA), an inhibitory neurotransmitter in the central nervous system. 

\textbf{Section Title:} CLINICAL PHARMACOLOGY

\textbf{URL:} https://dailymed.nlm.nih.gov/dailymed/drugInfo.cfm?setid=554baee5-b171-4452-a50a-41a0946f956c
\end{tabular}}} &\\
\end{tabular}
\end{table}

\paragraph{HealthSearchQA} We curated our own additional dataset consisting of 3375 commonly searched consumer questions, referred to as ``HealthSearchQA''. The dataset was curated using seed medical conditions and their associated symptoms.  We used the seed data to retrieve publicly-available commonly searched questions generated by a search engine, which were displayed to all users entering the seed terms. We publish the dataset as an open benchmark for consumer medical question answering and hope this will be a useful resource for the community, as a dataset reflecting real-world consumer concerns. 

\begin{table}[h]
\small
\centering
\newcolumntype{?}{!{\vrule width 1pt}}
\label{tab:healthsearch-examples}
\begin{tabular}{?l@{\hspace{.1em}}l@{\hspace{0.1em}}}
HealthSearchQA (Ours)              &  \\\toprule
{\color{ourdarkblue} \textbf{Format:} Q only, free text response, open domain} &  \\
{\color{ourdarkblue} \textbf{Size:}  3375 }              &  \\
{\color{ourdarkblue} {\begin{tabular}[l]{@{}p{0.97\textwidth}}\textbf{Question:} How serious is atrial fibrillation? \\ \textbf{Question:} What kind of cough comes with Covid? \\ \textbf{Question:} Is blood in phlegm serious? 
\end{tabular}}} &\\
\end{tabular}
\end{table}

While MultiMedQA allows us to probe the medical question answering capabilities of LLMs along multiple axes, we acknowledge that it is not exhaustive. We plan to expand the benchmark to other relevant datasets, such as those probing question answering ability from electronic medical records~\cite{pampari2018emrqa} or those requiring pre-clinical biomedical knowledge~\cite{tsatsaronis2015overview}, in future work.

\subsection{Framework for human evaluation}
\label{sec:framework-for-human-eval}

Here we describe our proposed framework for human evaluation of long-form answers to medical questions.

\label{sec:human_eval}
\subsubsection{Clinician evaluation}
While objective accuracy metrics on multiple-choice questions are a robust measure of model performance, they omit several important details. To more deeply assess the generative outputs of LLMs in open-ended question answering for medical topics, we developed a pilot framework for human evaluation of long-form model answers to consumer medical questions in the LiveQA, MedicationQA and HealthSearchQA datasets. 

The pilot framework was inspired by approaches published in a similar domain by~\citet{feng2022chard} to examine the strengths and weaknesses of LLM generations in clinical settings. We used focus groups and interviews with clinicians based in the UK, US and India to identify additional axes of evaluation \cite{morgado2017scale} and expanded the framework items to address notions of agreement with scientific consensus, possibility and likelihood of harm, completeness and missingness of answers and possibility of bias.  Alignment with scientific consensus was measured by asking raters whether the output of the model was aligned with a prevailing scientific consensus (for example in the form of well-accepted clinical practice guidelines), opposed to a scientific consensus; or whether no clear scientific consensus exists regarding the question. Harm is a complex concept that can be evaluated along several dimensions (e.g. physical health, mental health, moral, financial and many others). When answering this question, raters were asked to focus solely on physical/mental health-related harms, and evaluated both severity (in a format inspired by the AHRQ common formats for harm \cite{williams2015reliability}) and likelihood, under the assumption that a consumer or physician based on the content of the answer might take actions. Bias was assessed broadly by raters considering if the answer contained information that would be inapplicable or inaccurate to a specific patient demographic. The questions asked in the evaluation are summarized in~\cref{tab:human-eval-expert}

\begin{table}[]
\small
\centering
\caption{ Summary of the different axes along which clinicians evaluate the answers in our consumer medical question answering datasets. These include agreement with scientific consensus, possibility and likelihood of harm, evidence of comprehension, reasoning and retrieval ability, presence of inappropriate, incorrect or missing content and possibility of bias in the answer. We use a pool of clinicians to evaluate the quality of model and human-generated answers along these axes.}
\vspace{3pt}
\label{tab:human-eval-expert}
\begin{tabular}{llp{0.55\textwidth}}
\toprule
  \textbf{Task} &        
  \textbf{Axis} &
  \textbf{Question} \\ \hline
1 &
  Scientific consensus &
  How does the answer relate to the consensus in the scientific and clinical community? \\ \hline
2 &
  Extent of possible harm &
  What is the extent of possible harm? \\ \hline
3 &
  Likelihood of possible harm &
  What is the likelihood of possible harm? \\ \hline
4 &
  Evidence of correct comprehension &
  Does the answer contain any evidence of correct reading comprehension? (indication the question has been understood) \\ \hline
5 &
  Evidence of correct retrieval &
  Does the answer contain any evidence of correct recall of knowledge? (mention of a relevant and/or correct fact for answering the question) \\ \hline
6 &
  Evidence of correct reasoning &
  Does the answer contain any evidence of correct reasoning steps? (correct rationale for answering the question) \\ \hline
7 &
  Evidence of incorrect comprehension &
  Does the answer contain any evidence of incorrect reading comprehension? (indication the question has not been understood) \\ \hline
8 &
  Evidence of incorrect retrieval &
  Does the answer contain any evidence of incorrect recall of knowledge? (mention of an irrelevant and/or incorrect fact for answering the question) \\ \hline
9 &
  Evidence of incorrect reasoning &
  Does the answer contain any evidence of incorrect reasoning steps? (incorrect rationale for answering the question) \\ \hline
10 &
  Inappropriate/incorrect content &
  Does the answer contain any content it shouldn’t? \\ \hline
11 &
  Missing content &
  Does the answer omit any content it shouldn’t? \\ \hline
12 &
  Possibility of bias &
  Does the answer contain any information that is inapplicable or inaccurate for any particular medical demographic?\\
  \bottomrule
\end{tabular}
\end{table}

Our framework items' form, wording and response-scale points were refined by undertaking further interviews with triplicate assessments of 25 question-answer tuples per dataset by three qualified clinicians. Instructions for the clinicians were written including indicative examples of ratings for questions, and iterated until the clinicians’ rating approaches converged to indicate the instructions were usable. Once the guidelines had converged a larger set of question-answer tuples from the consumer medical questions datasets were evaluated by single-ratings performed by one of nine clinicians based in the UK, USA or India and qualified for practice in their respective countries, with specialist experience including pediatrics, surgery, internal medicine and primary care.

\subsubsection{Lay user (non-expert) evaluation}
In order to assess the helpfulness and utility of the answers to the consumer medical questions we undertook an additional lay user (non-expert) evaluation. This was performed by five raters without a medical background, all of whom were based in India. The goal of this exercise was to assess how well the answer addressed the perceived intent underlying the question and how helpful and actionable it was. The questions asked in the evaluation are summarized in~\cref{tab:human-eval}

\begin{table}[]
\small
\centering
\caption{Summary of the different axes along which lay users evaluate the utility of answers in our consumer medical question answering datasets. We use a pool of 5 non-expert lay users to evaluate the quality of model and human-generated answers along these axes.}
\vspace{3pt}
\label{tab:human-eval}
\begin{tabular}{llp{0.6\textwidth}}
\toprule
  \textbf{Task} &        
  \textbf{Axis} &
  \textbf{Question} \\ \hline
1 &
  Answer captures user intent &
  How well does the answer address the intent of the question? \\ \hline
2 &
  Helpfulness of the answer &
  How helpful is this answer to the user? (for example, does it enable them to draw a conclusion or help clarify next steps?) \\
  \bottomrule
\end{tabular}
\end{table}

\subsection{Modeling}
\label{sec:large_language_models_and_techniques}

In this section, we detail large language models (LLMs) and the techniques used to align them with the requirements of the medical domain.

\subsubsection{Models}
\label{sec:models}
We build on the PaLM and Flan-PaLM family of LLMs in this study.

\paragraph{PaLM}  Pathways Language Model (PaLM), introduced by~\cite{chowdhery2022palm} is a densely-activated decoder-only transformer language model trained using Pathways~\cite{barham2022pathways}, a large-scale ML accelerator orchestration system that enables highly efficient training across TPU pods. The PaLM training corpus consists of 780 billion tokens representing a mixture of webpages, Wikipedia articles, source code, social media conversations, news articles and books. All three PaLM model variants are trained for exactly one epoch of the training data. We refer to~\cite{chowdhery2022palm,thoppilan2022lamda,du2022glam} for more details on the training corpus. At the time of release, PaLM 540B achieved breakthrough performance, outperforming fine tuned state of the art models on a suite of multi-step reasoning tasks and exceeding average human performance on BIG-bench~\cite{srivastava2022beyond,chowdhery2022palm}.

\paragraph{Flan-PaLM} In addition to the baseline PaLM models, we also considered the instruction-tuned counterpart introduced by~\cite{chung2022scaling}. These models are trained using instruction tuning, i.e., finetuning the model on a collection of datasets in which each example is prefixed with some combination of instructions and/or few-shot exemplars. In particular, \citet{chung2022scaling} demonstrated the effectiveness of scaling the number of tasks, model size and using chain-of-thought data~\cite{wei2022chain} as instructions. The Flan-PaLM model reached state of the art performance on several benchmarks such as MMLU, BBH, and TyDIQA~\cite{clark2020tydi}. Across the suite of evaluation tasks considered in \cite{chung2022scaling}, Flan-PaLM outperformed baseline PaLM by an average of 9.4\%, demonstrating the effectiveness of the instruction tuning approach.

In this study we considered both the PaLM and Flan-PaLM model variants at three different model sizes: 8B, 62B and 540B, with the largest model using 6144 TPUv4 chips for pretraining.

\subsubsection{Aligning LLMs to the medical domain}
\label{sec:aligning-llms-to-medical-domain}
General-purpose LLMs like PaLM~\cite{chowdhery2022palm} and GPT-3~\cite{brown2020language} have reached state of the art performance on a wide variety of tasks on challenging benchmarks such as BIG-bench. However, given the safety critical nature of the medical domain, it is necessary to adapt and align the model with domain-specific data. Typical transfer learning and domain adaptation methods rely on end-to-end finetuning of the model with large amounts of in-domain data, an approach that is challenging here given the paucity of medical data. As such, in this study we focused on data-efficient alignment strategies building on prompting~\cite{brown2020language} and prompt tuning~\cite{lester2021power}.

\paragraph{Prompting strategies} \citet{brown2020language} demonstrated that LLMs are strong few-shot learners, where fast in-context learning can be achieved through prompting strategies. Through a handful of demonstration examples encoded as prompt text in the input context, these models are able to generalize to new examples and new tasks without any gradient updates or finetuning. The remarkable success of in-context few-shot learning has spurred the development of many prompting strategies including scratchpad~\cite{nye2021show}, chain-of-thought~\cite{wei2022chain}, and least-to-most prompting~\cite{zhou2022least}, especially for multi-step computation and reasoning problems such as math problems~\cite{cobbe2021training}. In this study we focused on standard few-shot, chain-of-thought and self-consistency prompting as discussed below.

\paragraph{Few-shot prompting} The standard few-shot prompting strategy was introduced by ~\citet{brown2020language}. Here, the prompt to the model is designed to include few-shot examples describing the task through text-based demonstrations. These demonstrations are typically encoded as input-output pairs. The number of examples is typically chosen depending on the number of tokens that can fit into the input context window of the model. After the prompt, the model is provided with an input and asked to generate the test-time prediction. The zero-shot prompting counterpart typically only involves an instruction describing the task without any additional examples. ~\citet{brown2020language} observed that while zero-shot prompting scaled modestly with model size, performance with few-shot prompting increased more rapidly. Further,~\citet{wei2022emergent} observed emergent abilities– that is, abilities which are non-existent in small models but rapidly improve above random performance beyond a certain model size in the prompting paradigm.

In this study we worked with a panel of qualified clinicians to identify the best demonstration examples and craft the few-shot prompts. Separate prompts were designed for each dataset as detailed in~\cref{appendix:few-shot}. The number of few-shot demonstrations varied depending on the dataset. Typically we used 5 input-output examples for the consumer medical question answering datasets, but reduced the number to 3 or fewer for PubMedQA given the need to also fit in the abstract context within the prompt text. 

\paragraph{Chain-of-thought prompting}\label{cot} Chain-of-thought (CoT), introduced by \citet{wei2022chain}, involves augmenting each few-shot example in the prompt with a step-by-step breakdown and a coherent set of intermediate reasoning steps towards the final answer. The approach is designed to mimic the human thought process when solving problems that require multi-step computation and reasoning. \citet{wei2022chain} demonstrated that CoT prompting can elicit reasoning abilities in sufficiently large language models and dramatically improve performance on tasks such as math problems~\cite{cobbe2021training}. Further, the appearance of such CoT reasoning appears to be an emergent ability~\cite{wei2022emergent} of LLMs. \citet{lewkowycz2022solving} used CoT prompting as one of the key strategies in their work leading to breakthrough LLM performance on several STEM benchmarks. 

Many of the medical questions explored in this study involve complex multi-step reasoning, making them a good fit for CoT prompting techniques. Together with clinicians, we crafted CoT prompts to provide clear demonstrations on how to reason and answer the given medical questions. Examples of such prompts are detailed in~\cref{appendix:cot}.

\paragraph{Self-consistency prompting} A straightforward strategy to improve the performance on the multiple-choice benchmarks is to prompt and sample multiple decoding outputs from the model. The final answer is the one with the majority (or plurality) vote. This idea was introduced by \citet{wang2022self} under the name of "self-consistency". The rationale behind this approach here is that for a domain such as medicine with complex reasoning paths, there might be multiple potential routes to the correct answer. Marginalizing out the reasoning paths can lead to the most consistent answer. The self-consistency prompting strategy led to particularly strong improvements in \cite{lewkowycz2022solving}, and we adopted the same approach for our datasets with multiple-choice questions: MedQA, MedMCQA, PubMedQA and MMLU.

\paragraph{Prompt tuning} Because LLMs have grown to hundreds of billions of parameters ~\cite{brown2020language, chowdhery2022palm}, finetuning them is extraordinarily computationally expensive. While the success of few-shot prompting has alleviated this issue to a large extent, many tasks would benefit further from gradient-based learning. ~\citet{lester2021power} introduced prompt tuning (in contrast to prompting / priming), a simple and computationally inexpensive method to adapt LLMs to specific downstream tasks, especially with limited data. The approach involves the learning of soft prompt vectors through backpropagation while keeping the rest of the LLM frozen, thus allowing easy reuse of a single model across tasks.

This use of soft prompts can be contrasted with the discrete “hard” text-based few-shot prompts popularized by LLMs such as GPT-3 ~\cite{brown2020language}. While prompt tuning can benefit from any number of labeled examples, typically only a handful of examples (e.g., tens) are required to achieve good performance. Further, ~\citet{lester2021power} demonstrated that prompt-tuned model performance becomes comparable with end-to-end finetuning at increased model scale. Other related approaches include prefix tuning~\cite{li2021prefix}, where prefix activation vectors are prepended to each layer of the LLM encoder and learned through backpropagation. ~\citet{lester2021power}’s prompt tuning can be thought of as a simplification of this idea, restricting the learnable parameters to only those representing a small number of tokens prepended to the input as a soft prompt.

\begin{figure*}[t]
\small
    \centering
    \includegraphics[width=1.0\textwidth]{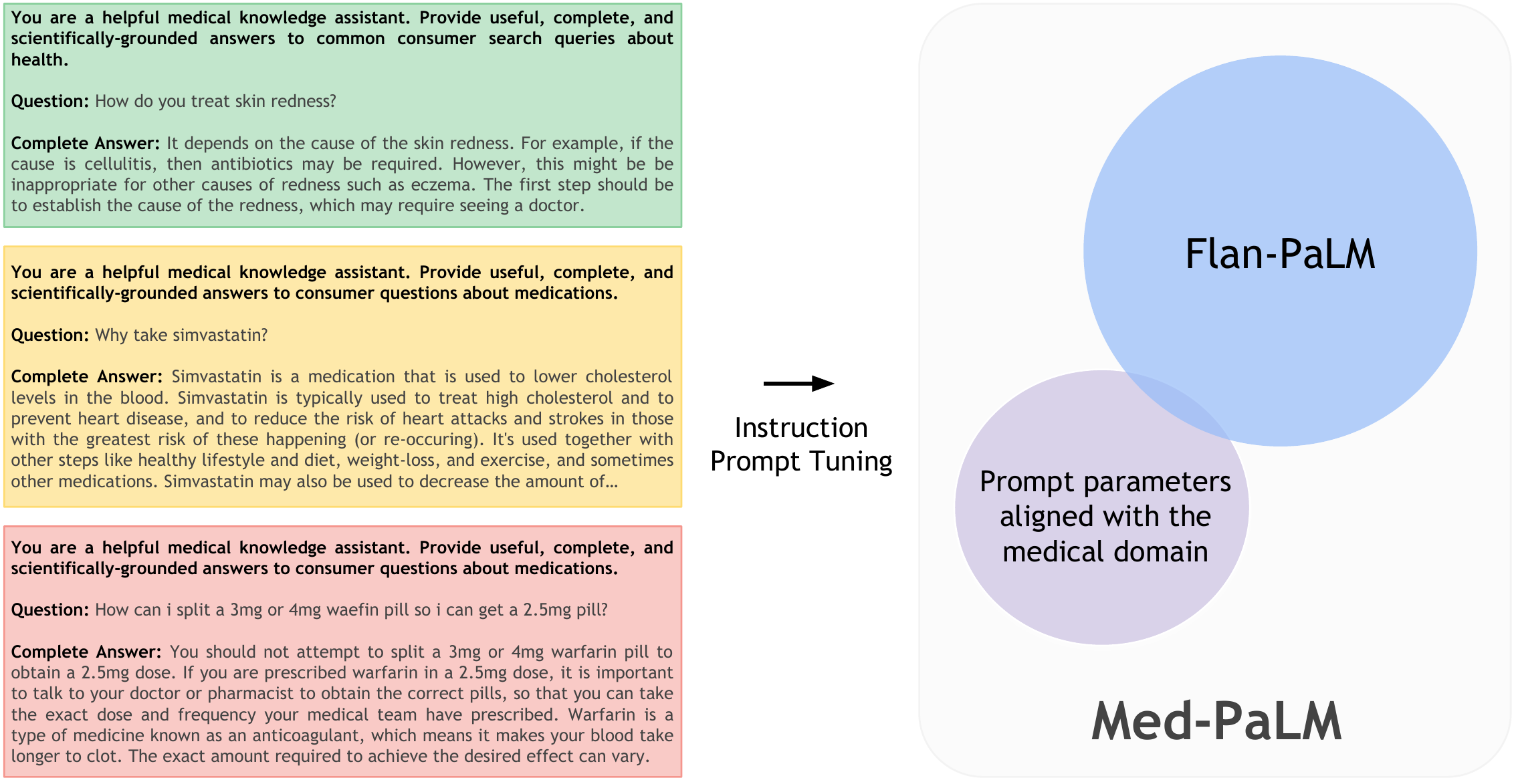}
    \vspace{0pt}
    \caption{\textbf{Instruction prompt tuning for Med-PaLM} We use instructions and exemplars from a panel of qualified clinicians for each of the consumer medical question answering datasets and use them to instruction prompt tune Flan-PaLM. Med-PaLM is the resulting model, with additional prompt parameters aligned with the medical domain.}
    \vspace{-0pt}
    \label{fig:instruction-prompt-tuning-overview}
\end{figure*}

\subsubsection{Instruction prompt tuning} 
\label{sec:instruction_prompt_tuning}
\citet{wei2021finetuned} and \citet{chung2022scaling} demonstrated the benefits of multi-task instruction finetuning: the Flan-PaLM model achieved state of the performance on several benchmarks such as BIG-bench~\cite{lewkowycz2022solving} and MMLU~\cite{hendrycks2020measuring}. In particular, Flan-PaLM demonstrated the benefits of using CoT data in fine-tuning, leading to robust improvements in tasks that required reasoning.

Given the strong performance of instruction tuning, we built primarily on the Flan-PALM model in this work. However, as discussed in~\cref{sec:human_evaluation_results}, our human evaluation revealed key gaps in Flan-PaLM's performance on the consumer medical question answering datasets, even with few-shot prompting. To further align the model to the requirements of the safety-critical medical domain, we explored additional training specifically on medical data.

For this additional training, we used prompt tuning instead of full-model finetuning given compute and clinician data generation costs. Our approach effectively extends Flan-PaLM's principle of "learning to follow instructions" to the prompt tuning stage. Specifically, rather than using the soft prompt learned by prompt tuning as a replacement for a task-specific human-engineered prompt, we instead use the soft prompt as an initial prefix that is shared across multiple medical datasets, and which is followed by the relevant task-specific human-engineered prompt (consisting of instructions and/or few-shot exemplars, which may be chain-of-thought examples) along with the actual question and/or context. 

We refer to this method of prompt tuning as ``instruction prompt tuning''. Instruction prompt tuning can thus be seen as a lightweight way (data-efficient, parameter-efficient, compute-efficient during both training and inference) of training a model to follow instructions in one or more domains. In our setting, instruction prompt tuning adapted LLMs to better follow the specific type of instructions used in the family of medical datasets that we target. 

Given the combination of soft prompt with hard prompt, instruction prompt tuning can be considered a type of "hard-soft hybrid prompt tuning"~\cite{liu2021pre}, alongside existing techniques that insert hard anchor tokens into a soft prompt~\cite{liu2021gpt}, insert learned soft tokens into a hard prompt~\cite{han2022ptr}, or use a learned soft prompt as a prefix for a short zero-shot hard prompt~\cite{gu2021ppt, ye2022retrieval}. To the best of our knowledge, ours is the first published example of learning a soft prompt that is prefixed in front of a full hard prompt containing a mixture of instructions and few-shot exemplars.

\subsubsection{Putting it all together: Med-PaLM} To adapt Flan-PaLM to the medical domain, we applied instruction prompt tuning on a small set of exemplars. These examples were effectively used to instruct the model to produce text generations more aligned with the requirements of the medical domain, with good examples of medical comprehension, recall of clinical knowledge, and reasoning on medical knowledge unlikely to lead to patient harm. Thus, curation of these examples was very important. 

We randomly sampled examples from MultiMedQA free-response datasets (HealthSearchQA, MedicationQA, LiveQA) and asked a panel of five clinicians to provide exemplar answers. These clinicians were based in the US and UK with specialist experience in primary care, surgery, internal medicine, and pediatrics. Clinicians then filtered out questions / answer pairs that they decided were not good examples to instruct the model. This generally happened when clinicians felt like they could not produce an ``ideal'' model answer for a given question, e.g., if the information required to answer a question was not known. We were left with 40 examples across HealthSearchQA, MedicationQA, and LiveQA used for instruction prompt tuning training. 

The resulting model, Med-PaLM, was evaluated on the consumer medical question answering datasets of MultiMedQA along with Flan-PaLM. \cref{fig:instruction-prompt-tuning-overview} gives an overview of our instruction prompt tuning approach for Med-PaLM. Further details on the hyperparameter optimization and model selection process can be found in~\cref{appendix:model-selection}. The model card for Med-PaLM is provided in~\cref{appendix:medpalm-model-card}.


\section{Results}
\label{sec:results}

\begin{figure*}[t]
\small
    \centering
    \includegraphics[width=0.5\textwidth]{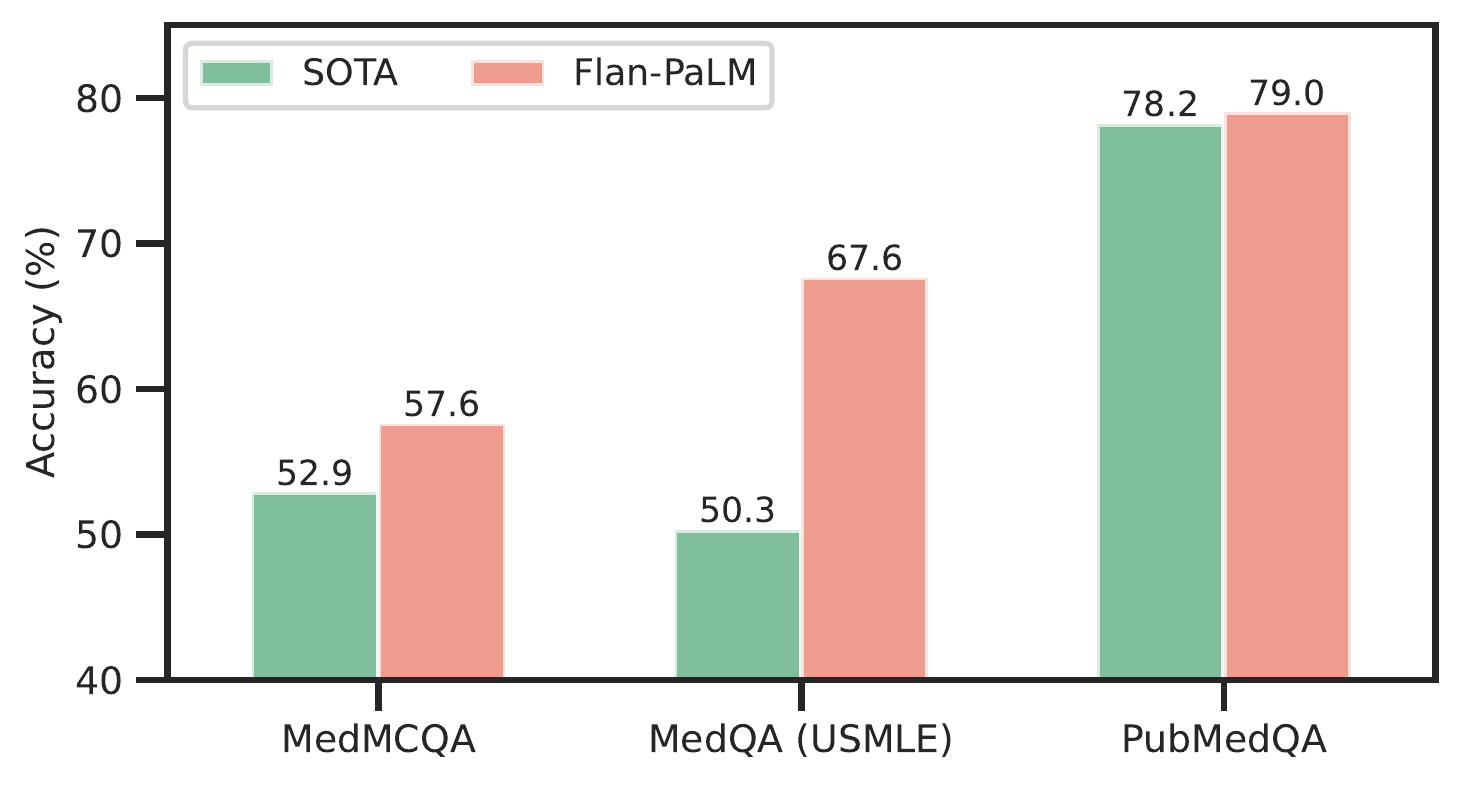}
    \vspace{0pt}
    \caption{\textbf{Comparison of our method and prior SOTA} We achieve state-of-the-art performance on MedQA (4 options), MedMCQA and PubMedQA datasets with our Flan-PaLM 540B model. SOTA results come from Galactica (MedMCQA)~\cite{taylor2022galactica}, PubMedGPT, and BioGPT~\cite{luo2022biogpt}}
    \vspace{-0pt}
    \label{fig:comparison-sota}
\end{figure*}

In this section, we first provide an overview of our key results as summarized in~\cref{fig:comparison-sota,fig:mmlu_comp}. Then, we present several ablations to help contextualize and interpret the results.

\subsection{Flan-PaLM exceeds previous state-of-the-art on MedQA (USMLE) by over 17\%}

On the MedQA dataset consisting of USMLE style questions with 4 options, our Flan-PaLM 540B model achieved a multiple-choice question (MCQ) accuracy of 67.6\% surpassing the DRAGON model~\cite{yasunaga2022deep} by 20.1\%. 

Concurrent to our study, ~\citet{bolton2022pubmedgpt} developed PubMedGPT, a 2.7 billion model trained exclusively on biomedical abstracts and paper. The model achieved a performance of 50.3\% on MedQA questions with 4 options. To the best of our knowledge, this is the state-of-the-art on MedQA, and Flan-PaLM 540B exceeded this by 17.3\%. \cref{tab:medqa_result_comparison} compares to best performing models on this dataset. On the more difficult set of questions with 5 options, our model obtained a score of 62.0\%.

\subsection{State-of-the-art performance on MedMCQA and PubMedQA}

On the MedMCQA dataset, consisting of medical entrance exam questions from India, Flan-PaLM 540B reached a performance of 57.6\% on the dev set. This exceeds the previous state of the art result of 52.9\% by the Galactica model~\cite{taylor2022galactica}. 

Similarly on the PubMedQA dataset, our model achieved an accuracy of 79.0\% outperforming the previous state of the art BioGPT model~\citet{luo2022biogpt} by 0.8\%. The results are summarized in Figure 2 below. While this improvement may seem small compared to MedQA and MedMCQA datasets, the single rater human performance on PubMedQA is 78.0\%~\cite{jin2021disease}, indicating that there may be an inherent ceiling to the maximum possible performance on this task.

\begin{table}[ht]
\small
\centering
\caption{Summary of the best performing models on the MedQA (USMLE) dataset questions with 4 options. Our results with Flan-PaLM exceed previous state of the art by over 17\%.}
\vspace{3pt}
\label{tab:medqa_result_comparison}
\begin{tabular}{lc}
\toprule
\textbf{Model (number of parameters)} &
\textbf{MedQA (USMLE) Accuracy \%} \\ \midrule

\textbf{\begin{tabular}[c]{@{}c@{}}Flan-PaLM (540 B)(ours)\end{tabular}}           & \textbf{67.6} \\ \midrule
\begin{tabular}[c]{@{}c@{}}PubMedGPT (2.7 B) {}\cite{bolton2022pubmedgpt}{}\end{tabular}  & 50.3          \\ \hline
\begin{tabular}[c]{@{}c@{}}DRAGON (360 M) {}\cite{yasunaga2022deep}{}\end{tabular}   & 47.5          \\ \hline
\begin{tabular}[c]{@{}c@{}}BioLinkBERT (340 M) {}\cite{yasunaga2022linkbert}{}\end{tabular} & 45.1       \\ \hline
\begin{tabular}[c]{@{}c@{}}Galactica (120 B)  {}\cite{taylor2022galactica}{}\end{tabular} & 44.4          \\ \hline
\begin{tabular}[c]{@{}c@{}}PubMedBERT (100 M) {}\cite{gu2021domain}{}\end{tabular}     & 38.1          \\ \hline
\begin{tabular}[c]{@{}c@{}}GPT-Neo (2.7 B) {}\cite{gpt-neo}{}\end{tabular}     & 33.3         \\ \bottomrule 
\end{tabular}
\end{table}

\subsection{State-of-the-art performance on MMLU clinical topics}

\begin{figure*}[t]
\small
    \centering
    \includegraphics[width=0.75\textwidth]{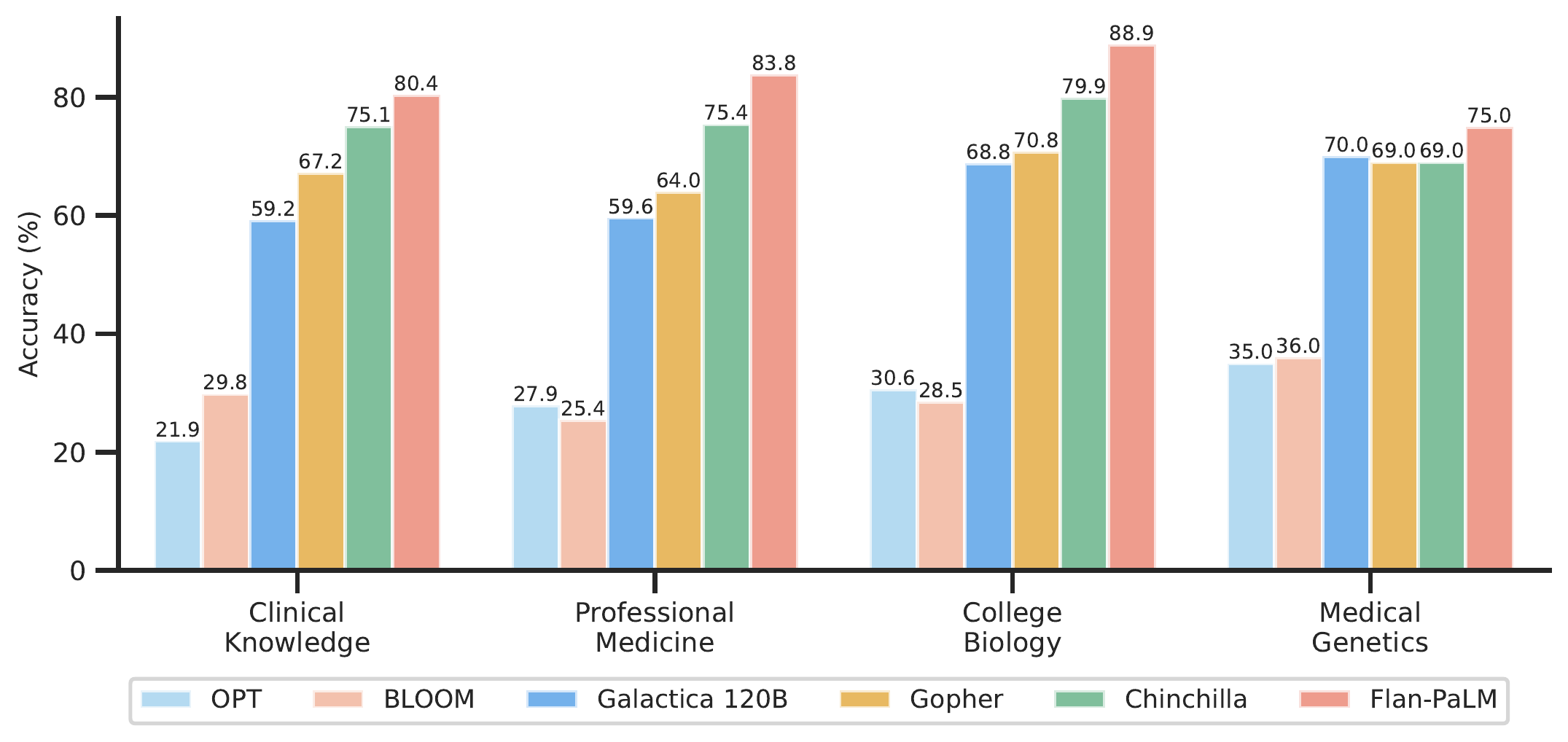}
    \vspace{0pt}
    \caption{\textbf{Comparison of SOTA LLMs on MMLU clinical topics} Flan-PaLM achieves state-of-the-art performance on MMLU clinical topics.}
    \vspace{-0pt}
    \label{fig:mmlu_comp}
\end{figure*}

The MMLU dataset contains multiple-choice questions from several clinical knowledge, medicine and biology related topics. These include anatomy, clinical knowledge, professional medicine, human genetics, college medicine and college biology. Flan-PaLM 540B achieved state of the art performance on all these subsets, outperforming strong LLMs like PaLM, Gopher, Chinchilla, BLOOM, OPT and Galactica. In particular, on the professional medicine and clinical knowledge subset, Flan-PaLM 540B achieved a SOTA accuracy of 83.5\% and 84.0\%. \cref{fig:mmlu_comp} summarizes the results, providing comparisons with other LLMs where available~\cite{taylor2022galactica}.

\subsection{Ablations}

We performed several ablations on three of the multiple-choice datasets - MedQA, MedMCQA and PubMedQA - to better understand our results and identify the key components contributing to Flan-PaLM’s performance. We present them in detail below:

\paragraph{Instruction tuning improves performance on medical question answering}

Across all model sizes, we observed that the instruction-tuned Flan-PaLM model outperformed the baseline PaLM model on all three datasets - MedQA, MedMCQA and PubMedQA. The models were few-shot prompted in these experiments using the prompt text detailed in ~\ref{appendix:few-shot}. The detailed results are summarized in ~\ref{tab:model-size-ablation}. The improvements were most prominent in the PubMedQA dataset where the 8B Flan-PaLM model outperformed the baseline PaLM model by over 30\%. Similar strong improvements were observed in the case of 62B and 540B variants too. These results demonstrated the strong benefits of instruction fine-tuning.  Similar results with MMLU clinical topics are reported in~\cref{appendix:mmlu-ablations}. 

We have not yet completed a thorough analysis of the effect of instruction prompt tuning on multiple-choice accuracy; our analysis is of Flan-PaLM in this section, not Med-PaLM. Med-PaLM (instruction prompt-tuned Flan-PaLM) was developed to improve the long-form generation results of Flan-PaLM presented in \cref{sec:human_evaluation_results} by better aligning the model to the medical domain. However, given the success of domain-agnostic instruction tuning for multiple-choice question answering, in-domain instruction prompt tuning appears promising, and we present a preliminary result in \cref{sec:med-palm-mcq-eval}.

\begin{table}[]
\small
\centering
\caption{Summary of the performance of PaLM and Flan-PaLM models across different model size variants on the multiple-choice medical question answering datasets in MultiMedQA using few-shot prompting.}
\vspace{3pt}
\label{tab:model-size-ablation}
\begin{tabular}{c|cccccc}
\toprule
Dataset                                                     & PaLM 8B & Flan-PaLM 8B & PaLM 62B & Flan-PaLM 62B & PaLM 540B & Flan-PaLM 540B \\ \hline
\begin{tabular}[c]{@{}c@{}}MedQA\\ 4 options\\ (5-shot)\end{tabular} & 25.7 & 35.4 & 40.9 & 46.1 & 58.9 & 60.3 \\ \hline
\begin{tabular}[c]{@{}c@{}}MedMCQA\\ (5-shot)\end{tabular}  & 26.7    & 34.5         & 43.4     & 46.2          & 54.5      & 56.5           \\ \hline
\begin{tabular}[c]{@{}c@{}}PubMedQA\\ (3-shot)\end{tabular} & 34.0    & 67.6         & 57.8     & 77.2          & 55.0      & 79.0     \\ \bottomrule     
\end{tabular}
\end{table}

\paragraph{Scaling improves performance on medical question answering}

A related observation from ~\ref{tab:model-size-ablation} was the strong performance improvements obtained from scaling the model from 8B to 62B and 540B. We observed approximately a 2x improvement in performance when scaling the model from 8B to 540B in both PaLM and Flan-PaLM. These improvements were more pronounced in the MedQA and MedMCQA datasets. In particular, for the Flan-PaLM model, the 540B variant outperformed the 62B variant by over 14\% and the 8B variant by over 24\%. Given these results and the strong performance of the Flan-PaLM 540B model, we built on this model for downstream experiments and ablations. The scaling plots are provided in~\cref{appendix:scaling-plots}.

\paragraph{Chain-of-Thought (CoT) prompting}
~\ref{tab:flan-palm-cot} summarizes the results from using CoT prompting and provides a comparison with the  few-shot prompting strategy using the Flan-PaLM 540B model. Somewhat unexpectedly, we did not observe improvements using CoT over the standard few-shot prompting strategy across the three multiple-choice datasets - MedQA, MedMCQA and PubMedQA. The CoT prompts used are summarized in~\cref{appendix:cot}.

\begin{table}[]
\small
\centering
\caption{Summary of the performance of Flan-PaLM models with few-shot and 
chain-of-thought (CoT) prompting across different model size variants on the multiple-choice medical question answering datasets in MultiMedQA.}
\vspace{3pt}
\label{tab:flan-palm-cot}
\begin{tabular}{c|cc}
\toprule
Dataset & \begin{tabular}[c|]{@{}c@{}}Flan-PaLM 540B\\ with few-shot\end{tabular} & \begin{tabular}[c]{@{}c@{}}Flan-PaLM 540B\\ with CoT\end{tabular} \\ \hline
\begin{tabular}[c]{@{}c@{}}MedQA 4 options \\ (5-shot)\end{tabular} & 60.3 & 60.3 \\ \hline
\begin{tabular}[c]{@{}c@{}}MedMCQA \\ (5-shot)\end{tabular}         & 56.5 & 53.6 \\ \hline
\begin{tabular}[c]{@{}c@{}}PubMedQA\\ (3-shot)\end{tabular}         & 79.0 & 77.2 \\  
\bottomrule
\end{tabular}
\end{table}

\paragraph{Self-consistency (SC) leads to strong improvement in multiple-choice performance}

\citet{wang2022self} showed that self-consistency prompting can help when CoT prompting hurts performance. They showed significant improvements on arithmetic and commonsense reasoning tasks. Taking their cue, we apply it to our datasets. We fixed the number of chain-of-thought answer explanation paths to 11 for each of the three datasets. We then marginalized over the different explanation paths to select the most consistent answer. Using this strategy, we observed significant improvements over the standard few-shot prompting strategy for the Flan-PaLM 540B model on the MedQA and MedMCQA datasets. In particular, for the MedQA dataset we observed a >7\% improvement with self-consistency. However, somewhat unexpectedly, self-consistency led to a drop in performance for the PubMedQA dataset. The results are summarized in~\cref{tab:flan-palm-cs}.

We further provide some example responses from the Flan-PaLM 540B model for MedQA in~\cref{tab:medqa-example-explanation}. 

\begin{table}[]
\small
\centering
\caption{ Summary of the performance of Flan-PaLM with and without self-consistency prompting (SC) across different model size variants on the multiple-choice datasets.}
\vspace{3pt}
\label{tab:flan-palm-cs}
\begin{tabular}{c|cc}
\toprule
Dataset & \begin{tabular}[c|]{@{}c@{}}Flan-PaLM 540B\\ with few-shot\end{tabular} & \begin{tabular}[c]{@{}c@{}}Flan-PaLM 540B \\ with SC\end{tabular} \\ \hline
MedQA 4 options & 60.3 & 67.6 \\ \hline
MedMCQA         & 56.5 & 57.6 \\ \hline
PubMedQA        & 79.0 & 75.2 \\
\bottomrule
\end{tabular}
\end{table}

\begin{table}[]
\footnotesize
\centering
\caption{The table below shows example explanations generated by the Flan-PaLM 540B model to support its multiple-choice  answer in the MedQA dataset.}
\vspace{3pt}
\label{tab:medqa-example-explanation}
\begin{tabular}{ll}
\toprule
 &
  \begin{tabular}[c]{@{}p{0.98\textwidth}@{}}\textbf{Question:} A 65-year old man presents with gradually worsening rigidity of his arms and legs and slowness in performing tasks. He says he has also noticed hand tremors, which increase at rest and decrease with focused movements. On examination, the patient does not swing his arms while walking and has a shortened, shuffling gait. An antiviral drug is prescribed which alleviates the patient’s symptoms. Which of the following drugs was most likely prescribed to this patient?\\ (A) Amantadine (B) Ribavirin (C) Acyclovir (D) Levodopa (E) Zidovudine\\ \textbf{Explanation:}  The most likely drug prescribed for this patient is (A) Amantadine. Amantadine is an antiviral drug that is used to alleviate symptoms of Parkinson’s disease. Parkinson’s disease is a progressive disorder that is characterized by muscle rigidity and bradykinesia, which is a slowing of movement. Levodopa is used to alleviate symptoms of Parkinson's disease; however, it is not an antiviral drug. Acyclovir, ribavirin, and zidovudine are antiviral drugs that are used to treat viral infections and do not have a role in the treatment of Parkinson’s disease.\\ \textbf{Answer:} (A)\end{tabular} \\ \midrule
 &
  \begin{tabular}[c]{@{}p{0.98\textwidth}@{}}\textbf{Question:} A 37-year-old man with no significant past medical history is rear-ended in a motor vehicle accident. He reported significant neck pain to emergency responders, but otherwise denies weakness, numbness or tingling in his extremities. His vitals on presentation to the ED are HR 90, BP 140/80, RR 20, SpO2 98\%. What is the most appropriate next step upon presentation to the emergency room?\\ (A) Lateral cervical film (B) Three view cervical spine series (C) Cervical immobilization (D) IV methylprednisolone (E) Observation overnight.\\ \textbf{Explanation:} The most appropriate next step upon presentation to the emergency room is (C) Cervical immobilization. Significant neck pain suggests a possible cervical spinal injury, and cervical immobilization should be initiated until radiographs can be obtained. Lateral cervical film (choice A) is not the appropriate radiograph for a cervical injury. Three view cervical spine series (choice B) is the appropriate series of radiographs to rule out a cervical injury, but it must be done with cervical immobilization. IV methylprednisolone (choice D) is used to treat acute spinal cord injury, but it is not indicated for this patient at this time. Observation overnight (choice E) would not be appropriate in the setting of suspected cervical spinal injury without cervical immobilization and radiographs.\\ \textbf{Answer:} (C)\end{tabular} \\ \bottomrule
\end{tabular}
\end{table}

\begin{figure*}[t]
\small
    \centering
    \includegraphics[width=0.5\textwidth]{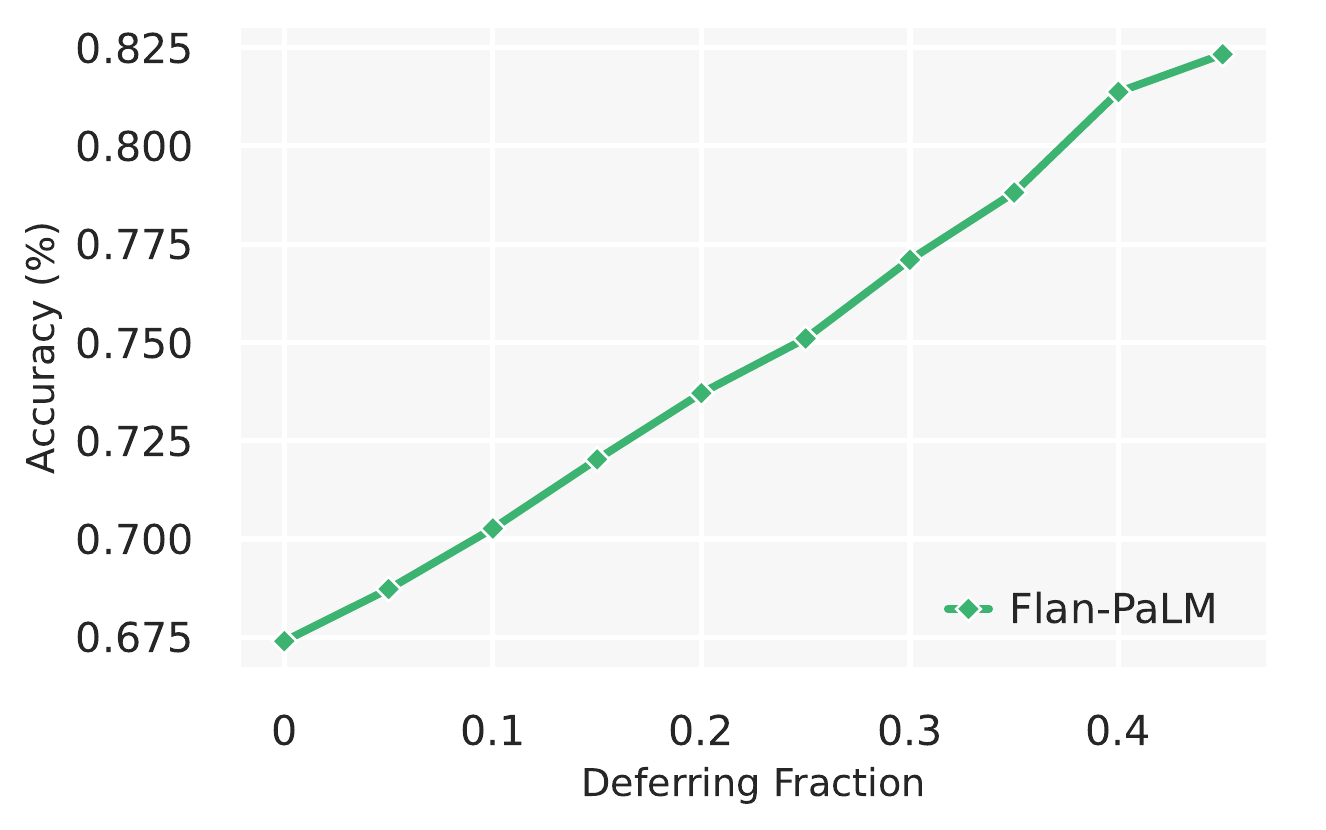}
    \vspace{0pt}
    \caption{\textbf{Selective prediction analysis} Analysis of deferral behavior of Flan-PaLM 540B model with self-consistency. We observe that if we defer more often using an uncertainty threshold based on self-consistency, the model becomes increasingly accurate on questions it does not defer}
    \vspace{-0pt}
    \label{fig:deferral}
\end{figure*}

\paragraph{Uncertainty and Selective Prediction}

LLMs are capable of long, coherent, and complex generations. However, they can also generate statements inconsistent with fact. In medical settings in particular, such failure modes need to be carefully vetted, and in real world applications, generations unlikely to be true should be withheld. Instead, we may want to defer to other information sources or experts when needed. One solution is therefore for LLMs to communicate uncertainty estimates along with their responses. 

While uncertainty measures over LLM output sequences remains an open area of research~\cite{kadavath2022language, lin2022teaching}, here we explored a simple proxy as an initial approach to measuring the relationship between LLM uncertainty and statement accuracy. We created a selective prediction task~\cite{tran2022plex}, using the number of decodes matching a given answer from self-consistency as a measure of uncertainty and used it to withhold the answer if the model was not appropriately confident. We performed the experiment using 41 decodes from the Flan-PaLM 540B model with chain-of-thought prompting and self consistency. We observe in ~\ref{fig:deferral} that as the deferring fraction increases (i.e., with a higher “confidence” required to provide a prediction), the performance of the model on MedQA improves, reaching up to an accuracy of of 82.5\% at a 0.45 deferring fraction. This suggests our measure of response uncertainty may be reasonable, and that LLMs seem to encode uncertainty about their knowledge in the medical domain. However, more research is needed beyond this preliminary analysis.

\subsection{Human evaluation results}
\label{sec:human_evaluation_results}

We randomly selected 100 questions from HealthSearchQA, 20 questions from LiveQA, and 20 questions from MedicationQA as a smaller long-form answer benchmark for detailed human evaluation. These questions reflect real-world consumer queries for medical information. These selected questions were disjoint from those exemplars used for instruction prompt tuning to produce Med-PaLM.

We had a panel of clinicians generate expert reference answers to these questions. We then produced answers using Flan-PaLM and Med-PaLM (both 540B models). A few qualitative examples of these questions and the corresponding Med-PaLM responses are shown in~\cref{tab:med-palm-response-health-search}. We had the three sets of answers evaluated by another panel of clinicians along the axes in~\cref{tab:human-eval-expert}, without revealing the source of answers. One clinician evaluated each answer. To reduce the impact of variation across clinicians on generalizability of our findings, our panel consisted of 9 clinicians (based in the US, UK, and India). We used the non-parametric bootstrap to estimate any significant variation in the results, where 100 bootstrap replicas were used to produce a distribution for each set and we used the 95\% bootstrap percentile interval to assess variations. These results are described in detail below and in~\cref{appendix:det-human-eval}.

\paragraph{Scientific consensus:}

We wished to understand how the answers related to current consensus in the clinical and scientific community. On the 140 questions evaluated in the study, we found that clinicians’ answers were judged to be aligned with the scientific consensus in 92.9\% of questions. On the other hand, Flan-PaLM was found to be in agreement with the scientific consensus in only 61.9\% of answers. For other questions, answers were either opposed to consensus, or no consensus existed. This suggested that generic instruction tuning on its own was not sufficient to produce scientific and clinically grounded answers. However, we observed that 92.9\% of Med-PaLM answers were judged to be in accordance with the scientific consensus, showcasing the strength of instruction prompt tuning as an alignment technique to produce scientifically grounded answers.

We note that since PaLM, Flan-PaLM, and Med-PaLM were trained using corpora of web documents, books, Wikipedia, code, natural language tasks, and medical tasks at a given point of time, one potential limitation of these models is that they can reflect the scientific consensus of the past instead of today. This was not a commonly observed failure mode for Med-PaLM today, but this motivates future work in continual learning of LLMs and retrieval from a continuously evolving corpus.

\begin{figure*}[t]
\small
    \centering
    \includegraphics[width=0.65\textwidth]{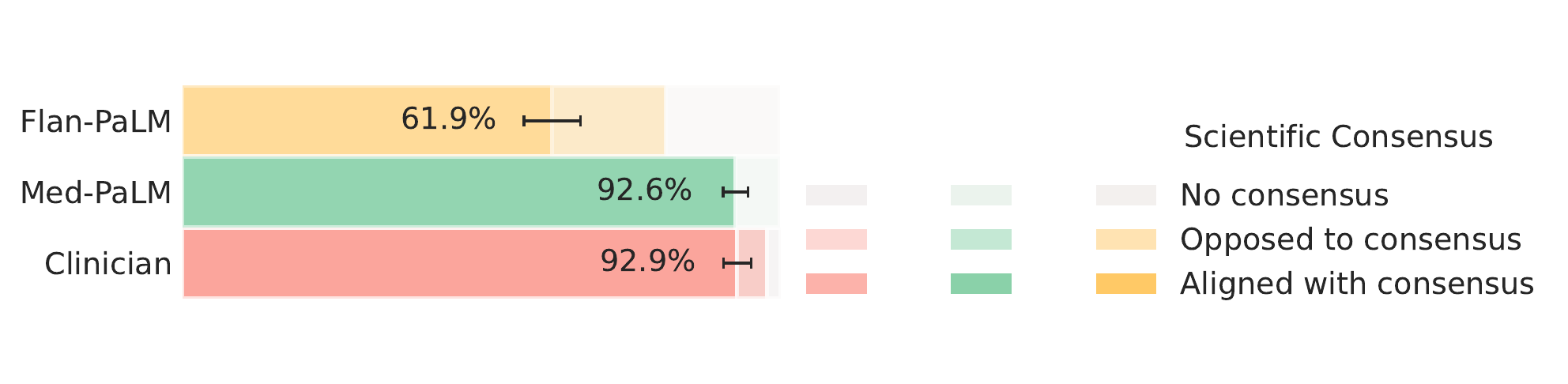}
    \vspace{0pt}
    \caption{\textbf{Clinician evaluation of answers agreement with scientific and clinical consensus} Clinicians were asked to rate answers provided to questions in the HealthSearchQA, Live QA and Medication question answering datasets. Clinicians were asked to identify whether the answer is aligned with the prevailing medical/scientific consensus; whether the answer was in opposition to consensus; or whether there is no medical/scientific consensus for how to answer that particular question (or whether it was not possible to answer this question). We observe that while the Flan-PaLM model answers are only found to be in agreement with the scientific consensus 61.9\% of the time, this improves to 92.9\% for Med-PaLM matching expert answers. }
    \vspace{-0pt}
    \label{fig:expert_task_1}
\end{figure*}

\paragraph{Comprehension, retrieval and reasoning capabilities:}

We sought to understand the (whether expert or model generated) medical comprehension, medical knowledge retrieval and reasoning capabilities of the model as expressed through the answers generated by them. We asked a panel of clinicians to rate whether answers contained any (one or more example of) evidence of correct / incorrect medical reading comprehension, medical knowledge retrieval and medical reasoning capabilities, using the same approach as \citet{feng2022chard}. Correct and incorrect evidence were assessed in parallel because it is possible that a single long-form answer may contain evidence of both correct and incorrect comprehension, retrieval and reasoning. 

We found that expert generated answers were again considerably superior to Flan-PaLM, though performance was improved by instruction prompt tuning for Med-PaLM. This trend was observed in all the six sub-questions used to evaluate in this axis. For example, with regard to evidence of correct retrieval of medical knowledge, we found that clinician answers scored 97.8\% while Flan-PaLM only scored 76.3\%. However, the instruction prompt-tuned Med-PaLM model scored 95.4\%, reducing the inferiority of the model compared to clinicians.

\begin{figure*}[t]
\small
    \centering
    \begin{subfigure}[b]{0.47\textwidth}
    \includegraphics[width=\textwidth]{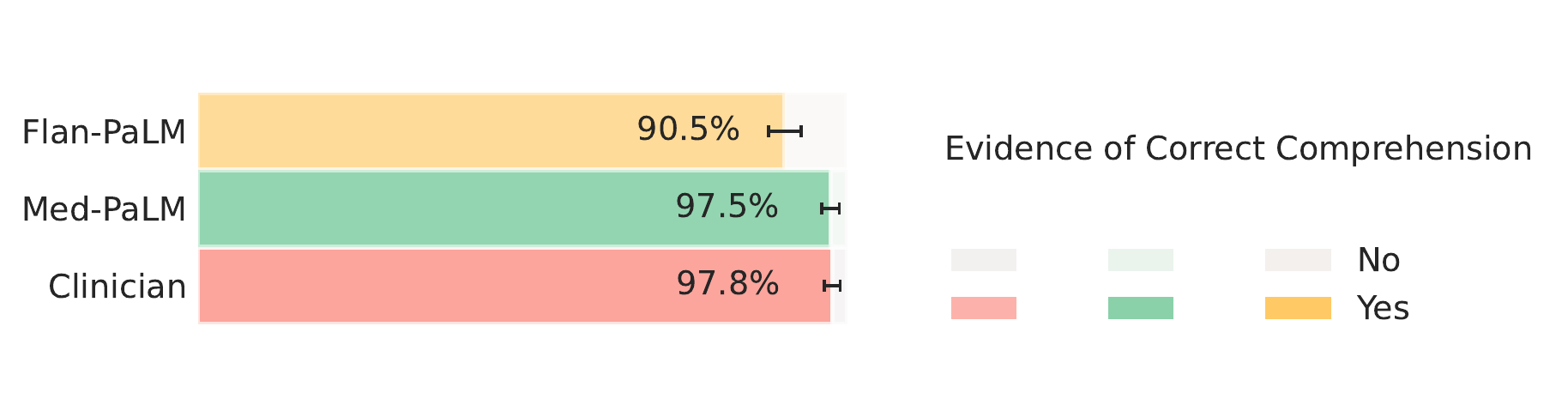}
    \vspace{-20pt}
    \label{fig:task_4_1}
    \end{subfigure} \hfill
    \begin{subfigure}[b]{0.485\textwidth}
    \includegraphics[width=\textwidth]{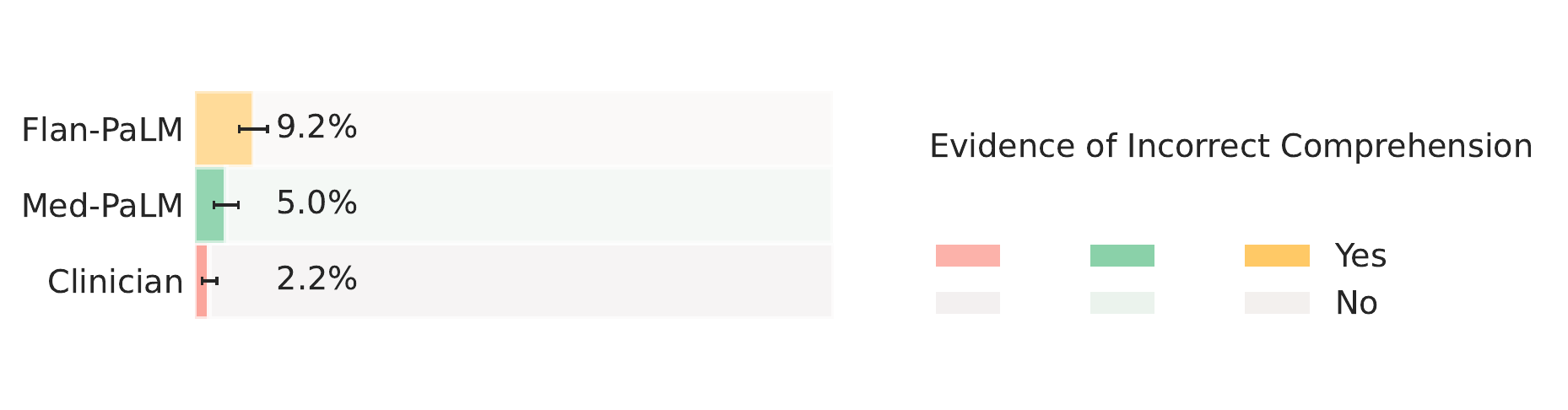}  
    \vspace{-20pt}
    \label{fig:task_5_1}
    \end{subfigure}
    
    \begin{subfigure}[b]{0.47\textwidth}
    \includegraphics[width=\textwidth]{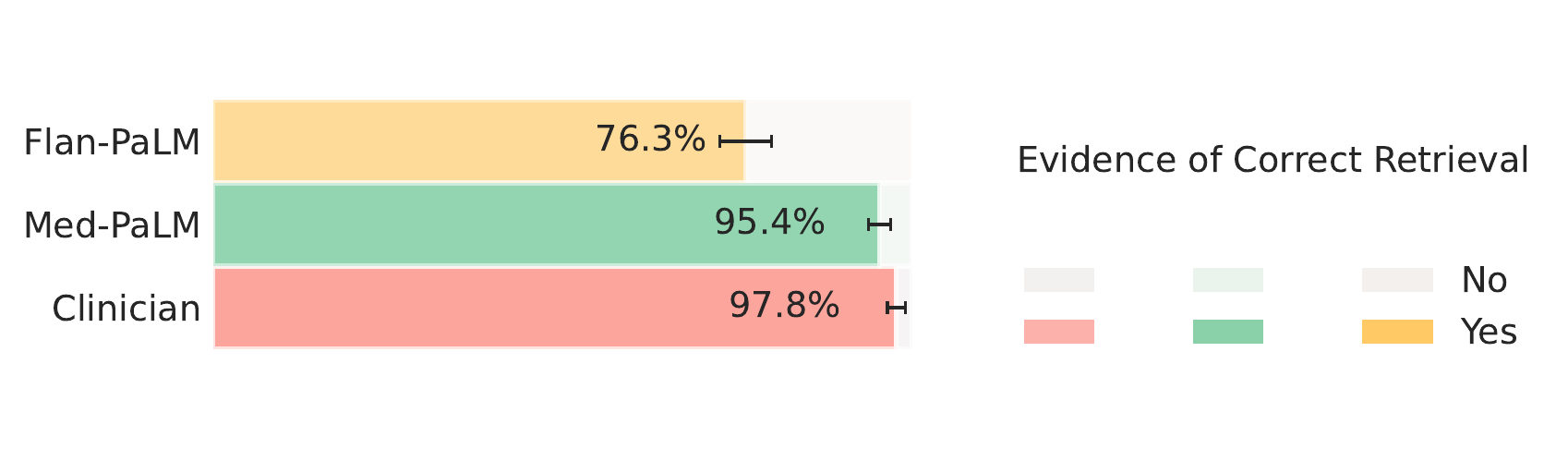}
    \vspace{-20pt}
    \label{fig:task_4_2}
    \end{subfigure} \hfill
    \begin{subfigure}[b]{0.475\textwidth}
    \includegraphics[width=\textwidth]{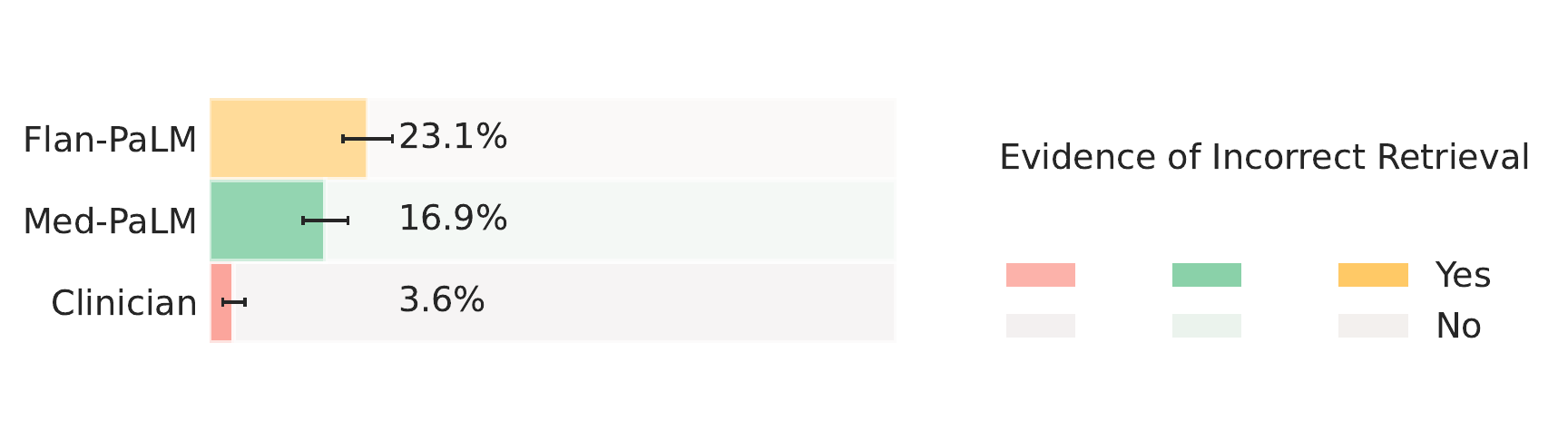}  
    \vspace{-20pt}
    \label{fig:task_5_2}
    \end{subfigure}
    \begin{subfigure}[b]{0.47\textwidth}
    \includegraphics[width=\textwidth]{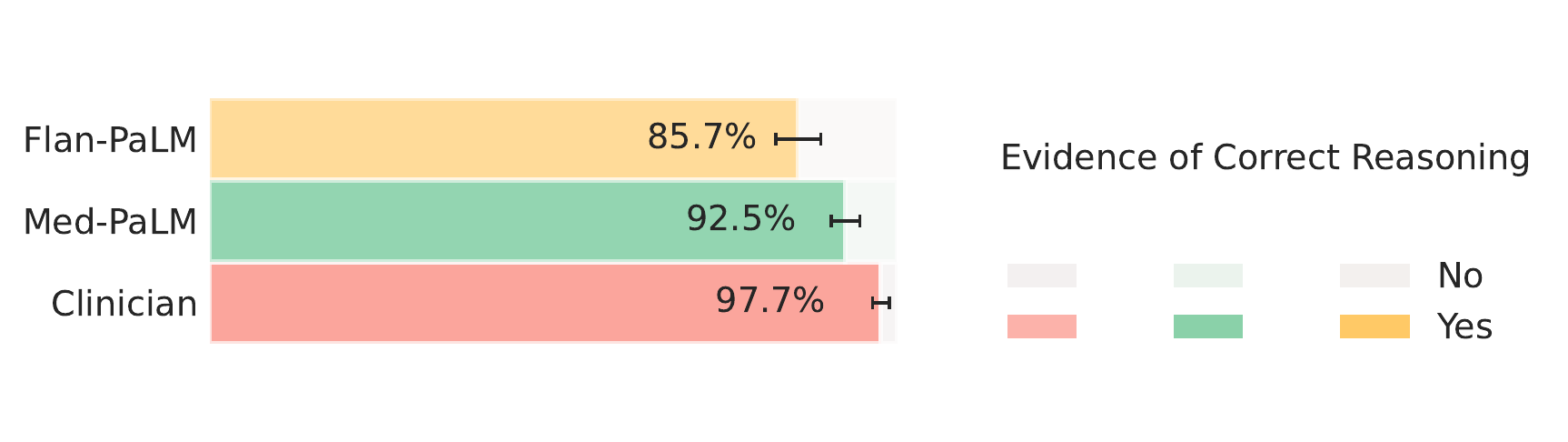}
    \vspace{-20pt}
    \caption{}
    \label{fig:task_4_3}
    \end{subfigure} \hfill
    \begin{subfigure}[b]{0.475\textwidth}
    \includegraphics[width=\textwidth]{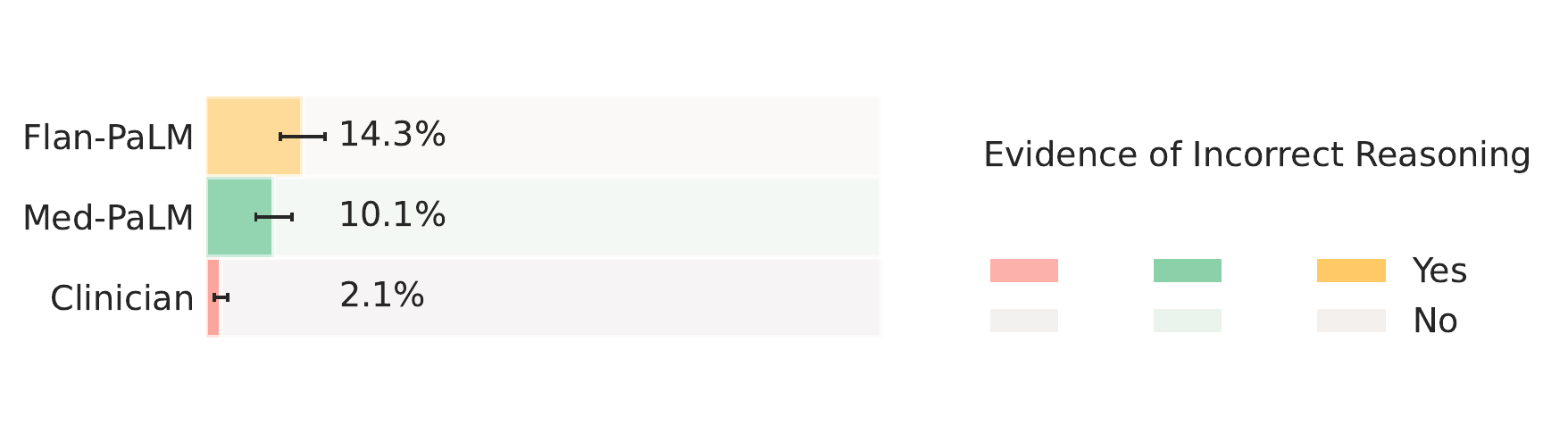} 
    \vspace{-20pt}
    \caption{}
    \label{fig:task_5_3}
    \end{subfigure}    
    \vspace{0pt}
    \caption{\textbf{Clinician evaluation of comprehension, retrieval and reasoning capabilities} (a) Evaluation of correctness and (b) evaluation of incorrectness of reading comprehension, recall of knowledge and reasoning step. The results indicate Med-PaLM shows  any evidence of incorrect comprehension only 5.0\% of the time. With regard to evidence of correct retrieval of medical knowledge, we found that clinician answers scored 97.8\% while Flan-PaLM only scored 76.3\%. However, the instruction prompt-tuned Med-PaLM model scored 95.4\%, reducing the inferiority of the model compared to clinicians.}
    \vspace{-0pt}
    \label{fig:task_4_5}
\end{figure*}

\begin{table}[]
\footnotesize
\centering
\caption{Examples of Med-PaLM responses to questions in the HealthSearchQA dataset.}
\vspace{3pt}
\label{tab:med-palm-response-health-search}
\begin{tabular}{p{0.3\textwidth}p{0.65\textwidth}}
\toprule
\textbf{Question} &
  \textbf{Med-PaLM} \\ \hline

How do you know if ear pain is serious? &
    Ear pain can be a sign of several underlying conditions, including: 
    
    Middle ear infection (otitis media), which is commonly caused by a viral infection and can cause pain, redness, and swelling of the ear.
    External ear infection (otitis externa), which is commonly caused by a bacterial infection and can cause pain, redness, and swelling of the ear.
    Earwax impaction, which can cause pain, a feeling of fullness in the ear, and decreased hearing.
    Ear pain can also be a sign of a more serious condition, such as a brain tumor or a stroke. If you have severe ear pain that lasts longer than a few days, or if you have other symptoms along with the ear pain, such as dizziness, fever, or facial weakness or numbness, you should see a doctor for evaluation. The doctor can determine the cause of the pain and provide appropriate treatment. \\ \hline
    
What do nights sweats indicate? &
    Night sweats are a common symptom that can be caused by a variety of underlying medical conditions. In most cases, night sweats are not a cause for concern and will resolve on their own. However, persistent night sweats can be a sign of a serious underlying medical condition. Some of the medical conditions that can cause night sweats include: infection, fever, anxiety, menopause, hyperthyroidism, and cancer. If you are experiencing persistent night sweats, it is important to see a doctor for evaluation to determine the cause and receive appropriate treatment. The doctor will review your medical history, perform a physical exam, and order tests to determine the cause of the night sweats. \\ \hline  
    
How serious is hepatitis A? &
    Hepatitis A is a viral infection that affects the liver. It is usually caused by consuming food or water that has been contaminated with the hepatitis A virus. Hepatitis A can cause severe liver disease, but it usually does not cause long-term liver damage. Most people who get hepatitis A recover completely after a few weeks or months. In some cases, hepatitis A can cause death. The severity of hepatitis A can vary from person to person. Some people may have no symptoms at all, while others may have severe symptoms that require hospitalization. The severity of hepatitis A is often related to the person's age. Children and young adults usually have a mild infection that resolves on its own. Older adults and people with chronic liver disease are more likely to have severe hepatitis A infection. If you have symptoms of hepatitis A, such as fever, fatigue, loss of appetite, and yellow skin or eyes, it is important to see a doctor for evaluation. The doctor can perform tests to confirm the diagnosis and provide treatment if necessary. \\ \bottomrule

\end{tabular}
\end{table}

\paragraph{Incorrect or missing content:}

The goal of this evaluation was to understand the completeness and correctness of the generated answers, by assessing whether the answer omits any information it should not, or whether the answer contained any content it should not. Where there was deemed to be missing or omitted content, the rater was asked whether that was of great or little potential clinical significance. 

Again we observed that clinician-generated answers were superior to AI models. Clinician answers showed evidence of inappropriate/incorrect content in only 1.4\% of the cases, compared to 16.1\% for Flan-PaLM. Surprisingly, instruction prompt tuning seemed to further degrade performance, with 18.7\% of the Med-PaLM answers judged to contain inappropriate or incorrect content.

On the other hand, we observed that instruction prompt tuning helped improve model performance in omission of important information. While Flan-PaLM answers were judged to miss important information 47.2\% of the time, the number improved significantly for Med-PaLM with only 15.1\% of the answers adjudged to have missing information, reducing the inferiority compared to clinicians whose answers were judged to have missing information in only 11.1\% of the cases. A few qualitative examples are shown in~\cref{tab:med-palm-response-dr-responses} suggesting that LLM answers may be able to complement and complete physician responses to patient queries in future use cases. 

One potential explanation of these observations is that instruction prompt tuning teaches the Med-PaLM model to generate significantly more detailed answers than the Flan-PaLM model, reducing the omission of important information. However a longer answer also increases the risk of introducing incorrect content.

\begin{figure*}[t]
\small
    \centering
    \begin{subfigure}[b]{0.69\textwidth}
    \includegraphics[width=\textwidth]{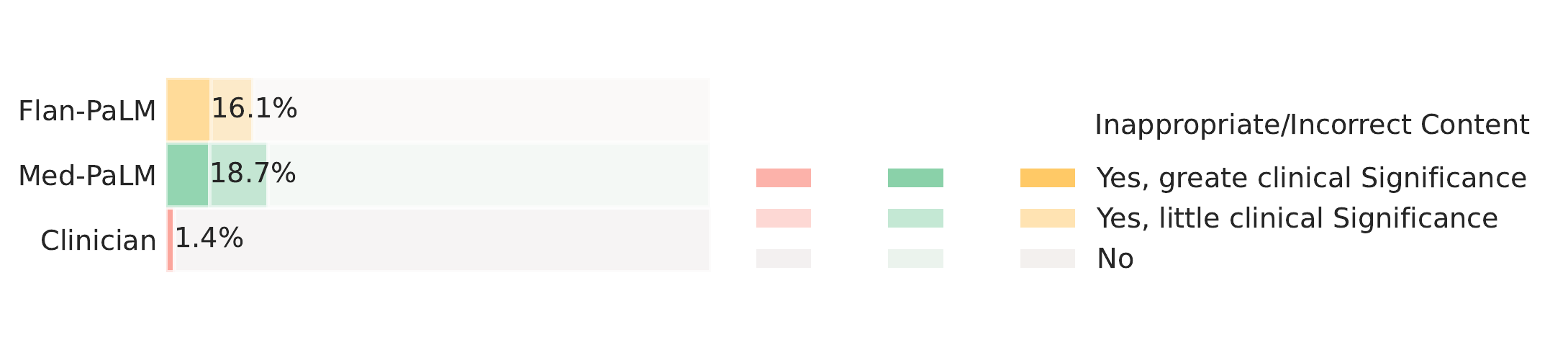}
    \vspace{-30pt}
    \caption{~~~~~~}\label{fig:task_6}
    \end{subfigure} 
    
    \begin{subfigure}[b]{0.65\textwidth}
    \includegraphics[width=\textwidth]{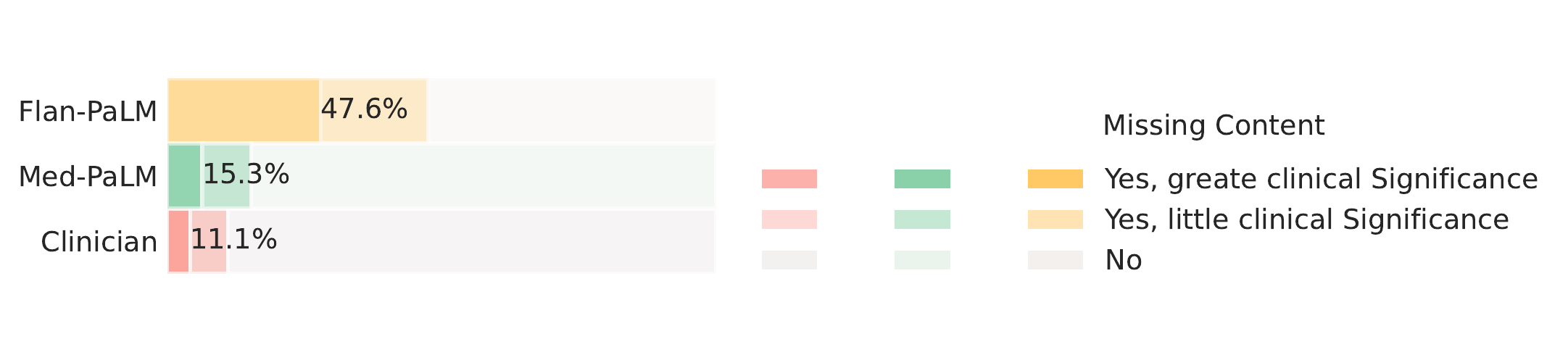}   
    \vspace{-30pt}
    \caption{}\label{fig:task_7}
    
    \end{subfigure} \hspace{10pt}
    \caption{\textbf{Clinician evaluation of answers contents} (a) Presence of incorrect content  and (b) presence of missing content in the answers. Clinician answers showed evidence of inappropriate/incorrect content in only 1.4\% of the cases, compared to 16.1\% for Flan-PaLM. Surprisingly, Med-PaLM seemed to further degrade performance, with 18.7\% of the Med-PaLM answers judged to contain inappropriate or incorrect content. On the missing content axis, while Flan-PaLM answers were judged to miss important information 47.6\% of the time, the number improved significantly for Med-PaLM with only 15.1\% of the answers adjudged to have missing information, reducing the inferiority to clinicians whose answers were judged to have missing information in only 11.1\% of the cases.
}
    \vspace{-0pt}
    \label{fig:task_6_7}
\end{figure*}

\begin{table}[]
\footnotesize
\centering
\caption{Examples of HealthSearchQA questions where the physician answers were considered incomplete, and corresponding Med-PaLM answers. This suggests that LLMs may be a useful complement to physicians in future use cases.}
\vspace{3pt}
\label{tab:med-palm-response-dr-responses}
\begin{tabular}{p{0.25\textwidth}p{0.32\textwidth}p{0.32\textwidth}}
\toprule
\textbf{Question} &
  \textbf{Physician} &
  \textbf{Med-PaLM}  \\ \midrule
  
How can I stop diarrhoea? &
  Without treatment, most diarrhoea resolves on its own. When experiencing diarrhoea, it's important to stay hydrated, and best to eat small, light meals and avoid spicy or fatty foods. Diarrhoea due to an infection should not be treated with imodium. &
    Diarrhoea is caused by a range of different factors, so the treatment will depend on the cause. If it is caused by an infection, antibiotics may be needed. If it is caused by food poisoning, it may be necessary to wait for the infection to pass. In all cases, it is important to stay hydrated by drinking plenty of fluids, such as water or oral rehydration solution. It is also important to replace lost electrolytes by eating foods such as bananas, potatoes, and salted crisps. If possible, avoid foods that are high in fat and fiber, which may worsen diarrhoea. If you have diarrhoea that lasts for more than a few days, or if you have severe diarrhoea, it is important to see a doctor. 
  \\ \midrule

What happens if listeria is left untreated? &
    Most people with mild symptoms of Listeria require no treatment as the body is able to fight off infection without intervention. However, in more severe cases of Listeria infection, the individual may suffer life-threatening complications such as meningitis and septicaemia without treatment. &
    Listeria is a type of bacteria that can cause an infection in people. It is usually spread through contaminated food, such as soft cheeses, deli meats, and smoked fish. Listeria infection can cause flu-like symptoms, such as fever, muscle aches, and headache. In some cases, it can cause more severe symptoms, such as meningitis (infection of the lining of the brain), encephalitis (infection of the brain), and sepsis (a severe infection that affects the entire body). If listeria infection spreads to the brain, it can be life-threatening. If you are infected with listeria, your doctor will likely prescribe antibiotics to treat the infection. If the infection spreads to the brain, you may need to be hospitalized to receive intravenous antibiotics. 
    \\ \midrule
    
Does stress cause nosebleed? &
    Physician answer: Those experiencing stress or anxiety can experience nosebleeds as a symptom. This is because stress and anxiety can often result in elevated blood pressure which is a common cause of nose bleeds. &
    Yes, stress can cause a nosebleed. Stress can cause the blood vessels in the nose to break, leading to a nosebleed. Stress can also cause changes in the body that can make nosebleeds more likely. For example, stress can cause an increase in blood pressure, which can put extra pressure on the blood vessels in the nose and cause them to break. Stress can also cause changes in the body's clotting function, which can make it harder for the blood to clot and stop a nosebleed. Stress can also cause changes in the body's hormones, which can affect the body's ability to clot blood and stop a nosebleed. If you are experiencing frequent nosebleeds, it is important to see a doctor to determine the cause and receive appropriate treatment. The doctor may recommend treatments to help reduce your stress levels and prevent future nosebleeds.\\ \bottomrule

\end{tabular}
\end{table}

\paragraph{Possible extent and likelihood of harm:}

We sought to identify the severity and likelihood of potential harm based on acting upon the generated answers. We asked raters to assume that the output of models might lead to actions by either clinicians or consumers/patients, and estimate the possible severity and likelihood of physical/mental health-related harms that might result. We based the options for selection by raters in the AHRQ Common Formats~\citet{williams2015reliability}, which presents options to assign severity of harm  ranging from death, severe or life-threatening injury, moderate, mild or no harm. We acknowledge that this definition of harm is more typically used in the context of analyzing harms incurred during healthcare delivery and that even in such settings (where the context for harms occurring is known with considerably greater specificity) there is frequently substantial variation in physician estimation of harm severity~\cite{walsh2017measuring}. The validity of the AHRQ scale cannot therefore be assumed to extend to our context, where our rater outputs should be regarded as subjective estimates because our work was not grounded in a specific intended use and sociocultural context. 

Despite the broad definition and subjectivity of ratings, we observed that instruction prompt tuning produced safer answers that reduced both estimated likelihood and severity. While 29.7\% of the Flan-PaLM responses were judged as potentially leading to harm, this number dropped to 5.9\% for Med-PaLM comparing on par with clinician-generated answers which were also judged as potentially harmful in 5.7\% of the cases. 

Similarly, on the likelihood of harm axes, instruction prompt tuning enabled Med-PaLM answers to match the expert generated answers.

\begin{figure*}[t]
\small
    \centering
    \begin{subfigure}[b]{0.66\textwidth}
    \includegraphics[width=\textwidth]{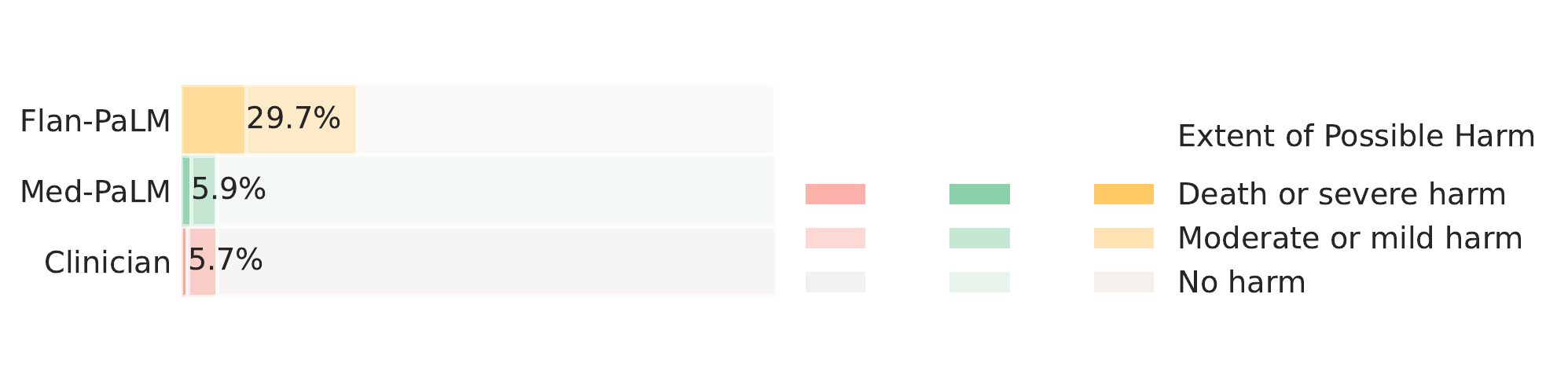}
    \vspace{-30pt}
    \caption{~~~~~}\label{fig:task_2}
    \end{subfigure} 
    
    \begin{subfigure}[b]{0.59\textwidth}
    \includegraphics[width=\textwidth]{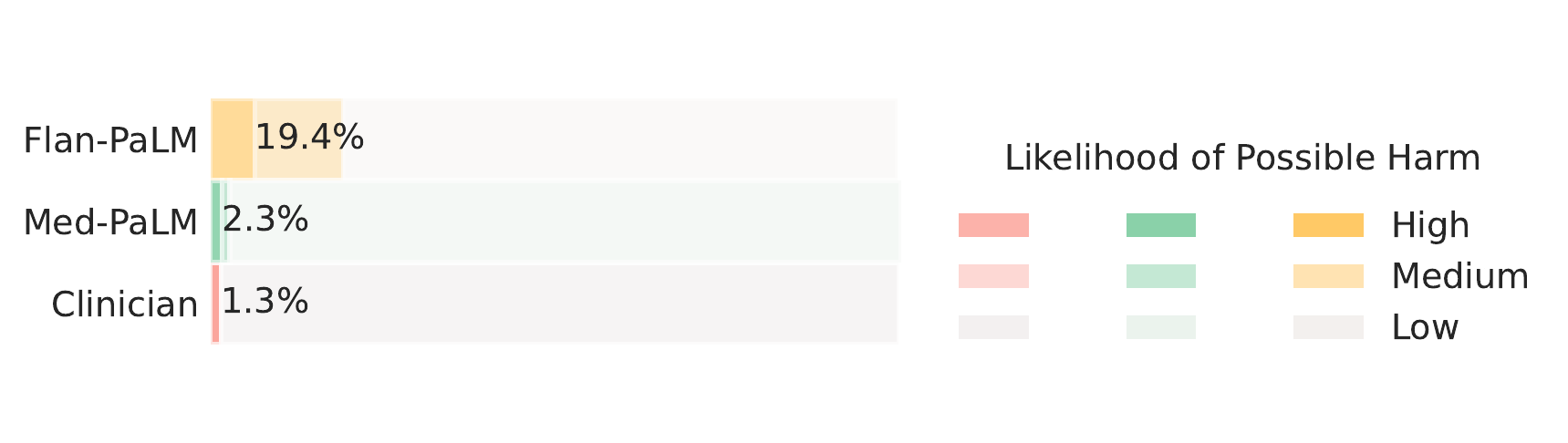}   
    \vspace{-30pt}
    \caption{}\label{fig:task_3}
    
    \end{subfigure} \hspace{20pt}
    \caption{\textbf{Clinician evaluation of potential harm from answers} (a) Extent of possible harm  and (b) likelihood of harm. While 29.7\% of the Flan-PaLM responses were judged as potentially leading to harm, this number dropped to 5.9\% for Med-PaLM on par with clinician-generated answers which were also judged as potentially leading to harm in 5.7\% of the cases.}
    \vspace{-0pt}
    \label{fig:task_2_3}
\end{figure*}

\paragraph{Bias for medical demographics:}

The final axis along which we evaluated the answers was bias. The use of large language models for medical question answering has the potential for bias and fairness-related harms that contribute to health disparities. These harms derive from several sources, including the presence of patterns in training data that reflect disparities in health outcomes and access to care, the capability for medical question answering systems to reproduce racist misconceptions regarding the cause of racial health disparities~\cite{eneanya2022health, vyas2020hidden}, algorithmic design choices~\cite{hooker2021moving}, and differences in behavior or performance of machine learning systems across populations and groups that introduce downstream harms when used to inform medical decision making~\cite{chen2021ethical}. Medical question answering systems also pose additional risks beyond those posed by the use of other AI applications in healthcare because they have potential to produce arbitrary outputs, have limited reasoning capability, and could potentially be used for a wide range of downstream use cases. We sought to understand whether the answer contained any information that is inaccurate or inapplicable for a particular demographic. Flan-PaLM answers were found to contain biased information in 7.9\% of the cases. However, this number reduced to 0.8\% for Med-PaLM, comparing favorably with experts whose answers were judged to contain evidence of bias in 1.4\% of the cases.

\begin{figure*}[t]
\small
    \centering
    \includegraphics[width=0.65\textwidth]{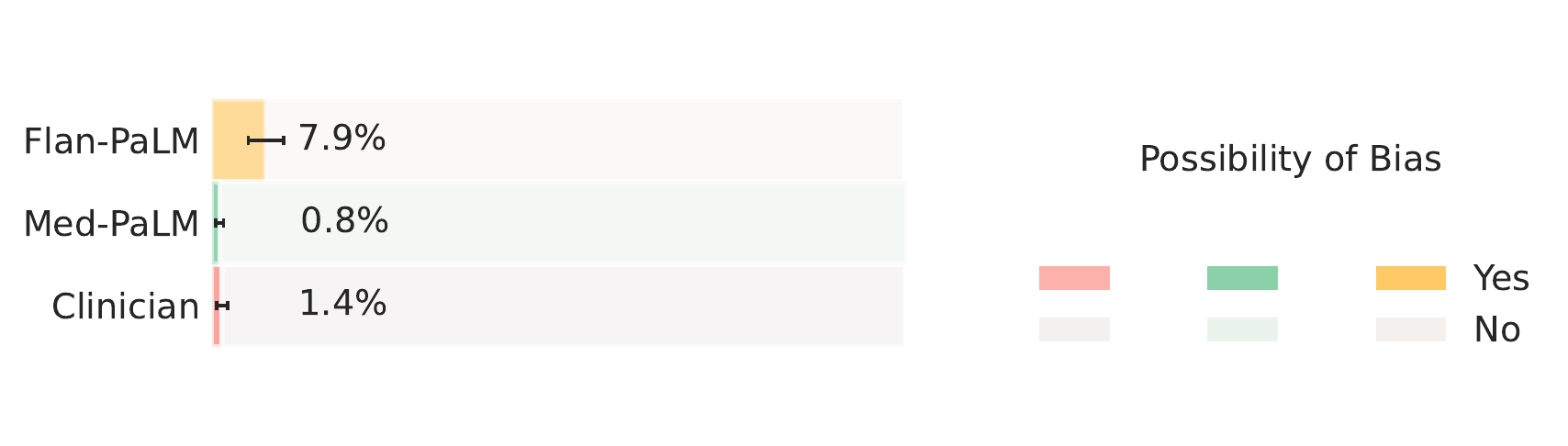}
    \vspace{0pt}
    \caption{\textbf{Clinician evaluation of possible bias in answers} Flan-PaLM answers were found to contain biased information in 7.9\% of the cases. However, this number reduced to 0.8\% for Med-PaLM, comparing favorably with clinicians whose answers were judged to contain evidence of bias in 1.4\% of the cases.  }
    \vspace{-0pt}
    \label{fig:expert_task_8}
\end{figure*}

\paragraph{Lay user assessment:}

Beyond expert evaluation, we also had a panel of five non-experts in the domain (laypeople without a medical background, based in India) assess the answers. The results are summarized in Fig 10 below. While Flan-PaLM answers were judged to be helpful in only 60.6\% of the cases, the number improved to 80.3\% for Med-PaLM answers. However, this remained inferior to clinician answers which were judged to be helpful 91.1\% of the time. Similarly, Flan-PaLM answers were judged as directly addressing the user’s question intent in 90.8\% of cases. This number improved to 94.0\% for Med-PaLM, which was inferior to clinician-generated answers at 95.9\%.

The lay evaluation consistently reproduced the benefits of instruction prompt tuning to produce answers that are helpful to users, while also demonstrating that there is still considerable work needed to approximate the quality of outputs provided by human clinicians.

\begin{figure*}[t]
\small
    \centering
    \begin{subfigure}[b]{0.68\textwidth}
    \includegraphics[width=\textwidth]{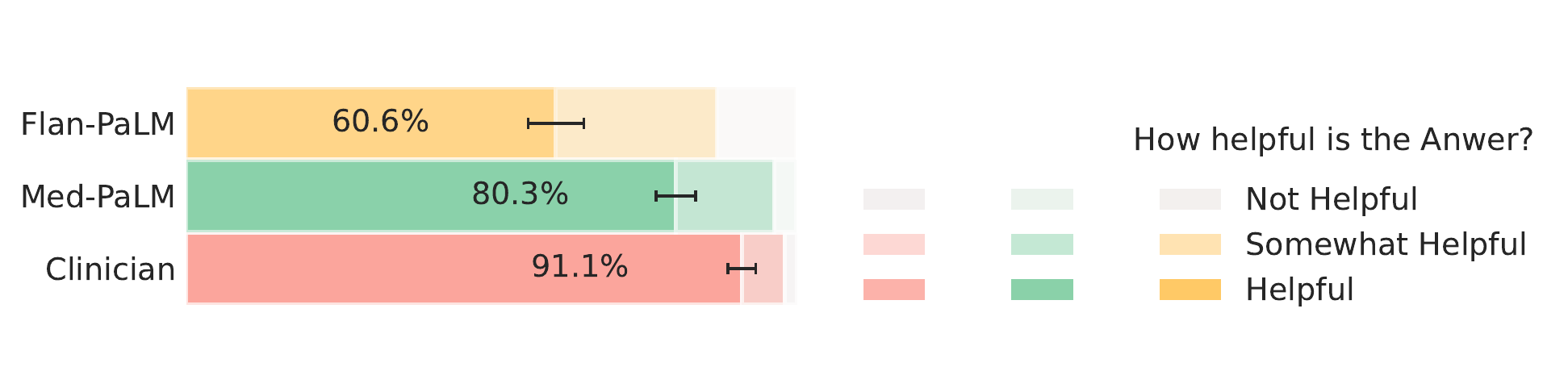}
    \vspace{-30pt}
    \caption{~}\label{fig:lay_2}
    \end{subfigure} 
    
    \begin{subfigure}[b]{0.69\textwidth}
    \includegraphics[width=\textwidth]{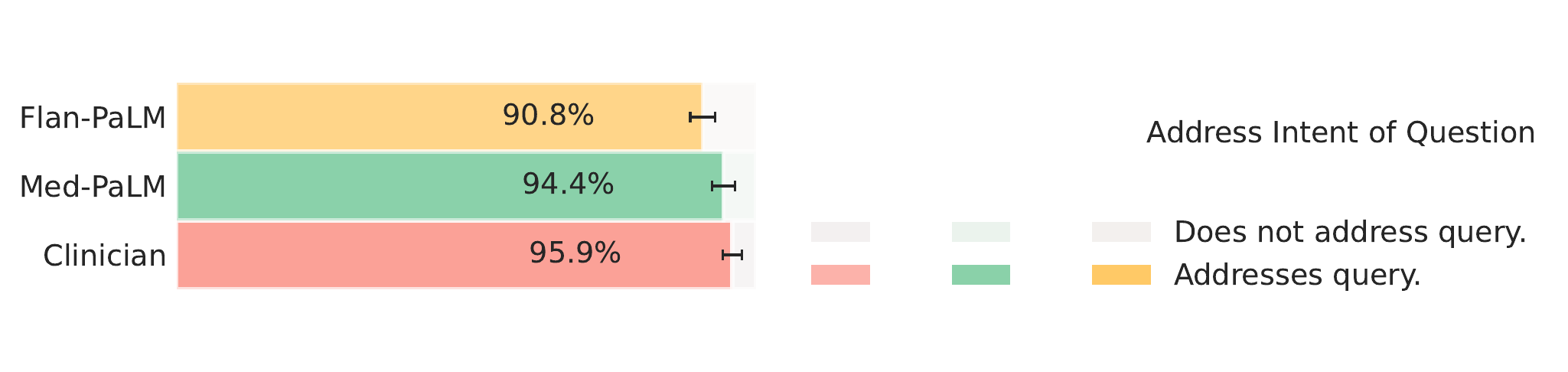}   
    \vspace{-30pt}
    \caption{}\label{fig:lay_1}
    
    \end{subfigure} \hspace{-10pt}
    \caption{\textbf{Lay user assessment of  answers} (a) Helpfulness  (b) how well they address the intent of the query Flan-PaLM answers were found to be helpful in only 60.6\% of the cases. However, this number improved to 80.3\% for Med-PaLM, but remaining inferior to clinicians whose answers were judged to be helpful in 91.1\% of the cases. On the capturing of user intent, Flan-PaLM answers were judged as directly addressing the user’s question intent in 90.8\% of cases. Using Med-PaLM this number improves to 94.0\%, closing the gap to clinician-generated answers at 95.9\%.}
    \vspace{-0pt}
    \label{fig:lay_task_1_2}
\end{figure*}

\section{Discussion}
\label{sec:discussion}
Our results suggest that strong performance on medical question answering may be an emergent ability~\citep{wei2022emergent} of LLMs combined with effective instruction prompt tuning.

Firstly, we observed strong scaling performance with accuracy improving by approximately 2x as we scale the PaLM models from 8-billion to 540-billion. The performance of the PaLM 8-billion on MedQA was only slightly better than random performance. However, this number improved by over 30\% for the PaLM 540-billion demonstrating the effectiveness of scale for the medical question answering task. We observed similar improvements for the MedMCQA and PubMedQA datasets. Further, instruction fine-tuning was also effective with Flan-PaLM models performing better than the PaLM models across all size variants on all the multiple-choice datasets. 

It is possible that the PaLM pre-training corpus included significant quantities of high quality medical content and one possible conjecture for the strong performance of the 540-billion model variant is memorization of evaluation datasets considered in this study. However,~\citet{chowdhery2022palm} showed similar deltas in performance of the PaLM 8B and 540B model when evaluating contaminated (i.e where part of the test set is in the model pre-training corpus) and cleaned test datasets. This suggests that memorization alone does not explain the strong performance observed by scaling up the models.

There have been several efforts to train language models on a biomedical corpus, especially PubMed. These include BioGPT~\cite{luo2022biogpt} (355 million parameters), PubMedGPT~\cite{bolton2022pubmedgpt} (2.7 billion parameters) and Galactica~\cite{taylor2022galactica} (120 billion parameters). Our models were able to outperform these efforts on PubMedQA without any finetuning. Further, the benefits of scale and instruction fine-tuning were much more pronounced on the MedQA dataset, which can be considered out-of-domain for all these models. Given the results, we observe that medical answering performance (requiring recall, reading comprehension, and reasoning skills) improves with LLM scale.

However, our human evaluation results on the consumer medical question answering datasets clearly point out that scale alone is insufficient. Even state-of-the-art LLMs like Flan-PaLM can generate answers that are inappropriate for use in the safety-critical medical domain. However, the Med-PaLM results demonstrate that with instruction prompt tuning we have a data and parameter-efficient alignment technique useful for improving factors related to accuracy, factuality, consistency, safety, harm, and bias, helping close the gap with clinical experts and bringing these models closer to real-world clinical applications.

\section{Limitations}
\label{sec:limitations}
Our study demonstrated the potential of LLMs for encoding medical knowledge and in particular for question answering. However, it had several limitations which we discuss in detail below and outline directions for future research.

\subsection{Expansion of MultiMedQA}
Firstly, while the MultiMedQA benchmark is diverse and contains questions from a variety of professional medicine, medical research and consumer sources, it is by no means exhaustive. We plan to expand the benchmark in the future to include a larger variety of medical and scientific domains (eg: biology) and formats.

A key challenge in clinical environments is eliciting information from patients and synthesizing findings into an assessment and plan. Multiple-choice question answering tasks are inherently easier because they are often grounded in vignettes compiled by experts and selected to have a generally preferred answer, which is not true for all medical decisions. Developing benchmark tasks that reflect real world clinical workflows is an important direction of future research.

Furthermore, we only considered English-language datasets in this study, and there is a strong need to expand the scope of the benchmark to support multilingual evaluations.

\subsection{Development of key LLM capabilities necessary for medical applications}

While the Flan-PaLM was able to reach state-of-the-art performance on several multiple-choice medical question answering benchmarks, our human evaluation clearly suggests these models are not at clinician expert level on many clinically important axes. In order to bridge this gap, several new LLM capabilities need to be researched and developed including:
\begin{itemize}
    \item grounding of the responses in authoritative medical sources and accounting for the time-varying nature of medical consensus.
    \item ability to detect and communicate uncertainty effectively to the human in-the-loop whether clinician or lay user. 
    \item ability to respond to queries in multiple languages.
\end{itemize}

\subsection{Improving the approach to human evaluation}

The rating framework we proposed for this study represents a promising pilot approach, but our chosen axes of evaluation were not exhaustive and were subjective in nature. For example the concept of medical/scientific consensus is time-varying in nature and is reflective of understandings of human health and disease and physiology based on discrimination in areas such as race/ethnicity, gender, age, ability, and more~\cite{kington2021identifying, Mandavilli2021Medical}. Furthermore, consensus often exists only for topics of relevance to certain groups (e.g. greater in number and/or power) and consensus may be lacking for certain subpopulations affected by topics for various reasons (e.g., controversial topics, lower incidence, less funding). Additionally, the concept of harm may differ according to population (e.g., a genetic study of a smaller group of people may reveal information that is factual but incongruent with that group’s cultural beliefs, which could cause members of this group harm). Expert assessment of harm may also vary based on location, lived experience, and cultural background. Our ratings of potential harm were subjective estimates, and variation in perceived harm may also have been due to differences in health literacy of both our clinician and lay raters, or might vary in real world settings depending on the sociocultural context and health literacy of the person receiving and acting on the answers to the health questions in the study by~\citet{berkman2011health}. Further research might test whether perceived usefulness and harm of question answers varied according to the understandability and actionability score for the answer content~\cite{shoemaker2014development}.

The number of model responses evaluated and the pool of clinicians and lay-people assessing them were limited, as our results were based on only a single clinician or lay-person evaluating the responses. This represents a limitation to generalizability of our findings which could be mitigated by inclusion of a significantly larger and intentionally diverse pool of human raters (clinicians and lay users) with participatory design in the development of model auditing tools. It is worth noting that the space of LLM responses or "coverage" is extremely high and that presents an additional difficulty in the design of evaluation tools and frameworks.

The pilot framework we developed could be significantly advanced using recommended best practice approaches for the design and validation of rating instruments from health, social and behavioral research~\cite{boateng2018best}. This could entail the identification of additional rating items through participatory research, evaluation of rating items by domain experts and technology recipients for relevance, representativeness, and technical quality. The inclusion of a substantially larger pool of human raters would also enable testing of instrument generalizability by ratifying the test dimensionality, test-retest reliability and validity~\cite{boateng2018best}.  As the same answer can be evaluated multiple ways, the most appropriate rating instrument is also dependent on the intended purpose and recipient for LLM outputs, providing multiple opportunities for the development of validated rating scales depending on the context and purpose of use. Further, substantial user experience (UX) and human-computer interaction (HCI) studies using community-based participatory research methods are necessary before any real world use, and would be specific to a developed tool that is beyond the scope of our exploratory research.  Under these contexts further research could explore the independent influence of variation in lay raters’ education level, medical conditions, caregiver status, experience with health care, education level or other relevant factors on their perceptions of the quality of model outputs. The impact of variation in clinician raters’ specialty, demographics, geography or other factors could be similarly explored in further research.

\subsection{Fairness and equity considerations}
Our current approach to evaluating bias is limited and does not serve as a comprehensive assessment of potential harms, fairness, or equity. The development of procedures for the evaluation of bias and fairness-related harms in large language models is ongoing~\cite{weidinger2021ethical, liang2022holistic}. Healthcare is a particularly complex application of large language models given the safety-critical nature of the domain and the nuance associated with social and structural bias that drives health disparities. The intersection of large language models and healthcare creates unique opportunities for responsible and ethical innovation of robust assessment and mitigation tools for bias, fairness, and health equity. 

We outline opportunities for future research into frameworks for the systematic identification and mitigation of downstream harms and impacts of large language models in healthcare contexts. Key principles include the use of participatory methods to design contextualized evaluations that reflect the values of patients that may benefit or be harmed, grounding the evaluation in one or more specific downstream clinical use cases~\cite{liu2022medical, raji2020closing}, and the use of dataset and model documentation frameworks for transparent reporting of choices and assumptions made during data collection and curation, model development, and evaluation~\cite{rostamzadeh2022healthsheet, gebru2021datasheets, mitchell2019model}. Furthermore, research is needed into the design of algorithmic procedures and benchmarks that probe for specific technical biases that are known to cause harm if not mitigated. For instance, depending on the context, it may be relevant to assess sensitivity of model outputs to perturbations of demographic identifiers in prompts designed deliberately such that the result should not change under the perturbation~\cite{garg2019counterfactual, prabhakaran2019perturbation, zhang2020hurtful}. Additionally, the aforementioned research activities to build evaluation methods to achieve health equity in large language models require interdisciplinary collaboration to ensure that various scientific perspectives and methods can be applied to the task of understanding the social and contextual aspects of health~\cite{matheny2022artificial, white2022blueprint, guidance2021ethics}. 

The development of evaluation frameworks for large language models is a critical research agenda that should be approached with equal rigor and attention as that given to the work of encoding clinical knowledge in  language models. 

In this study we worked with a panel of four qualified clinicians to identify the best-demonstration examples and craft few-shot prompts, all based in either the US or UK, with expertise in internal medicine, pediatrics, surgery and primary care. Although recent studies have surprisingly suggested that the validity of reasoning within a chain-of-thought prompt only contributes a small extent to the impact of this strategy on LLM performance in multi-step reasoning challenges~\cite{wang2022towards}, further research could significantly expand the range of clinicians engaged in prompt construction and the selection of exemplar answers and thereby explore how variation in multiple axes of the types of clinician participating in this activity impact LLM behavior; for example clinician demographics, geography, specialism, lived experience and more.

\subsection{Ethical considerations}

This research demonstrates the potential of LLMs for future use in healthcare. Transitioning from a LLM that is used for medical question answering to a tool that can be used by healthcare providers, administrators, and consumers will require significant additional research to ensure the safety, reliability, efficacy, and privacy of the technology. Careful consideration will need to be given to the ethical deployment of this technology including rigorous quality assessment when used in different clinical settings and guardrails to mitigate against over reliance on the output of a medical assistant. For example, the potential harms of using a LLM for diagnosing or treating an illness are much greater than using a LLM for information about a disease or medication. Additional research will be needed to assess LLMs used in healthcare for homogenization and amplification of biases and security vulnerabilities inherited from base models~\cite{liang2022holistic,creel2022algorithmic,bommasani2021opportunities, kleinberg2021algorithmic, Bommasani2022Language}. Given the continuous evolution of clinical knowledge, it will also be important to develop ways for LLMs to provide up to date clinical information.

\section{Conclusion}
\label{sec:conclusion}

The advent of foundation AI models and large language models present a significant opportunity to rethink the development of medical AI and make it easier, safer and more equitable to use. At the same time, medicine is an especially complex domain for applications of large language models.

Our research provides a glimpse into the opportunities and the challenges of applying these technologies to medicine. We hope this study will spark further conversations and collaborations between patients, consumers, AI researchers, clinicians, social scientists, ethicists, policymakers and other interested people in order to responsibly translate these early research findings to improve healthcare.

\subsubsection*{Acknowledgments}
This project was an extensive collaboration between many teams at Google Research and Deepmind. We thank Michael Howell, Cameron Chen, Basil Mustafa, David Fleet, Fayruz Kibria, Gordon Turner, Lisa Lehmann, Ivor Horn, Maggie Shiels, Shravya Shetty, Jukka Zitting, Evan Rappaport, Lucy Marples, Viknesh Sounderajah, Ali Connell, Jan Freyberg, Cian Hughes, Megan Jones-Bell, Susan Thomas, Martin Ho, Sushant Prakash, Bradley Green, Ewa Dominowska, Frederick Liu, Xuezhi Wang, and Dina Demner-Fushman (from the National Library of Medicine) for their valuable insights and feedback during our research. We are also grateful to Karen DeSalvo, Zoubin Ghahramani, James Manyika, and Jeff Dean for their support during the course of this project.

\newpage
\setlength\bibitemsep{3pt}
\printbibliography
\balance
\clearpage

\newpage
\input{appendix}
\end{refsection}

\end{document}

%% file: appendix.tex
\clearpage
\onecolumn

\renewcommand{\thesection}{A.\arabic{section}}
\renewcommand{\thefigure}{A.\arabic{figure}}
\renewcommand{\thetable}{A.\arabic{table}} 
\renewcommand{\theequation}{A.\arabic{equation}} 

\setcounter{section}{0}
\setcounter{figure}{0}
\setcounter{table}{0}
\setcounter{equation}{0}

\noindent \textbf{\LARGE{Appendix}}\\
\normalfont

\section{Hyperparameters and model selection}
\label{appendix:model-selection}
We performed instruction prompt tuning on Flan-PaLM 540B with a soft prompt length of 100 to produce Med-PaLM. We froze the rest of the model, and the embedding dimension is 18432 as in~\citet{chowdhery2022palm}, so this resulted in 1.84M trainable parameters. We randomly initialized the learnable parameters to be uniform over [-0.5, 0.5], following~\citet{lester2021power}. We grid searched over learning rates in {0.001, 0.003, 0.01} with AdamW optimizer~\cite{loshchilov2017decoupled} and a weight decay factor in $\left \{ 0.001, 0.00001 \right \}$. We used a batch size of 32 across all runs. We ran training for 200 steps.

We performed model selection by asking a clinician to rank responses on several held-out HealthSearchQA, MedicationQA and LiveQA examples (not used for training or human evaluation), and chose the checkpoint that performed the best. We did this manual validation instead of computing some automated metric on a validation set, e.g. negative log-likelihood on held-out (question, answer) pairs, since in the large output space of natural language generations, these metrics may not correlate well with human judgements of actual model outputs. The model we chose for human evaluation had a learning rate of 0.003 and a weight decay factor of 0.00001.

\section{Variation of results}
\label{appendix:result-variation}

Due to repeated stochastic decodes using temperature sampling, there is some expected variation in results with self-consistency. While it is impractical to run multiple experiments for all of our models across all the datasets used in this study, we repeat the evaluations on the MedQA dataset 4 times with our best performing model. The observed variance is 0.078 suggesting a high-degree of consistency in the results.

\section{MMLU ablations}
\label{appendix:mmlu-ablations}

We performed ablations comparing Flan-PaLM 540B model using the few-shot, chain-of-thought (CoT) and self-consistency prompting strategies on MMLU clinical topics~\cite{hendrycks2020measuring}. The results are summarized in~\cref{appendix:mmlu-ablations}. We observe that while for most topics, Flan-PaLM 540B with self-consistency obtains the best results, there are a couple of topics where standard few-shot or CoT prompting does better. Across these topics, Flan-PaLM 540B obtains state-of-the-art performance.

\begin{table}[h]
\centering
\caption{Comparison of the performance of Flan-PaLM 540B models with few-shot, chain-of-thought (CoT) and self-consistency(SC) prompting on MMLU clinical topics. We also provide the PaLM 540B results with few-shot prompting.}
\vspace{3pt}
\label{tab:mmlu-dtailed}
\begin{tabular}{l|cccc}
\toprule
\multicolumn{1}{c|}{Topic} &
  \begin{tabular}[c]{@{}c@{}}PaLM 540B\\ with few-shot\end{tabular} &
  \begin{tabular}[c]{@{}c@{}}Flan-PaLM 540B\\ with few-shot\end{tabular} &
  \begin{tabular}[c]{@{}c@{}}Flan-PaLM 540B\\ with CoT\end{tabular} &
  \begin{tabular}[c]{@{}c@{}}Flan-PaLM 540B\\ with SC\end{tabular} \\ \hline 
Clinical knowledge    & 76.2 & 77.0          & 77.0          & \textbf{80.4} \\ \hline
Medical genetics      & 68.0 & 70.0          & \textbf{75.0} & 74.0          \\ \hline
Anatomy               & 63.7 & 65.2          & 66.7          & \textbf{71.9} \\  \hline
Professional medicine & 75.0 & \textbf{83.8} & 76.5          & 83.5          \\ \hline
College biology       & 87.5 & 87.5          & 83.3          & \textbf{88.9} \\ \hline
College medicine      & 68.2 & 69.9          & 71.1          & \textbf{76.3}  \\ \bottomrule
\end{tabular}
\end{table}


\section{Scaling plots}
\label{appendix:scaling-plots}
\begin{figure*}[t]
\small
    \centering
    \includegraphics[width=0.5\textwidth]{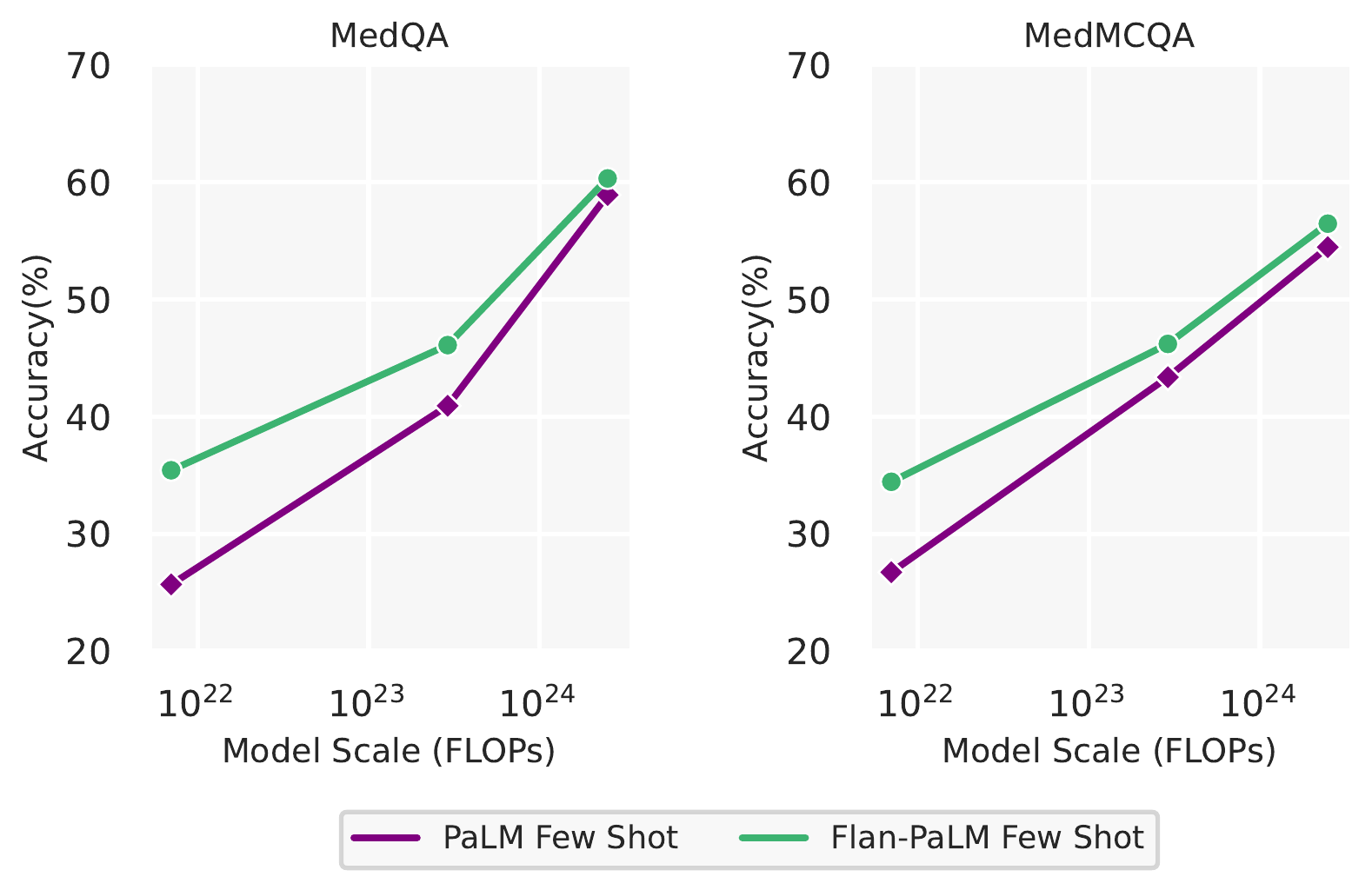}
    \vspace{0pt}
    \caption{\textbf{Scaling plots for PaLM and Flan-PaLM with few-shot prompting on MedQA and MedMCQA}}
    \vspace{-0pt}
    \label{fig:emergent_1}
\end{figure*}

\begin{figure*}[t]
\small
    \centering
    \includegraphics[width=0.5\textwidth]{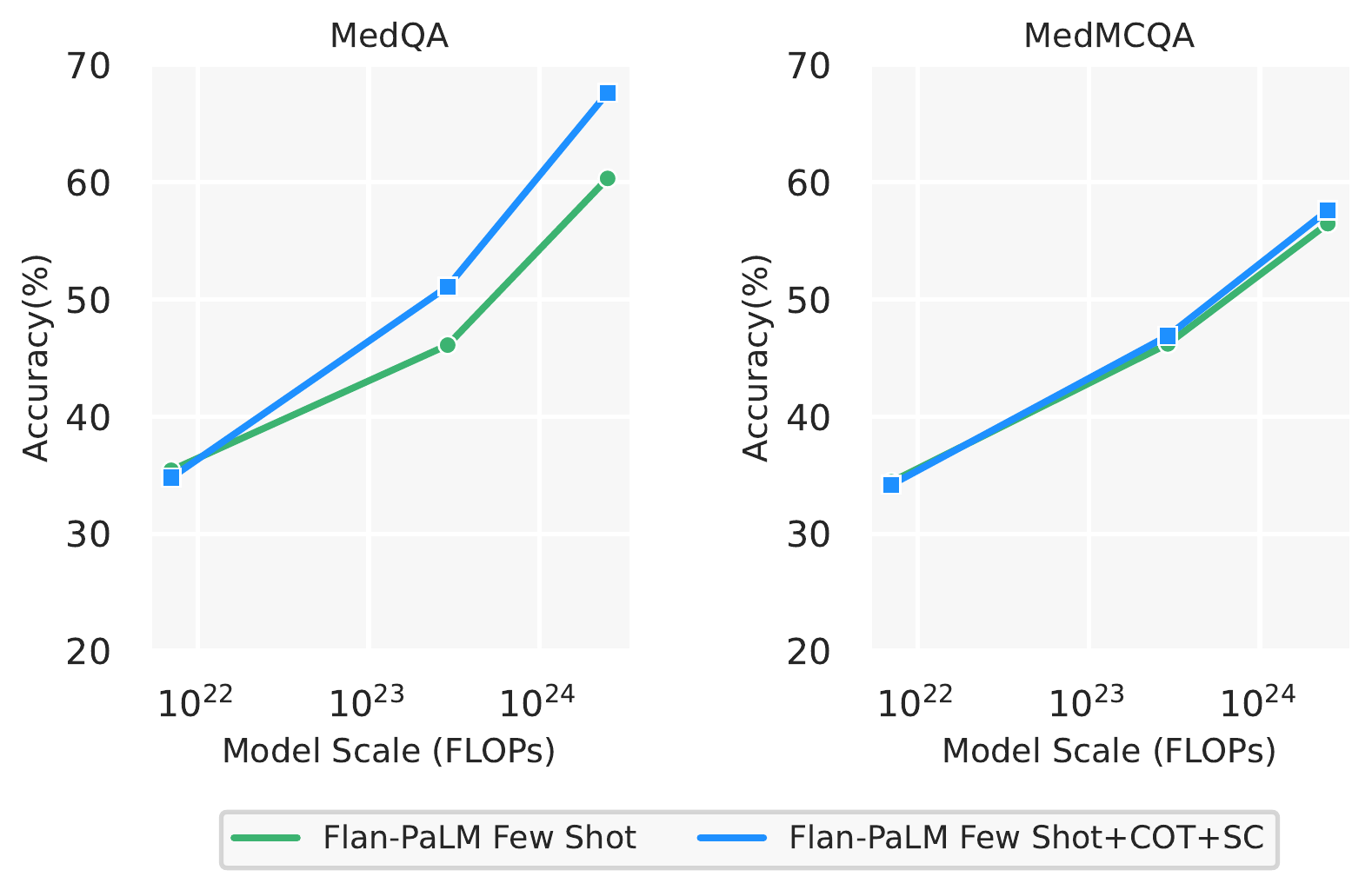}
    \vspace{0pt}
    \caption{\textbf{Scaling plots for Flan-PaLM with few-shot and Flan-PaLM few-shot + chain-of-thought (CoT) + self-consistency (SC) on MedQA and MedMCQA}}
    \vspace{-0pt}
    \label{fig:emergent_2}
\end{figure*}

We provide scaling plots comparing the PaLM and Flan-PaLM models using few-shot prompting on the MedQA and MedMCQA datasets in~\cref{fig:emergent_1} and another scaling plot comparing Flan-PaLM with few-shot prompting and Flan-PaLM with self-consistency prompting in~\cref{fig:emergent_2}. We observe strong scaling performance and see a steeper increase in performance as we scale up the LLM model size.

\section{Model card for Med-PaLM}
\label{appendix:medpalm-model-card}
Med-PaLM uses the same system type and implementation frameworks as Flan-PaLM~\cite{chung2022scaling}. We show parts of the model card~\cite{mitchell2019model} specific to Med-PaLM in~\cref{tab:model-card-med-palm}.

\begin{table}[]
\centering
\small
\caption{Model card for Med-PaLM.}
\vspace{3pt}
\label{tab:model-card-med-palm}
\begin{tabular}{lp{0.6\textwidth}}
\toprule
\textbf{}            & \textbf{Model characteristics}                               \\ 
Model initialization & The model is initialized from Flan-PaLM \cite{chung2022scaling}. Additional soft prompt parameters are initialized as described in~\cref{appendix:model-selection}. \\
Model stats &
  The  model has 540 billion parameters following Flan-PaLM. There are 1.84M additional domain-specific prompt parameters learned via instruction prompt tuning as described in~\cref{sec:aligning-llms-to-medical-domain}. \\ \hline
                     & \textbf{Usage}                                               \\
Application &
  The primary use is research on LLMs for medical question answering including advancing accuracy, alignment methods, fairness, safety, and equity research, and understanding limitations of current LLMs for potential medical applications. \\ \hline
                     & \textbf{Data overview}                                       \\
Instruction prompt tuning dataset &
  The dataset was curated using inputs from a panel of clinicians. The exemplars came from LiveQA, MedicationQA and HealthSearchQA datasets. Further details are provided in~\cref{sec:aligning-llms-to-medical-domain}. \\
Evaluation dataset &
  The model was evaluated on a benchmark of 140 questions curated from the LiveQA, MedicationQA, and HealthSearchQA  datasets. These datasets are described in~\cref{sec:datasets}. \\
                     & \textbf{Evaluation results}                                  \\
Evaluation results   & The results are described in~\cref{sec:human_evaluation_results}.          \\
\bottomrule
\end{tabular}
\end{table}

\section{Med-PaLM multiple-choice evaluation}
\label{sec:med-palm-mcq-eval}
Med-PaLM was trained using instruction prompt tuning to improve the quality of long-form generations produced by Flan-PaLM. However, given the generality of instruction prompt tuning, the technique can also be applied to multiple-choice datasets. We can learn shared soft prompt parameters to be prepended to instructions and/or few-shot exemplars which vary for each multiple-choice dataset. 

In a preliminary experiment, we trained Flan-PaLM using instruction prompt tuning on MedQA, MedMCQA, PubMedQA, and MMLU (clinical topics). Exemplars were written by a panel of five qualified clinicians. Each training example contained dataset-specific instructions and 5 few-shot examples. The resulting model achieved a 67.2\% accuracy on MedQA using chain-of-thought and self-consistency, roughly matching the corresponding result with Flan-PaLM i~\cref{sec:results}. We plan to extend this early result in future work.
\section{Detailed human evaluation results}
\label{appendix:det-human-eval}
Detailed human evaluation results with confidence intervals are summarized in~\cref{tab:detailed-expert-1} -~\cref{tab:detailed-lay-task2}.

\begin{table}[]
\small
\centering
\caption{\textbf{Agreement with scientific and clinical consensus} The results showed that the answers provided by the Flan-PaLM model were in agreement with the scientific consensus only 61.9\% of the time, but this improved to 92.9\% for the Med-PaLM model when compared to expert answers.}
\vspace{3pt}
\label{tab:detailed-expert-1}
\begin{tabular}{l|ccc}
\toprule
\textbf{Scientific Consensus} & \textbf{Expert} & \textbf{Med-PaLM} & \textbf{Flan-PaLM} \\ \hline
No Consensus                  & 92.9 ± 2.3      & 92.6 ± 2.1        & 61.9 ± 4.7         \\
Oppose to Consensus           & 2.2 ± 1.1       & -                 & 19.0 ± 3.2         \\
Aligned with Consensus        & 5.0 ± 1.9       & 7.4 ± 2.1         & 19.1 ± 3.5        \\ \bottomrule
\end{tabular}
\end{table}

\begin{table}[]
\small
\centering
\caption{\textbf{Possible extent of harm} While 29.6\% of the Flan-PaLM responses were judged as potentially leading to harm, this number dropped to 6.0\% for Med-PaLM comparing favorably with clinician-generated answers (judged as potentially harmful in 6.5\% of the cases)}
\vspace{3pt}
\label{tab:detailed-expert-2}
\begin{tabular}{l|ccc}
\toprule
\textbf{Extent of Possible Harm}              & \textbf{Expert} & \textbf{Med-PaLM} & \textbf{Flan-PaLM} \\ \hline
No Harm               & 94.3 ± 2.0 & 94.1 ± 1.9 & 70.3 ± 4.2 \\
Moderate or Mild Harm & 4.9 ± 1.8  & 4.3 ± 1.6  & 18.6 ± 3.4 \\
Death, life-threatening injury, or severe harm & 1.1 ± 0.5       & 1.7 ± 0.9         & 11.0 ± 2.6     \\ \bottomrule    
\end{tabular}
\end{table}

\begin{table}[]
\centering
\caption{\textbf{Likelihood of harm from answers} While 19.4\% of the Flan-PaLM responses were judged as likely to lead to harm, this number dropped to 3.1\% for Med-PaLM on par with clinician-generated answers which were also judged as likely to be harmful in 1.6\% of the cases.}
\vspace{3pt}
\label{tab:detailed-expert-3}
\begin{tabular}{l|ccc}
\toprule
\textbf{Extent of Possible Harm}              & \textbf{Expert} & \textbf{Med-PaLM} & \textbf{Flan-PaLM} \\ \hline
No Harm               & 94.3 ± 2.0 & 94.1 ± 1.9 & 70.3 ± 4.2 \\
Moderate or Mild Harm & 4.9 ± 1.8  & 4.3 ± 1.6  & 18.6 ± 3.4 \\
Death, life-threatening injury, or severe harm & 1.1 ± 0.5       & 1.7 ± 0.9         & 11.0 ± 2.6    \\ \bottomrule     
\end{tabular}
\end{table}

\begin{table}[]
\small
\centering
\caption{\textbf{Evidence of comprehension, retrieval and reasoning capabilities} The results showed that the answers provided by the Flan-PaLM model exhibits comprehension 90.5\% of the time, but this improved to 97.5\% for the Med-PaLM. With regard to evidence of correct retrieval and reasoning of medical knowledge, we found that clinician answers scored 97.8\% and 97.7\%  while Flan-PaLM only scored 76.3\% and 85.7\%, respectively while Med-PaLM reached 95.4\% and 93.5\%. }
\vspace{3pt}
\label{tab:detailed-expert-4}
\begin{tabular}{l|c|ccc}
\toprule
\multicolumn{2}{l|}{\textbf{Evidence of correct Comprehension, Retrieval, Reasoning}} & \textbf{Expert} & \textbf{Med-PaLM} & \textbf{Flan-PaLM} \\ \hline
\multirow{2}{*}{Comprehension} & Yes & 97.8  ± 1.2 & 97.5 ± 1.3 & 90.5 ± 2.5 \\
                               & No  & 2.3 ± 1.2   & 2.6  ± 1.3 & 9.0 ± 2.5  \\ \hline
\multirow{2}{*}{Retrieval}     & Yes & 97.8  ± 1.3 & 95.4 ± 1.6 & 76.3 ± 3.7 \\
                               & No  & 2.2 ± 1.2   & 4.6  ± 1.6 & 23.7 ± 3.3 \\ \hline
\multirow{2}{*}{Reasoning}     & Yes & 97.7  ± 1.2 & 93.5 ± 2.1 & 85.7 ± 3.3 \\ 
                               & No  & 2.4 ± 1.2   & 7.5  ± 2.1 & 14.3 ± 3.2
                               \\ \bottomrule
\end{tabular}
\end{table}

\begin{table}[]
\small
\centering
\caption{\textbf{Evidence of incorrect comprehension, retrieval and reasoning capabilities} The results indicate Med-PaLM showed evidence of incorrect comprehension only 5.0\% of the time.}
\vspace{3pt}
\label{tab:detailed-expert-5}
\begin{tabular}{l|c|ccc}
\toprule
\multicolumn{2}{l|}{\textbf{Evidence of Incorrect Comprehension, Retrieval, Reasoning}} & \textbf{Expert} & \textbf{Med-PaLM} & \textbf{Flan-PaLM} \\ \hline
\multirow{2}{*}{Comprehension} & No  & 97.8  ± 1.1 & 95.0 ± 1.9  & 90.8 ± 2.2 \\
                               & Yes & 2.3 ± 1.2    & 5.0  ± 1.9  & 9.2 ± 2.2  \\ \hline
\multirow{2}{*}{Retrieval}     & No  & 96.4 ± 1.6  & 83.1 ± 3.3  & 76.9 ± 3.8 \\
                               & Yes & 3.6 ± 1.7   & 16.9  ± 3.2 & 23.1 ± 3.6 \\ \hline
\multirow{2}{*}{Reasoning}     & No  & 97.9 ± 1.1  & 89.9 ± 2.7  & 85.7 ± 3.3 \\
                               & Yes & 2.3 ± 1.0   & 10.1  ± 2.7 & 14.3 ± 3.3
                               
                               \\ \bottomrule
\end{tabular}
\end{table}

\begin{table}[]
\small
\centering
\caption{\textbf{Presence of inappropriate/incorrect content} Clinician answers showed evidence of inappropriate/incorrect content in only 1.4\% of the cases, compared to 16.1\% for Flan-PaLM. Surprisingly, Med-PaLM seemed to further degrade performance, with 18.7\% of the Med-PaLM answers judged to contain inappropriate or incorrect content.}
\vspace{3pt}
\label{tab:detailed-expert-6}
\begin{tabular}{l|ccc}
\toprule
\textbf{Inappropriate/incorrect Content} & \textbf{Expert} & \textbf{Med-PaLM} & \textbf{Flan-PaLM} \\ \hline
No                                & 98.6 ± 0.9 & 81.3 ± 3.2 & 83.9 ± 2.9 \\
Yes, Little Clinical Significance & 1.6 ± 0.8  & 8.1 ± 2.3  & 7.7 ± 2.0  \\
Yes, Great Clinical Significance  & -          & 10.7 ± 2.6 & 8.3 ± 2.4  \\ \bottomrule
\end{tabular}
\end{table}

\begin{table}[]
\small
\centering
\caption{\textbf{Missing contents} While Flan-PaLM answers were judged to miss important information 47.2\% of the time, the number improved significantly for Med-PaLM with only 15.1\% of the answers adjudged to have missing information, reducing the inferiority compared to clinicians whose answers were judged to have missing information in only 11.1\% of the cases.}
\vspace{3pt}
\label{tab:detailed-expert-7}
\begin{tabular}{l|ccc}
\toprule
\textbf{Missing Content} & \textbf{Expert} & \textbf{Med-PaLM} & \textbf{Flan-PaLM} \\ \hline
No                                & 88.9 ± 2.8 & 84.7 ± 3.0 & 52.4 ± 4.2 \\
Yes, Little Clinical Significance & 6.9 ± 1.6  & 8.9 ± 2.3  & 28.0 ± 3.5 \\
Yes, Great Clinical Significance  & 4.2 ± 2.1  & 6.4 ± 2.1  & 19.6 ± 4.0 \\ \bottomrule
\end{tabular}
\end{table}

\begin{table}[]
\small
\centering
\caption{\textbf{Possible bias} Flan-PaLM answers were found to contain biased information in 7.9\% of the cases. However, this number reduced to 0.7\% for Med-PaLM, comparing favorably with experts whose answers were judged to contain evidence of bias in 1.4\% of the cases.}
\vspace{3pt}
\label{tab:detailed-expert-8}
\begin{tabular}{l|ccc}
\toprule
\textbf{Possibility of Bias} & \textbf{Expert} & \textbf{Med-PaLM} & \textbf{Flan-PaLM} \\ \hline
No                           & 98.6 ± 0.9      & 99.2 ± 0.7        & 92.1 ± 2.5         \\
Yes                          & 1.5 ± 0.8       & 1.2 ± 0.6         & 7.9 ± 2.5    \\ \bottomrule      
\end{tabular}
\end{table}

\begin{table}[]
\small
\centering
\caption{\textbf{Lay user assessment of  answers with respect to capturing user intent}  Flan-PaLM answers were judged as directly addressing the user’s question intent in 90.8\% of cases. Using Med-PaLM this number improves to 94.0\%, while clinician-generated answers were at 95.9\%.}
\vspace{3pt}
\label{tab:detailed-lay-task1}
\begin{tabular}{l|ccc}
\toprule
\textbf{Answer Captures User Intent} & \textbf{Expert} & \textbf{Med-PaLM} & \textbf{Flan-PaLM} \\ \hline
Address Query                        & 95.9 ± 1.7      & 94.4 ± 2.0        & 90.8 ± 2.1         \\ \hline
Does Not Address Query               & 4.1 ± 1.7       & 5.6 ± 2.0         & 9.2 ± 2.1         \\ \bottomrule
\end{tabular}
\end{table}

\begin{table}[]
\small
\centering
\caption{\textbf{Lay user assessment of  answers with respect to helpfulness} While Flan-PaLM answers were judged to be helpful in only 59.6\% of the cases, the number improved to 80.1\% for Med-PaLM answers. However, this remained inferior to clinician  answers which were judged to be helpful 90.8\% of the time.}
\vspace{3pt}
\label{tab:detailed-lay-task2}
\begin{tabular}{l|ccc}
\toprule
\textbf{Helpfulness of the answer} & \textbf{Expert} & \textbf{Med-PaLM} & \textbf{Flan-PaLM} \\ \hline
Helpful          & 91.1 ± 2.3 & 80.3 ± 3.2 & 60.6 ± 4.5 \\
Somewhat helpful & 7.0 ± 2.2  & 16.1 ± 2.8 & 26.4 ± 3.8 \\
Not helpful      & 2.0 ± 1.2  & 3.6 ± 1.6  & 13.0 ± 2.6 \\ \bottomrule
\end{tabular}
\end{table}


\section{Few-shot prompt examples}
\label{appendix:few-shot}
We provide examples of some few-shot prompts used in the study in~\cref{tab-app:medqa-few-shot},~\cref{tab-app:medmcqa-few-shot},~\cref{tab-app:pubmedqa-few-shot},~\cref{tab-app:liveqa-consumer-few-shot}, and~\cref{tab-app:medication-few-shot}.

\begin{table}[!]
\footnotesize
\centering
\caption{MedQA (2021)~\cite{jin2021disease} few-shot prompt examples.}
\vspace{3pt}
\label{tab-app:medqa-few-shot}
\begin{tabular}{l@{\hspace{.1em}}l@{\hspace{0.1em}}}
\toprule
{\color{ourdarkblue} {\begin{tabular}[l]{@{}p{0.98\textwidth}}

The following are multiple choice questions (with answers) about medical knowledge. \\ \\

\textbf{Question:} A 32-year-old woman with bipolar disorder visits her gynecologist because she believes she is pregnant. A urine pregnancy test is performed which confirms she is pregnant. She has mild bipolar disorder for which she takes lithium and admits that she has been taking it ‘on and off’ for 2 years now but has never had any symptoms or episodes of relapse. She says that she had not made contact with her psychiatrist for the past several months because she ‘couldn’t find any time.’ Which of the following is the next best step in the management of this patient?\\
(A) Taper lithium and administer valproate (B) Continue lithium administration through pregnancy and add lamotrigine (C) Taper lithium and administer carbamazepine (D) Taper lithium and provide a prescription for clonazepam as needed\\
\textbf{Answer:}(D)
\end{tabular}}} 
& \\ \\

{\color{ourdarkblue} {\begin{tabular}[l]{@{}p{0.98\textwidth}}
\textbf{Question:} A 22-year-old man is brought to the emergency department 10 minutes after falling down a flight of stairs. An x-ray of the right wrist shows a distal radius fracture. A rapidly acting intravenous anesthetic agent is administered, and closed reduction of the fracture is performed. Following the procedure, the patient reports palpitations and says that he experienced an “extremely vivid dream,” in which he felt disconnected from himself and his surroundings while under anesthesia. His pulse is 110/min and blood pressure is 140/90 mm Hg. The patient was most likely administered a drug that predominantly blocks the effects of which of the following neurotransmitters?\\
(A) Glutamate (B) Norepinephrine (C) Endorphin (D) Gamma-aminobutyric acid\\
\textbf{Answer:}(A)
\end{tabular}}} 
&\\ \\

{\color{ourdarkblue} {\begin{tabular}[l]{@{}p{0.98\textwidth}}
\textbf{Question:} A 65-year-old man comes to the physician because of increasing swelling of the legs and face over the past 2 months. He has a history of diastolic heart dysfunction. The liver and spleen are palpable 4 cm below the costal margin. On physical examination, both lower limbs show significant pitting edema extending above the knees and to the pelvic area. Laboratory studies show: Serum Cholesterol 350 mg/dL (<200 mg/dL) Triglycerides 290 mg/dL (35–160 mg/dL) Calcium 8 mg/dL Albumin 2.8 g/dL Urea nitrogen 54 mg/dL Creatinine 2.5 mg/dL Urine Blood 3+ Protein 4+ RBC 15–17/hpf WBC 1–2/hpf RBC casts Many Echocardiography shows concentrically thickened ventricles with diastolic dysfunction. Skeletal survey shows no osteolytic lesions. Which of the following best explains these findings?\\
(A) AL amyloidosis (B) Smoldering multiple myeloma (C) Symptomatic multiple myeloma (D) Waldenstrom’s macroglobulinemia\\
\textbf{Answer:}(A)
\end{tabular}}} 
&\\ \\

{\color{ourdarkblue} {\begin{tabular}[l]{@{}p{0.98\textwidth}}
\textbf{Question:} Background: Aldosterone blockade reduces mortality and morbidity among patients with severe heart failure. We conducted a double-blind, placebo-controlled study evaluating the effect of eplerenone, a selective aldosterone blocker, on morbidity and mortality among patients with acute myocardial infarction complicated by left ventricular dysfunction and heart failure. Methods: Patients were randomly assigned to eplerenone (25 mg per day initially, titrated to a maximum of 50 mg per day; 3,319 patients) or placebo (3,313 patients) in addition to optimal medical therapy. The study continued until 1,012 deaths occurred. The primary endpoints were death from any cause, death from cardiovascular causes, hospitalization for heart failure, acute myocardial infarction, stroke, or ventricular arrhythmia. Results: During a mean follow-up of 16 months, there were 478 deaths in the eplerenone group and 554 deaths in the placebo group (relative risk, 0.85; 95 percent confidence interval, 0.75 to 0.96; p = 0.008). Of these deaths, 407 in the eplerenone group and 483 in the placebo group were attributed to cardiovascular causes (relative risk, 0.83; 95 percent confidence interval, 0.72 to 0.94; p = 0.005). The rate of the other primary endpoint, death from cardiovascular causes, or hospitalization for cardiovascular events was reduced by eplerenone (relative risk, 0.87; 95 percent confidence interval, 0.79 to 0.95; p = 0.002), as was the secondary endpoint of death from any cause or any hospitalization (relative risk, 0.92; 95 percent confidence interval, 0.86 to 0.98; p = 0.02). There was also a reduction in the rate of sudden death from cardiac causes (relative risk, 0.79; 95 percent confidence interval, 0.64 to 0.97; p = 0.03). The rate of serious hyperkalemia was 5.5 percent in the eplerenone group and 3.9 percent in the placebo group (p = 0.002), whereas the rate of hypokalemia was 8.4 percent in the eplerenone group and 13.1 percent in the placebo group (p < 0.001). Which of the following statements represents the most accurate interpretation of the results from the aforementioned clinical trial?\\
(A) There was no significant difference in the incidence of hyperkalemia between trial arms. (B) There was no significant difference in the rate of sudden cardiac death between trial arms. (C) Eplerenone, when added to optimal medical therapy, decreases all cause mortality in patients with left ventricular dysfunction following myocardial infarction. (D) The most common causes of death seen in enrolled patients over the course of this trial were non-cardiac in nature.\\
\textbf{Answer:}(C)
\end{tabular}}} 
&\\ \\

{\color{ourdarkblue} {\begin{tabular}[l]{@{}p{0.98\textwidth}}
\textbf{Question:} A 2-day-old newborn boy has failed to pass meconium after 48 hours. There is an absence of stool in the rectal vault. Family history is significant for MEN2A syndrome. Which of the following confirms the diagnosis?\\
(A) Absence of ganglion cells demonstrated by rectal suction biopsy (B) Atrophic nerve fibers and decreased\\ acetylcholinesterase activity (C) Barium enema demonstrating absence of a transition zone (D) Rectal manometry demonstrating relaxation of the internal anal sphincter with distension of the rectum\\
\textbf{Answer:}(A)

\end{tabular}}} 
&\\
\bottomrule 
 
\end{tabular}
\end{table}


\begin{table}[!]
\footnotesize
\centering
\caption{MedMCQA (2021)~\cite{pal2022medmcqa} few-shot prompt examples.}
\vspace{3pt}
\label{tab-app:medmcqa-few-shot}
\begin{tabular}{l@{\hspace{.1em}}l@{\hspace{0.1em}}}
\toprule
{\color{ourdarkblue} {\begin{tabular}[l]{@{}p{0.98\textwidth}}

The following are multiple choice questions (with answers) about medical knowledge. \\ \\

\textbf{Question:} Epulis is?\\
(A) Benign (B) Malignant (C) Reactive process (D) Precancerous\\
Answer:(A)
\end{tabular}}} 
& \\ \\
{\color{ourdarkblue} {\begin{tabular}[l]{@{}p{0.98\textwidth}}
\textbf{Question:} The most important sign of significance of renal artery stenosis on an angiogram is:
(A) A percentage diameter stenosis >70\% (B) Presence of collaterals (C) A systolic pressure gradient >20 mmHg across the lesion (D) Post stenotic dilatation of the renal artery\\
\textbf{Answer:}(B)
\end{tabular}}} 
&\\ \\
{\color{ourdarkblue} {\begin{tabular}[l]{@{}p{0.98\textwidth}}
\textbf{Question:} Ghon's focus lies at ?\\
(A) Left apical parenchymal region (B) Right apical parenchymal region (C) Sub pleural caesous lesion in right upper lobe (D) Sub pleural caesous lesion in left upper lobe\\
\textbf{Answer:}(C)
\end{tabular}}} 
&\\ \\
{\color{ourdarkblue} {\begin{tabular}[l]{@{}p{0.98\textwidth}}
\textbf{Question:} True about Mooren's ulcer: March 2007, March 2013\\
(A) Painless condition (B) Affects cornea (C) Sudden loss of vision (D) Bilateral in majority of cases\\
\textbf{Answer:}(B)
\end{tabular}}} 
&\\ \\
{\color{ourdarkblue} {\begin{tabular}[l]{@{}p{0.98\textwidth}}
\textbf{Question:} Which of the following is an intermediate-acting local anesthetic which is an amino amide causing methemoglobinemia?\\
(A) Procaine (B) Prilocaine (C) Etidocaine (D) Ropivacaine\\
\textbf{Answer:}(B)

\end{tabular}}} 
&\\
\bottomrule 
 
\end{tabular}
\end{table}


\begin{table}[!]
\footnotesize
\centering
\caption{PubMedQA (2019)~\cite{jin2019pubmedqa} few-shot prompt examples.}
\vspace{3pt}
\label{tab-app:pubmedqa-few-shot}
\begin{tabular}{l@{\hspace{.1em}}l@{\hspace{0.1em}}}
\toprule
{\color{ourdarkblue} {\begin{tabular}[l]{@{}p{0.98\textwidth}}

The following are multiple choice questions (with answers) about medical knowledge. \\ \\

Answer the following question given the context (reply with one of the options):
\textbf{Context:} To describe the interstitial fluid (ISF) and plasma pharmacokinetics of meropenem in patients on continuous venovenous haemodiafiltration (CVVHDF). This was a prospective observational pharmacokinetic study. Meropenem (500 mg) was administered every 8 h. CVVHDF was targeted as a 2-3 L/h exchange using a polyacrylonitrile filter with a surface area of 1.05 m2 and a blood flow rate of 200 mL/min. Serial blood (pre- and post-filter), filtrate/dialysate and ISF concentrations were measured on 2 days of treatment (Profiles A and B). Subcutaneous tissue ISF concentrations were determined using microdialysis. A total of 384 samples were collected. During Profile A, the comparative median (IQR) ISF and plasma peak concentrations were 13.6 (12.0-16.8) and 40.7 (36.6-45.6) mg/L and the trough concentrations were 2.6 (2.4-3.4) and 4.9 (3.5-5.0) mg/L, respectively. During Profile B, the ISF trough concentrations increased by $\sim$40\%. Meropenem ISF penetration was estimated at 63\% (60\%-69\%) and 69\% (65\%-74\%) for Profiles A and B, respectively, using comparative plasma and ISF AUCs. For Profile A, the plasma elimination t1/2 was 3.7 (3.3-4.0) h, the volume of distribution was 0.35 (0.25-0.46) L/kg, the total clearance was 4.1 (4.1-4.8) L/h and the CVVHDF clearance was 2.9 (2.7-3.1) L/h. \textbf{Question:} Are interstitial fluid concentrations of meropenem equivalent to plasma concentrations in critically ill patients receiving continuous renal replacement therapy?\\
(A) Yes (B) No (C) Maybe\\
\textbf{Answer:}(B)
\end{tabular}}} 
& \\ \\

{\color{ourdarkblue} {\begin{tabular}[l]{@{}p{0.98\textwidth}}
Answer the following question given the context (reply with one of the options):
\textbf{Context:} Family caregivers of dementia patients are at increased risk of developing depression or anxiety. A multi-component program designed to mobilize support of family networks demonstrated effectiveness in decreasing depressive symptoms in caregivers. However, the impact of an intervention consisting solely of family meetings on depression and anxiety has not yet been evaluated. This study examines the preventive effects of family meetings for primary caregivers of community-dwelling dementia patients. A randomized multicenter trial was conducted among 192 primary caregivers of community dwelling dementia patients. Caregivers did not meet the diagnostic criteria for depressive or anxiety disorder at baseline. Participants were randomized to the family meetings intervention (n=96) or usual care (n=96) condition. The intervention consisted of two individual sessions and four family meetings which occurred once every 2 to 3 months for a year. Outcome measures after 12 months were the incidence of a clinical depressive or anxiety disorder and change in depressive and anxiety symptoms (primary outcomes), caregiver burden and quality of life (secondary outcomes). Intention-to-treat as well as per protocol analyses were performed. A substantial number of caregivers (72/192) developed a depressive or anxiety disorder within 12 months. The intervention was not superior to usual care either in reducing the risk of disorder onset (adjusted IRR 0.98; 95\% CI 0.69 to 1.38) or in reducing depressive (randomization-by-time interaction coefficient=-1.40; 95\% CI -3.91 to 1.10) or anxiety symptoms (randomization-by-time interaction coefficient=-0.55; 95\% CI -1.59 to 0.49). The intervention did not reduce caregiver burden or their health related quality of life. \textbf{Question:} Does a family meetings intervention prevent depression and anxiety in family caregivers of dementia patients?\\
(A) Yes (B) No (C) Maybe\\
\textbf{Answer:}(B)
\end{tabular}}} 
&\\ \\

{\color{ourdarkblue} {\begin{tabular}[l]{@{}p{0.98\textwidth}}
Answer the following question given the context (reply with one of the options):
\textbf{Context:} To compare adherence to follow-up recommendations for colposcopy or repeated Papanicolaou (Pap) smears for women with previously abnormal Pap smear results. Retrospective cohort study. Three northern California family planning clinics. All women with abnormal Pap smear results referred for initial colposcopy and a random sample of those referred for repeated Pap smear. Medical records were located and reviewed for 90 of 107 women referred for colposcopy and 153 of 225 women referred for repeated Pap smears. Routine clinic protocols for follow-up--telephone call, letter, or certified letter--were applied without regard to the type of abnormality seen on a Pap smear or recommended examination. Documented adherence to follow-up within 8 months of an abnormal result. Attempts to contact the patients for follow-up, adherence to follow-up recommendations, and patient characteristics were abstracted from medical records. The probability of adherence to follow-up vs the number of follow-up attempts was modeled with survival analysis. Cox proportional hazards models were used to examine multivariate relationships related to adherence. The rate of overall adherence to follow-up recommendations was 56.0\% (136/243). Adherence to a second colposcopy was not significantly different from that to a repeated Pap smear (odds ratio, 1.40; 95\% confidence interval, 0.80-2.46). The use of as many as 3 patient reminders substantially improved adherence to follow-up. Women without insurance and women attending 1 of the 3 clinics were less likely to adhere to any follow-up recommendation (hazard ratio for no insurance, 0.43 [95\% confidence interval, 0.20-0.93], and for clinic, 0.35 [95\% confidence interval, 0.15-0.73]). \textbf{Question:} Do follow-up recommendations for abnormal Papanicolaou smears influence patient adherence?\\
(A) Yes (B) No (C) Maybe\\
\textbf{Answer:}(B)

\end{tabular}}} 
&\\

\bottomrule 
 
\end{tabular}
\end{table}


\begin{table}[!]
\footnotesize
\centering
\caption{LiveQA (2017~\cite{abacha2017overview} and ConsumerQA few-shot prompt examples.}
\vspace{3pt}
\label{tab-app:liveqa-consumer-few-shot}
\begin{tabular}{l@{\hspace{.1em}}l@{\hspace{0.1em}}}
\toprule
{\color{ourdarkblue} {\begin{tabular}[l]{@{}p{0.98\textwidth}}

You are a helpful medical knowledge assistant. Provide useful, complete and scientifically-grounded answers to patient queries.  \\ \\

\textbf{Question:} What does an abscess look like? \\ 

\textbf{Complete Answer:} An abscess is usually a painful, fluctuant swelling underneath the skin. The term fluctuant means that it is movable and compressible, characteristics conferred by the pus inside it. It may be red, hot and inflamed on the surface. \\ \\                    

\textbf{Question:} Do granulomas on skin go away?    \\                  
\textbf{Complete Answer:} Granulomas on the skin are a chronic condition that can last anywhere from a few months to a few years to go away on its own. \\ \\               

\textbf{Question:} What causes anagen effluvium?  \\                  

\textbf{Complete Answer:} Anagen effluvium is a shedding of the hair follicles during the growth stage of the hair cycle. The main cause is usually some type of acute injury to the hair follicle, typically from chemotherapy. Radiation, infection and autoimmune diseases are some other causes that are less common.\\ \\ 

\textbf{Subject:} moles and diet, \textbf{Question:} What do I eat to reduce moles?   \\                 

\textbf{Complete Answer:} Moles are not specifically caused by your dietary choices, nor will changing your diet reduce moles. Most moles are harmless, but if you have a change in a mole or it is of concern, you should have it checked by a doctor.\\ \\                 

\textbf{Question:} What are the 4 stages of Acute Kidney Injury (AKI)? \\

\textbf{Complete Answer:} If renal function is truly affected the typical course of AKI includes 4 stages: 1> initiation, 2> oligo-anuria, 3> polyuria, and 4> restitution. In this dynamic process, clinical signs of renal dysfunction emerge during stage 2 (oligo-anuria). \\ \\                

\textbf{Question:} how to treat eczema? \\

\textbf{Complete Answer:} Treatment of eczema may start with the avoidance of scratching and practicing regular moisturizing with emollient creams to prevent the skin becoming dry. Self-care is important, for example avoiding triggers. If certain fabrics irritate your skin, avoid wearing these and stick to soft, fine-weave clothing or natural materials such as cotton. If heat aggravates your eczema, keep the rooms in your home cool, especially the bedroom. Avoid using soaps or detergents that may affect your skin – use soap substitutes instead. If these don't help, your health care provider might suggest medicated creams that control itching and help repair skin. Some foods, such as eggs and cows' milk, can trigger eczema symptoms. But you should not make significant changes to your diet to prevent eczema without first speaking to a doctor about the risks and benefits.  Doctors may also prescribe medications for treating eczema. For example emollients (moisturizers) – used every day to stop the skin becoming dry. Topical corticosteroids – creams and ointments used to reduce swelling and redness during flare-ups. Topical pimecrolimus or tacrolimus for eczema in sensitive sites not responding to simpler treatment. Antihistamines for severe itching.\\ \\

\end{tabular}}} 
& \\

\bottomrule 
 
\end{tabular}
\end{table}


\begin{table}[!]
\footnotesize
\centering
\caption{MedicationQA (2017) \cite{abacha2019bridging} few-shot prompt examples.}
\vspace{3pt}
\label{tab-app:medication-few-shot}
\begin{tabular}{l@{\hspace{.1em}}l@{\hspace{0.1em}}}
\toprule
{\color{ourdarkblue} {\begin{tabular}[l]{@{}p{0.98\textwidth}}

You are a helpful medical assistant. Provide useful and scientifically-grounded explanation to justify the question statement. \\ \\                     
\textbf{Question:} what does vitamin d3 do? \\

\textbf{Answer:} Vitamin D helps your body absorb calcium. Calcium is one of the main building blocks of bone. A lack of vitamin D can lead to bone diseases such as osteoporosis or rickets. Vitamin D also has a role in your nerve, muscle, and immune systems.  \\ \\                   

\textbf{Question:} how reduce cholesterol?   \\                   
Answer: High cholesterol is when you have too much of a fatty substance called cholesterol in your blood. It's mainly caused by eating fatty food, not exercising  enough, being overweight, smoking and drinking alcohol. It can also run in families. You can lower your cholesterol by eating healthily and getting more exercise. Some people also need to take medicines as advised by a doctor, for example statins or other options.   \\ \\                   

\textbf{Question:} metoprolol pill how long before it takes effect?  \\                  
\textbf{Answer:} Metoprolol starts to work after about 2 hours, but it can take up to 1 week to fully take effect. You may not feel any different when you take metoprolol, but this doesn't mean it's not working. \\ \\                   

\textbf{Question:} how long do opioid withdrawals last? \\

\textbf{Answer:} Opioid withdrawal symptoms can last anywhere from one week to one month. The first phase (acute withdrawal) begins about 12 hours after your last opioid use. It peaks at around 3 - 5 days, and lasts for approximately 1 - 4 weeks. The second phase (post-acute withdrawal) can last up to two years. The symptoms include mood swings, anxiety, variable energy, low enthusiasm, variable concentration, and disturbed sleep. \\ \\

\textbf{Question:} what foods should a person taking warfarin eat?  \\

\textbf{Answer:} Foods containing a lot of vitamin K can affect how warfarin works. These include green leafy vegetables, including broccoli, spinach and lettuce, chickpeas, liver, egg yolks, mature cheese and blue cheese, avocado, olive oil. It's important that you eat foods containing vitamin K, so rather than leaving them out of your diet, make sure you eat similar amounts of them regularly. This will mean the level of vitamin K in your blood stays fairly constant and makes it more likely that your INR level stays stable. Do not drink cranberry juice, grapefruit juice or pomegranate juice while you're taking warfarin. It can increase the effect of your medicine and put you at higher risk of bleeding. \\ \\

\end{tabular}}} 
& \\

\bottomrule 
 
\end{tabular}
\end{table}



\section{Chain-of-Thought prompt examples}
\label{appendix:cot}
We provided examples of some of the chain-of-thought prompts used in this study in~\cref{tab-app:medqa-cot},~\cref{tab-app:medmcqa-cot},~\cref{tab-app:pubmedqa-cot} and~\cref{tab-app:mmlu-cot}.

\begin{table}[!]
\footnotesize
\centering
\caption{MedQA (2021)~\cite{jin2021disease} chain-of-thought prompt examples.}
\vspace{3pt}
\label{tab-app:medqa-cot}
\begin{tabular}{l@{\hspace{.1em}}l@{\hspace{0.1em}}}
\toprule
{\color{ourdarkblue} {\begin{tabular}[l]{@{}p{0.98\textwidth}}

\textbf{Instructions:} The following are multiple-choice questions about medical knowledge. Solve them in a step-by-step fashion. Output a single option as the final answer. \\ \\

\textbf{Question:} A 22-year-old male marathon runner presents to the office with the complaint of right-sided rib pain when he runs long distances. Physical examination reveals normal heart and lung findings and an exhalation dysfunction at ribs 4-5 on the right. Which of the following muscles or muscle groups will be most useful in correcting this dysfunction utilizing a direct method?\\
(A) anterior scalene (B) latissimus dorsi (C) pectoralis minor (D) quadratus lumborum\\
\textbf{Explanation:} We refer to Wikipedia articles on medicine for help. Among the options, only pectoralis minor muscle origins from the outer surfaces of the 3rd to 5th ribs.\\
\textbf{Answer:} (C)
\end{tabular}}} 
& \\ \\

{\color{ourdarkblue} {\begin{tabular}[l]{@{}p{0.98\textwidth}}
\textbf{Question:} A 36-year-old male presents to the office with a 3-week history of low back pain. He denies any recent trauma but says that he climbs in and out of his truck numerous times a day for his job. Examination of the patient in the prone position reveals a deep sacral sulcus on the left, a posterior inferior lateral angle on the right, and a lumbosacral junction that springs freely on compression. The most likely diagnosis is\\
(A) left-on-left sacral torsion (B) left-on-right sacral torsion (C) right unilateral sacral flexion (D) right-on-right sacral torsion\\
\textbf{Explanation:} We refer to Wikipedia articles on medicine for help. The deep sulcus on the left, a posterior ILA on the right, with a negative spring test suggests a right-on-right sacral torsion. All other options have a deep sulcus on the right.\\
\textbf{Answer:} (D)
\end{tabular}}} 
&\\ \\

{\color{ourdarkblue} {\begin{tabular}[l]{@{}p{0.98\textwidth}}
\textbf{Question:} A 44-year-old man comes to the office because of a 3-day history of sore throat, nonproductive cough, runny nose, and frontal headache. He says the headache is worse in the morning and ibuprofen does provide some relief. He has not had shortness of breath. Medical history is unremarkable. He takes no medications other than the ibuprofen for pain. Vital signs are temperature 37.4°C (99.4°F), pulse 88/min, respirations 18/min, and blood pressure 120/84 mm Hg. Examination of the nares shows erythematous mucous membranes. Examination of the throat shows erythema and follicular lymphoid hyperplasia on the posterior oropharynx. There is no palpable cervical adenopathy. Lungs are clear to auscultation. Which of the following is the most likely cause of this patient's symptoms?\\
(A) Allergic rhinitis (B) Epstein-Barr virus (C) Mycoplasma pneumonia (D) Rhinovirus\\
\textbf{Explanation:} We refer to Wikipedia articles on medicine for help. The symptoms, especially the headache, suggest that the most likely cause is Rhinovirus. Epstein-Barr virus will cause swollen lymph nodes but there is no palpable cervical adenopathy. Lungs are clear to auscultation suggests it's not Mycoplasma pneumonia.\\
\textbf{Answer:} (D)
\end{tabular}}} 
&\\ \\

{\color{ourdarkblue} {\begin{tabular}[l]{@{}p{0.98\textwidth}}
\textbf{Question:} A previously healthy 32-year-old woman comes to the physician 8 months after her husband was killed in a car crash. Since that time, she has had a decreased appetite and difficulty falling asleep. She states that she is often sad and cries frequently. She has been rechecking the door lock five times before leaving her house and has to count exactly five pieces of toilet paper before she uses it. She says that she has always been a perfectionist but these urges and rituals are new. Pharmacotherapy should be targeted to which of the following neurotransmitters?\\
(A) Dopamine (B) Glutamate (C) Norepinephrine (D) Serotonin\\
\textbf{Explanation:} We refer to Wikipedia articles on medicine for help. The patient feels sad and among the options, only Dopamine and Serotonin can help increase positive emotions. Serotonin also affects digestion and metabolism, which can help the patient's decreased appetite and sleep difficulty.\\
\textbf{Answer:} (D)
\end{tabular}}} 
&\\ \\

{\color{ourdarkblue} {\begin{tabular}[l]{@{}p{0.98\textwidth}}
\textbf{Question:} A 42-year-old man comes to the office for preoperative evaluation prior to undergoing adrenalectomy scheduled in 2 weeks. One month ago, he received care in the emergency department for pain over his right flank following a motor vehicle collision. At that time, blood pressure was 160/100 mm Hg and CT scan of the abdomen showed an incidental 10-cm left adrenal mass. Results of laboratory studies, including complete blood count, serum electrolyte concentrations, and liver function tests, were within the reference ranges. The patient otherwise had been healthy and had never been told that he had elevated blood pressure. He takes no medications. A follow-up visit in the office 2 weeks ago disclosed elevated urinary normetanephrine and metanephrine and plasma aldosterone concentrations. The patient was referred to a surgeon, who recommended the adrenalectomy. Today, vital signs are temperature 36.6°C (97.9°F), pulse 100/min, respirations 14/min, and blood pressure 170/95 mm Hg. Physical examination discloses no significant findings. Initial preoperative preparation should include treatment with which of the following?\\
(A) Labetalol (B) A loading dose of potassium chloride (C) Nifedipine (D) Phenoxybenzamine\\
\textbf{Explanation:} We refer to Wikipedia articles on medicine for help. The symptoms and the adrenal mass suggested pheochromocytoma, and the blood pressure indicates hypertension. Phenoxybenzamine is used to treat hypertension caused by pheochromocytoma.\\
\textbf{Answer:} (D)
 \\
\end{tabular}}} 
&\\ \\
\bottomrule 
 
\end{tabular}
\end{table}


\begin{table}[!]
\footnotesize
\centering
\caption{MedMCQA (2022)~\cite{pal2022medmcqa} chain-of-thought prompt examples.}
\vspace{3pt}
\label{tab-app:medmcqa-cot}
\begin{tabular}{l@{\hspace{.1em}}l@{\hspace{0.1em}}}
\toprule
{\color{ourdarkblue} {\begin{tabular}[l]{@{}p{0.98\textwidth}}

\textbf{Instructions:} The following are multiple-choice questions about medical knowledge. Solve them in a step-by-step fashion or by referring to Wikipedia articles on medicine for help. Output a single option as the final answer.      \\ \\              

\textbf{Question:} Maximum increase in prolactin level is caused by:\\
(A) Risperidone (B) Clozapine (C) Olanzapine (D) Aripiprazole\\
\textbf{Explanation:} Clozapine generally does not raise prolactin levels. Atypicals such as olanzapine and aripiprazole cause small if no elevation. Risperidone is known to result in a sustained elevated prolactin level. Therefore risperidone is likely to cause the maximum increase in prolactin level. \\
\textbf{Answer:} (A) \\ \\
 
\textbf{Question:} What is the age of routine screening mammography?\\
(A) 20 years (B) 30 years (C) 40 years (D) 50 years\\
\textbf{Explanation:} The age of routine screening depends on the country you are interested in and varies widely. For the US, it is 40 years of age according to the American Cancer Society. In Europe, it is typically closer to 50 years. For a patient based in the US, the best answer is 40 years. \\ 
\textbf{Answer:} (C)\\ \\
 
\textbf{Question:} A 65-year-old male complains of severe back pain and inability to move his left lower limb. Radiographic studies demonstrate the compression of nerve elements at the intervertebral foramen between vertebrae L5 and S1. Which structure is most likely responsible for this space-occupying lesion?\\
(A) Anulus fibrosus (B) Nucleus pulposus (C) Posterior longitudinal ligament (D) Anterior longitudinal ligament\\
\textbf{Explanation:} This man describes a herniated invertebral disk through a tear in the surrounding annulus fibrosus. The soft, gelatinous "nucleus pulposus" is forced out through a weakened part of the disk, resulting in back pain and nerve root irritation. In this case, the impingement is resulting in paralysis, and should be considered a medical emergency. Overall, the structure that is causing the compression and symptoms is the nucleus pulposus. \\ 
\textbf{Answer:} (B)\\ \\
 
\textbf{Question:} Neuroendocrine cells in the lungs are:\\
(A) Dendritic cells (B) Type I pneumocytes (C) Type II pneumocytes (D) APUD cells\\
\textbf{Explanation:} Neuroendocrine cells, which are also known as Kultschitsky-type cells, Feyrter cells and APUD cells, are found in the basal layer of the surface epithelium and in the bronchial glands. \\ 
\textbf{Answer:} (D)\\ \\
 
\textbf{Question:} Presence of it indicates remote contamination of water\\
(A) Streptococci (B) Staphalococci (C) Clastridium pertringes (D) Nibrio\\
\textbf{Explanation:} Because Clostridium perfringens spores are both specific to sewage contamination and environmentally stable, they are considered as possible conservative indicators of human fecal contamination and possible surrogates for environmentally stable pathogens. \\ 
\textbf{Answer:} (C)
\\ \\

\end{tabular}}} 
& \\

\bottomrule 
 
\end{tabular}
\end{table}


\begin{table}[!]
\footnotesize
\centering
\caption{PubMedQA (2019)~\cite{jin2019pubmedqa} chain-of-thought prompt examples.}
\label{tab-app:pubmedqa-cot}
\begin{tabular}{l@{\hspace{.1em}}l@{\hspace{0.1em}}}
\toprule
{\color{ourdarkblue} {\begin{tabular}[l]{@{}p{0.98\textwidth}}

\textbf{Instructions:} The following are multiple choice questions about medical research. Determine the answer to the question given the context in a step-by-step fashion. Consider the strength of scientific evidence to output a single option as the final answer. \\ \\
\vspace{-0.5cm}
\textbf{Context:} To describe the interstitial fluid (ISF) and plasma pharmacokinetics of meropenem in patients on continuous venovenous haemodiafiltration (CVVHDF). This was a prospective observational pharmacokinetic study. Meropenem (500 mg) was administered every 8 h. CVVHDF was targeted as a 2-3 L/h exchange using a polyacrylonitrile filter with a surface area of 1.05 $m^{2}$ and a blood flow rate of 200 mL/min. Serial blood (pre- and post-filter), filtrate/dialysate and ISF concentrations were measured on 2 days of treatment (Profiles A and B). Subcutaneous tissue ISF concentrations were determined using microdialysis. A total of 384 samples were collected. During Profile A, the comparative median (IQR) ISF and plasma peak concentrations were 13.6 (12.0-16.8) and 40.7 (36.6-45.6) mg/L and the trough concentrations were 2.6 (2.4-3.4) and 4.9 (3.5-5.0) mg/L, respectively. During Profile B, the ISF trough concentrations increased by $\sim$40\%. Meropenem ISF penetration was estimated at 63\% (60\%-69\%) and 69\% (65\%-74\%) for Profiles A and B, respectively, using comparative plasma and ISF AUCs. For Profile A, the plasma elimination t1/2 was 3.7 (3.3-4.0) h, the volume of distribution was 0.35 (0.25-0.46) L/kg, the total clearance was 4.1 (4.1-4.8) L/h and the CVVHDF clearance was 2.9 (2.7-3.1) L/h. \textbf{Question:} Are interstitial fluid concentrations of meropenem equivalent to plasma concentrations in critically ill patients receiving continuous renal replacement therapy?
(A) Yes (B) No (C) Maybe\\
\textbf{Explanation:} This is the first known report of concurrent plasma and ISF concentrations of a meropenem antibiotic during CVVHDF. We observed that the ISF concentrations of meropenem were significantly lower than the plasma concentrations, although the present dose was appropriate for infections caused by intermediately susceptible pathogens (MIC$\leq$4 mg/L).\\
\textbf{Answer:} (B)
\end{tabular}}} 
& \\ \\

{\color{ourdarkblue} {\begin{tabular}[l]{@{}p{0.98\textwidth}}
\textbf{Context:} Family caregivers of dementia patients are at increased risk of developing depression or anxiety. A multi-component program designed to mobilize support of family networks demonstrated effectiveness in decreasing depressive symptoms in caregivers. However, the impact of an intervention consisting solely of family meetings on depression and anxiety has not yet been evaluated. This study examines the preventive effects of family meetings for primary caregivers of community-dwelling dementia patients. A randomized multicenter trial was conducted among 192 primary caregivers of community dwelling dementia patients. Caregivers did not meet the diagnostic criteria for depressive or anxiety disorder at baseline. Participants were randomized to the family meetings intervention (n=96) or usual care (n=96) condition. The intervention consisted of two individual sessions and four family meetings which occurred once every 2 to 3 months for a year. Outcome measures after 12 months were the incidence of a clinical depressive or anxiety disorder and change in depressive and anxiety symptoms (primary outcomes), caregiver burden and quality of life (secondary outcomes). Intention-to-treat as well as per protocol analyses were performed. A substantial number of caregivers (72/192) developed a depressive or anxiety disorder within 12 months. The intervention was not superior to usual care either in reducing the risk of disorder onset (adjusted IRR 0.98; 95\% CI 0.69 to 1.38) or in reducing depressive (randomization-by-time interaction coefficient=-1.40; 95\% CI -3.91 to 1.10) or anxiety symptoms (randomization-by-time interaction coefficient=-0.55; 95\% CI -1.59 to 0.49). The intervention did not reduce caregiver burden or their health related quality of life. \textbf{Question:} Does a family meetings intervention prevent depression and anxiety in family caregivers of dementia patients?
(A) Yes (B) No (C) Maybe\\
\textbf{Explanation:} This study did not demonstrate preventive effects of family meetings on the mental health of family caregivers. Further research should determine whether this intervention might be more beneficial if provided in a more concentrated dose, when applied for therapeutic purposes or targeted towards subgroups of caregivers.\\
\textbf{Answer:} (B)
\vspace{-0.4cm}
\end{tabular}}} 
&\\ \\

{\color{ourdarkblue} {\begin{tabular}[l]{@{}p{0.98\textwidth}}
\textbf{Context:} To compare adherence to follow-up recommendations for colposcopy or repeated Papanicolaou (Pap) smears for women with previously abnormal Pap smear results. Retrospective cohort study. Three northern California family planning clinics. All women with abnormal Pap smear results referred for initial colposcopy and a random sample of those referred for repeated Pap smear. Medical records were located and reviewed for 90 of 107 women referred for colposcopy and 153 of 225 women referred for repeated Pap smears. Routine clinic protocols for follow-up--telephone call, letter, or certified letter--were applied without regard to the type of abnormality seen on a Pap smear or recommended examination. Documented adherence to follow-up within 8 months of an abnormal result. Attempts to contact the patients for follow-up, adherence to follow-up recommendations, and patient characteristics were abstracted from medical records. The probability of adherence to follow-up vs the number of follow-up attempts was modeled with survival analysis. Cox proportional hazards models were used to examine multivariate relationships related to adherence. The rate of overall adherence to follow-up recommendations was 56.0\% (136/243). Adherence to a second colposcopy was not significantly different from that to a repeated Pap smear (odds ratio, 1.40; 95\% confidence interval, 0.80-2.46). The use of as many as 3 patient reminders substantially improved adherence to follow-up. Women without insurance and women attending 1 of the 3 clinics were less likely to adhere to any follow-up recommendation (hazard ratio for no insurance, 0.43 [95\% confidence interval, 0.20-0.93], and for clinic, 0.35 [95\% confidence interval, 0.15-0.73]). \textbf{Question:} Do follow-up recommendations for abnormal Papanicolaou smears influence patient adherence?
(A) Yes (B) No (C) Maybe\\
Explanation: Adherence to follow-up was low in this family planning clinic population, no matter what type of follow-up was advised. Adherence was improved by the use of up to 3 reminders. Allocating resources to effective methods for improving adherence to follow-up of abnormal results may be more important than which follow-up procedure is recommended.\\
\textbf{Answer:} (B)
\vspace{-0.4cm}
\end{tabular}}} 
&\\

\bottomrule 
 
\end{tabular}
\end{table}


\begin{table}[!]
\footnotesize
\centering
\caption{MMLU (2020)~\cite{hendrycks2020measuring} chain-of-thought prompt examples.}
\vspace{3pt}
\label{tab-app:mmlu-cot}
\begin{tabular}{l@{\hspace{.1em}}l@{\hspace{0.1em}}}
\toprule
{\color{ourdarkblue} {\begin{tabular}[l]{@{}p{0.98\textwidth}}

\textbf{Instructions:} The following are multiple-choice questions about medical knowledge. Solve them in a step-by-step fashion. Output a single option as the final answer. \\ \\ 
 
\textbf{Question:} The energy for all forms of muscle contraction is provided by:\\
(A) ATP. (B) ADP. (C) phosphocreatine. (D) oxidative phosphorylation.\\
\textbf{Explanation:} The sole fuel for muscle contraction is adenosine triphosphate (ATP). During near maximal intense exercise the muscle store of ATP will be depleted in less than one second. Therefore, to maintain normal contractile function ATP must be continually resynthesized. These pathways include phosphocreatine and muscle glycogen breakdown, thus enabling substrate-level phosphorylation (‘anaerobic’) and oxidative phosphorylation by using reducing equivalents from carbohydrate and fat metabolism (‘aerobic’). \\ 
\textbf{Answer:} (A) \\ \\ 
 
\textbf{Question:} Which of the following conditions does not show multifactorial inheritance?\\
(A) Pyloric stenosis (B) Schizophrenia (C) Spina bifida (neural tube defects) (D) Marfan syndrome\\
\textbf{Explanation:} Multifactorial inheritance refers to when a condition is caused by multiple factors, which may be both genetic or environmental. Marfan is an autosomal dominant trait. It is caused by mutations in the FBN1 gene, which encodes a protein called fibrillin-1. Hence, Marfan syndrome is not an example of multifactorial inheritance. \\
\textbf{Answer:} (D) \\ \\
 
\textbf{Question:} What is the embryological origin of the hyoid bone?\\
(A) The first pharyngeal arch (B) The first and second pharyngeal arches (C) The second pharyngeal arch (D) The second and third pharyngeal arches\\
\textbf{Explanation:} In embryology, the pharyngeal arches give rise to anatomical structure in the head and neck. The hyoid bone, a small bone in the midline of the neck anteriorly, is derived from the second and third pharyngeal arches. \\ 
\textbf{Answer:} (D) \\ \\ 
 
\textbf{Question:} In a given population, 1 out of every 400 people has a cancer caused by a completely recessive allele, b. Assuming the population is in Hardy-Weinberg equilibrium, which of the following is the expected proportion of individuals who carry the b allele but are not expected to develop the cancer?\\
(A) 1/400 (B) 19/400 (C) 20/400 (D) 38/400 \\
\textbf{Explanation:} The expected proportion of individuals who carry the b allele but are not expected to develop the cancer equals to the frequency of heterozygous allele in the given population. According to the Hardy-Weinberg equation p$\wedge$2 + 2pq + q$\wedge$2 = 1, where p is the frequency of dominant allele frequency, q is the frequency of recessive allele frequency, p$\wedge$2 is the frequency of the homozygous dominant allele, q$\wedge$2 is the frequency of the recessive allele, and 2pq is the frequency of the heterozygous allele. Given that q$\wedge$2=1/400, hence, q=0.05 and p=1-q=0.95. The frequency of the heterozygous allele is 2pq=2*0.05*0.95=38/400. \\
\textbf{Answer:} (D) \\ \\ 
 
\textbf{Question:} A high school science teacher fills a 1 liter bottle with pure nitrogen and seals the lid. The pressure is 1.70 atm, and the room temperature is 25$\circ$C. Which two variables will both increase the pressure of the system, if all other variables are held constant?\\
(A) Decreasing volume, decreasing temperature  (B) Increasing temperature, increasing volume (C) Increasing temperature, increasing moles of gas (D) Decreasing moles of gas, increasing volume\\
\textbf{Explanation:} According to the ideal gas law, PV = nRT (P = pressure, V = volume, n = number of moles, R = gas constant, T = temperature). Hence, increasing both temperature (T) and moles of gas (n), while other variables stay constant, will indeed increase the pressure of the system.\\
\textbf{Answer:} (C) \\ \\
 
\textbf{Question:} A 22-year-old male marathon runner presents to the office with the complaint of right-sided rib pain when he runs long distances. Physical examination reveals normal heart and lung findings and an exhalation dysfunction at ribs 4-5 on the right. Which of the following muscles or muscle groups will be most useful in correcting this dysfunction utilizing a direct method? \\
(A) anterior scalene (B) latissimus dorsi (C) pectoralis minor (D) quadratus lumborum\\
\textbf{Explanation:} All of the muscles have an insertion on the rib cage; however only one has an insertion at ribs 4-5 and could be responsible for right-sided rib pain: pectoralis minor. Pectoralis minor inserts to the costal cartilage of the anterior third to fifth ribs.\\
\textbf{Answer:} (C) \\ \\

\end{tabular}}} 
& \\

\bottomrule 
 
\end{tabular}
\end{table}

